\documentclass[
    numbers=noenddot,
    parskip=half-,
    fontsize=12pt,
    paper=a4,
    oneside,
    titlepage,
    bibliography=totoc,
    chapterprefix=false,
]{scrbook}

\usepackage{graphicx}
\usepackage[onehalfspacing]{setspace}

\usepackage{tocbasic}
\usepackage{booktabs}
\usepackage{multicol}
\usepackage{multirow}

\usepackage[]{scrlayer-scrpage}

\usepackage[citestyle=alphabetic, bibstyle=alphabetic, sorting=nyt, backend=biber, language=english, backref=true, maxcitenames=2]{biblatex}

\usepackage[autostyle,english=american,german=quotes]{csquotes}
\usepackage[titletoc]{appendix}

\graphicspath{{images/}}

\usepackage{hyperref}
\usepackage{xcolor}

\usepackage{afterpage}

\newcommand\blankpage{
    \null
    \thispagestyle{empty}
    \addtocounter{page}{-1}
    \newpage
    }

\usepackage{amsmath,amsfonts,amsthm} % Math packages

\usepackage{float}

\usepackage{longtable,array,ragged2e}
\newcolumntype{L}{>{\centering\arraybackslash}m{3cm}}
\usepackage{listings}
\colorlet{punct}{red!60!black}
\definecolor{background}{HTML}{EEEEEE}
\definecolor{delim}{RGB}{20,105,176}
\colorlet{numb}{magenta!60!black}

\lstdefinelanguage{json}{
    basicstyle=\normalfont\ttfamily,
    numbers=left,
    numberstyle=\scriptsize,
    stepnumber=1,
    numbersep=8pt,
    showstringspaces=false,
    breaklines=true,
    frame=lines,
    backgroundcolor=\color{background},
    literate=
     *{0}{{{\color{numb}0}}}{1}
      {1}{{{\color{numb}1}}}{1}
      {2}{{{\color{numb}2}}}{1}
      {3}{{{\color{numb}3}}}{1}
      {4}{{{\color{numb}4}}}{1}
      {5}{{{\color{numb}5}}}{1}
      {6}{{{\color{numb}6}}}{1}
      {7}{{{\color{numb}7}}}{1}
      {8}{{{\color{numb}8}}}{1}
      {9}{{{\color{numb}9}}}{1}
      {:}{{{\color{punct}{:}}}}{1}
      {,}{{{\color{punct}{,}}}}{1}
      {\{}{{{\color{delim}{\{}}}}{1}
      {\}}{{{\color{delim}{\}}}}}{1}
      {[}{{{\color{delim}{[}}}}{1}
      {]}{{{\color{delim}{]}}}}{1},
}
\newcommand{\SourceCodeRepo}{https://gitlab.com/malik.zohaib90/nlp-on-legal-datasets/}
\newcommand{\RepoRootPath}{-/blob/master/}
\newcommand{\GenPreTrainData}{utilities/generate_pretraining_data.py}

\newcommand{\LegalVocabulary}{data/legal_vocabulary/legal_dictionary.tsv}

\title{Comparing the Performance of NLP Toolkits and Evaluation measures in Legal Tech}

\author{Muhammad Zohaib Khan}

\date{\today}

\makeatletter
\let\thetitle\@title
\let\theauthor\@author
\let\thedate\@date
\makeatother

\begin{document}

%%%%%%%%%%%%%%%%%%%%%%%%%%%%%%%%%%%%%%%%%%%%%%%%%%%%%%%%%%%%%%%%%%%%%%%%%%%%%%%%%%%%%%%%%
\frontmatter
% CHOOSE ACCORDINGLY
%\include{includes/BA-titlepage}
% !TeX spellcheck = en_US
% !TeX encoding = UTF-8
\begin{titlepage}
    \centering
    \begin{onehalfspace}
    	
        	\includegraphics[width=7cm]{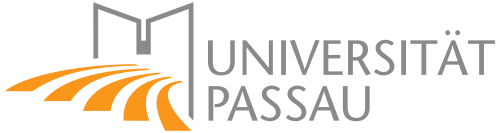}\\
        	\vspace{1.0cm}
        	\large {\bfseries Lehrstuhl f\"ur Data Science }\\

        	\vspace{2.5cm}

            \begin{doublespace}
            	{\textsf{\Huge{\thetitle}}}
            \end{doublespace}

        	\vspace{2cm}

            \Large{Masterarbeit von}\\

        	\vspace{1cm}

        	{\bfseries \large{\theauthor}}

            {\large
        		\begin{tabular}[l]{cc}
        			Supervised By: & Prof.~Dr.~Jelena Mitrovi\'{c}
        		\end{tabular}
        	}
        	\vfill

        	{\large
        		\begin{tabular}[l]{cc}
        			\textsc{1.~Pr\"ufer} & \textsc{2.~Pr\"ufer} \\
        			Prof.~Dr.~Jelena Mitrovi\'{c} & Prof.~Dr.~Michael Granitzer
        		\end{tabular}
        	}

        	\vspace{1cm}
        	
        	\parbox{\linewidth}{\hrule\strut}

            \vfill

	    \thedate
    \end{onehalfspace}
\end{titlepage}

\afterpage{\blankpage}
\tableofcontents
\afterpage{\blankpage}

% -- ABSTRACT
% !TeX spellcheck = en_US
% !TeX encoding = UTF-8
\chapter*{Abstract}

Recent developments in Natural Language Processing have led to the introduction of state-of-the-art Neural Language Models, enabled with unsupervised transferable learning, using different pretraining objectives. While these models achieve excellent results on the downstream NLP tasks, various domain adaptation techniques can improve their performance on domain-specific tasks. We compare and analyze the pretrained Neural Language Models, XLNet (autoregressive), and BERT (autoencoder) on the Legal Tasks. Results show that XLNet Model performs better on our Sequence Classification task of Legal Opinions Classification, whereas BERT produces better results on the NER task. We use domain-specific pretraining and additional legal vocabulary to adapt BERT Model further to the Legal Domain. We prepared multiple variants of the BERT Model, using both methods and their combination. Comparing our variants of the BERT Model, specializing in the Legal Domain, we conclude that both additional pretraining and vocabulary techniques enhance the BERT model's performance on the Legal Opinions Classification task. Additional legal vocabulary improves BERT's performance on the NER task. Combining the pretraining and vocabulary techniques further improves the final results. Our Legal-Vocab-BERT Model gives the best results on the Legal Opinions Task, outperforming the larger pretrained general Language Models, i.e., BERT-Base and XLNet-Base.

\afterpage{\blankpage}

% -- Acknowledgements (optional)
% !TeX spellcheck = en_US
% !TeX encoding = UTF-8
\chapter*{Acknowledgments}

Firstly, thanks to God for blessing me with the abilities to successfully complete the thesis. Then, I wish to express my utmost gratitude to my thesis advisor Prof.~Dr.~Jelena Mitrovi\'{c}, for her continuous guidance throughout the research. She was always available to help me with her insightful suggestions and feedback on the research work. She is an excellent research advisor, provided me with great ideas, and kept me in the right direction to accomplish the thesis work.

I wish to thank Prof. Dr. Michael Granitzer for his valuable time for reviewing and evaluating my thesis as the second examiner. His suggestions at the kickoff presentation of my research helped speed up the research process.

I would like to thank the University of Passau for providing the computational resources to prepare the models for the research. I am grateful to the Harvard Law School Library for providing me with the legal data for research scholars from the CASELAW Access Project. I am also thankful to Google's Colaboratory Service for providing the computational resources used in the research.

Finally, I would like to acknowledge the love and support of my family and friends. Special thanks to my parents, who always believed in me; without their endless support and encouragement, this work would not have been possible.

% I would first like to thank my thesis advisor ...
\afterpage{\blankpage}

% -- List of figures
\thispagestyle{empty}
\cleardoublepage
\listoffigures
\newpage

% -- List of tables
\thispagestyle{empty}
\cleardoublepage
\listoftables
\newpage

%%%%%%%%%%%%%%%%%%%%%%%%%%%%%%%%%%%%%%%%%%%%%%%%%%%%%%%%%%%%%%%%%%%%%%%%%%%%%%%%%%%%%%%%%
\mainmatter

% -- Chapters
% following IMRaD structure
% adjust for your liking
\chapter{Introduction}\label{chap:introduction}

\section{Motivation}\label{sec:motivation}

In the last quarter of 2018, the field of Natural Language Processing was hit with a major paradigm shift after the release of Google's BERT (Bidirectional Encoder Representations from Transformers) \cite{bert-model}, achieving state-of-the-art results on 11 NLP tasks. BERT introduced contextual representations learning in a bidirectional way, using masked language modeling techniques, which the previous efforts were lacking in capturing the contextual representations. A few months after BERT was released, another major development in the field was made by XLNet Model \cite{xlnet-model} released in June 2019, claiming to outperform BERT and achieving new state-of-the-art results on 20 NLP tasks. These new state-of-the-art Language models gained significant improvements by leveraging the power of deep learning techniques and the benefits of transfer learning. The major difference between BERT and XLNet is their objectives in learning the bidirectional context during the pretraining process, with BERT using autoencoder vs. XLNet using autoregressive language modeling.

In recent years, a significant increase of interest is seen in the application of NLP in the Legal Domain. Many researchers and tech companies are working on exploiting deep learning techniques, developing legal applications with NLP capabilities. In a field like Law, following a well-structured process with well-defined practices and detailed documentation makes it very ideal for applications of NLP. With NLP playing a vital role in law, and the new developments revolutionizing the field of NLP, it is very important for the Legal Domain and its community to understand better what new technologies are relevant for helping their cause and how would this impact the role that NLP is playing in the domain. Which of the new developments would fit their needs, and which techniques can enhance the performance of NLP tasks they require, and if the benefit gained is worth the effort needed to make use of these techniques.

% \underline{the problem part:}
\section{Problem}\label{sec:problem}

Transformer \cite{transformer} based language models like BERT and XLNet, use the techniques of unsupervised pretraining process on massive text to learn the features of languages and benefits of transfer learning to adapt to the downstream NLP tasks quickly by simple finetuning. Both Models (BERT and XLNet) provide advantages of transfer learning, easy finetuning, and faster computation in finetuning stage on the downstream NLP task, with the difference being the way each model learns the weights during the pretraining process. Deciding the model that fits the specific needs of natural language processing in the Legal Domain is an important part. It is helpful to have a meaningful comparison between the performance of two models with different pretraining strategies on the NLP tasks in the Legal Domain.

BERT uses self-supervised learning in the pretraining process and uses the WordPiece embedding vocabulary \cite{wordpiece-embedding} which has around 30K tokens. This opens the path to improvement of the performance of the BERT Model on domain specific NLP tasks. This includes additional pretraining of the BERT Model on the legal text by initializing with the pretrained general BERT Model, as it is claimed to help improve performance on the domain-specific NLP tasks. The second option for enhancing BERT's performance on the Legal Domain is to add domain-specific, i.e., legal vocabulary in the existing WordPiece vocabulary. However, it is important to know how these methods may impact BERT's performance in accuracy and computational time. It is important to know when to opt for these options and if it is worth the effort and resources to use these domain adaptation techniques. These problems are the underlying foundation of the research questions described in Section-\ref{sec:research-questions}.

% general introduction of common methodology of bert and xlnet.

% different techniques both models using to improve the performance on nlp tasks.

% different techniques with in bert models that can improve bert's performance on nlp tasks on domain specific corpora.

% deciding which one is the goto model and which techniques are worth applying to improve accuracy on nlp tasks in legal datasets. performance benefit over cost and effort in implementing those techniques.

% with bert and xlnet giving state-of-the-art results in nlp tasks, how do we know which model is best suitable for us and which gives us what kind of advantages over the other. 
% How to decide which model to pick for our cause.

% different legal tasks have their own priorities in what is most preferred over the other in taking advantage from nlp. do we prefer accuracy over performance? or vice versa.

% what techniques can be worth the effort and resources in improving results via bert?

% \underline{the contribution part:}
\section{Contribution}\label{sec:contribution}

Following contributions are made that will help to solve the problems described in Section-\ref{sec:problem}.

\begin{itemize}
	\item A fair comparison of performance on NLP tasks in the Legal Domain between transformer models using different pretraining objectives to learn deep bidirectional context.
	\item Finding out the impacts of further pretraining BERT on Legal text. This includes the impact on the performance of accuracy as well as computation time during the finetuning stage.
	\item Finding out the impacts of adding legal vocabulary to WordPiece Vocabulary used by BERT on NLP tasks in the Legal Domain. This also includes both the performance of accuracy and computation time.
	\item Selecting the Legal vocabulary that would actually contribute to better results in the model's performance on Legal tasks.
	\item Providing the guidelines on exploiting the techniques to enhance performance in the BERT model. When is it appropriate to choose a performance enhancement technique, and what are the constraints that need to be taken care of before opting for a technique.
	\item We also release models prepared during the research, including Legal-BERT, Vocab-BERT, and Legal-Vocab-BERT models. %TODO: link the name of models to their description once the methodology chapter is written.
\end{itemize}

%the impact
A meaningful comparison of new state-of-the-art transformer-based models will help select the model that fits the specific requirements of natural language processing in the Legal Domain. BERT Model provides excellent performance in natural language processing, with a strong understanding of the language using general text from different domains. However, with possibilities of further improvement within a domain, it will be very helpful to know which techniques would actually have a fruitful impact. Given that pretraining is an expensive procedure, it is not always easy to use the technique without knowing if going through these efforts would actually lead to the expectations.  

\section{Research Questions}\label{sec:research-questions}

\begin{enumerate} \bfseries
    \item Which of the Transformer based Language Models such as BERT and XLNet perform better in Legal Domain?

    \textnormal{Language Models are a vital part of the NLP pipeline. BERT and XLNet are the latest developments in this area, achieving state-of-the-art results in NLP tasks. Both models learn the deep bidirectional context during their pretraining stage and can be easily finetuned on downstream NLP tasks. Answering this question will provide the findings on a comparison between performances of autoencoder models like BERT and autoregressive models like the XLNet.}
    
    \item How does Legal Vocabulary affect the performance of BERT on Legal Tech?
    
    \textnormal{For tokenization, BERT uses the vocabulary from WordPiece embedding, which contains around 30K tokens. The focus of this question is to find out the impacts of adding Legal Vocabulary, along with WordPiece vocabulary, to the pretrained BERT model. The impact includes the performance of BERT in evaluation measures as well as computation time on the downstream NLP tasks in the Legal Domain. The answer will provide guidelines on when to opt for this technique and how to select the vocabulary for a task.}
    
    \item Is additional pretraining with Legal data on top of an existing pretrained general language model of BERT worthwhile for optimizing NLP performance in Legal Tech?
    
    \textnormal{BERT is pretrained on a huge amount of plain text, covering the data from different domains and understands the general language features. However, as per the BERT's authors recommendation, it is beneficial to add domain-specific pretraining when working with NLP on a specific domain. This question focuses on the benefits gained by additional pretraining and the effort required to run the model's additional pretraining. It compares different variants of the BERT Model, including a BERT model pretrained on general plain text, a BERT Model with additional pretraining on Legal text, and a BERT Model combining the benefits gained by adding Legal vocabulary along with the additional pretraining on Legal text.}

\end{enumerate}

% This question will also focus on comparison of different variants of BERT models we prepare to evaluate. It will also incorporate the benefits that BERT might gain from techniques discussed on second question. We will compare the performance of a general pretrained BERT model, BERT model after being fed with Legal Vocabulary, and BERT model with added pretraining on Legal Data. We will also combine the techniques from question 2 and question 3 to gain another variant of BERT model which will include pretraining on Legal dataset and also feeding with Legal Vocabulary to find which of the techniques will yield the best results in Legal Domain.

% General Requirements to the thesis:

% \begin{itemize}
% 	\item 60 pages of content in this format. Content does not include table of content, lists, appendices etc.
% 	\item Proper scientific referencing
% 	\item Introduction and Background should be less than 50\% of the thesis
% 	\item Images should be readable and in the proper size. 
% \end{itemize}

\section{Structure of the Thesis}

Thesis contains 8 Chapters in total. Chapter-\ref{chap:introduction} introduces the research topic, including motivation, problem, and contribution and impact made by the Paper in this research area. The background knowledge, state-of-the-art, with necessary understanding of the models being studied, is explained in Chapter-\ref{chap:background}. Related work is discussed in Chapter-\ref{chap:related-work}. Chapter-\ref{chap:methods} contains the details on the preparation of Datasets and Models for the NLP tasks performed. Experimental setup, used parameters, and implementation details of NLP tasks, are discussed in Chapter-\ref{chap:experimentation}. Chapter-\ref{chap:results} shows the results found from experimentation. A brief discussion on the findings and the rationality for the outcomes, is talked about in Chapter-\ref{chap:discussion}. Finally the conclusions and future work are discussed in Chapter-\ref{chap:conclusion}.

% everything below is commented, but shows examples of how to use a particular component of report

% \section{Example citation \& symbol reference}\label{sec:citation}
% For symbols look at \cite{latex_symbols_2017}.

% \section{Example reference}
% Example reference: Look at chapter~\ref{chap:introduction}, for sections, look at section~\ref{sec:citation}.

% \section{Example image}

% \begin{figure}
% 	\centering
% 	\includegraphics[width=0.5\linewidth]{uni-logo}
% 	\caption{Meaningful caption for this image}
% 	\label{fig:uniLogo}
% \end{figure}

% Example figure reference: Look at Figure~\ref{fig:uniLogo} to see an image. It can be \texttt{jpg}, \texttt{png}, or best: \texttt{pdf} (if vector graphic).

% \section{Example table}

% \begin{table}
% 	\centering
% 	\begin{tabular}{lr}
% 		First column & Number column \\
% 		\hline
% 		Accuracy & 0.532 \\
% 		F1 score & 0.87
% 	\end{tabular}
% 	\caption{Meaningful caption for this table}
% 	\label{tab:result}
% \end{table}

% Table~\ref{tab:result} shows a simple table\footnote{Check \url{https://en.wikibooks.org/wiki/LaTeX/Tables} on syntax}
\afterpage{\blankpage}

\chapter{Background}\label{chap:background}

Natural Language Processing has been around for a few decades now. It evolved over the years with new techniques of understanding the natural language, development of different language models, from N-grams to Neural Language Models. Even though NLP, in terms of advancements, is a fairly young field, the last few years have seen a rapid progress. Recent developments in the field have made a huge impact on its efficiency, attracting the attention of researchers as well as firms working with NLP.

\section{Natural Language Processing in Legal Domain}\label{sec:nlp-in-legal-domain}

Law is a domain, which works with well-structured language and well-organized processes. To leverage the benefits of software and technology, the Legal Domain is dedicating efforts for its digital transformation to facilitate, simplify, and optimize legal processes with the development of Legal Tech\footnote{\url{https://en.wikipedia.org/wiki/Legal_technology}}. Law has a formal language, with particular syntax, semantics, and its own specific vocabulary.
% , and is studied to learn the domain knowledge to communicate in terms of law more with more simplicity. 
As a matter of fact, Legal English is often referred to as a "sublanguage", and services like lawlinguists\footnote{\url{https://lawlinguists.com/training/}} are explicitly teaching it to the Lawyers and people working with the Legal Domain. These characteristics of Legal language make it a perfect subject for exploiting the Natural Language Processing's powers in the Legal Domain. Currently, NLP is playing a vital role in the Legal Technology in quite a few areas, including extracting relevant information about judicial decisions, automated review of contracts for error checking, automating the regular legal procedures and documentation, and Lawyer bots for Legal Advice. 

\section{Background of Language Modeling}\label{sec:language-modelling}
One of Natural Language Processing's fundamental tasks is predicting text conditioned on the previous sequence of words, which is the core of Language Modeling. The prominent statistical models were rule-based or probabilistic Language Models through the development of language modeling techniques, e.g., N-gram. The simplicity of a model like N-gram limits its performance over long texts. Neural Language Models helped to overcome these limitations and outperformed Statistical Language Models in language modeling tasks. Recurrent Neural Networks (RNNs) faced problems such as vanishing gradient, resulting in the inability to capture relevant dependencies in long sequences. Long Short Term Memory (LSTM) \cite{lstm-sequence-generation} solved problems other RNNs faced with the capability of capturing long-term dependencies in sequences. The refinement of language modeling through Deep Learning methods continued up to the Transformer's \cite{transformer} introduction, based one which the new state-of-the-art language models are built, including BERT and XLNet.

\subsection{Context Learning}\label{subsec:context-learning}

Word Representation is one of the main building blocks for Natural Language Processing, having a remarkable impact on performance, more specifically on Deep Learning models. There are different types and numerous techniques for creating word representations. Word representations can be fixed as generated by methods like "dictionary lookup" and "One-Hot Encoding" or distributed such as word embedding. Distributed Representations provide the key contribution to the efficiency of Deep Learning methods in NLP. A good word representation method should capture syntactic and semantic regularities of the natural language and incorporate the notion of word similarity. 

Contextualized Word Representations, as compared to Static Word Representations, yield significant improvements in performance on NLP tasks. Same words can change their meanings depending on the context of the sentence, which is not comprehended by context-free representations. Context-free models will generate the same representations for a word, irrespective of the words it is surrounded by. For example, given two sentences, "Opening a bank account" and "Walking along the river bank", they will generate the same representation for the word "bank". Models like GloVe\footnote{\url{https://nlp.stanford.edu/projects/glove/}} and word2vec\footnote{\url{https://www.tensorflow.org/tutorials/text/word2vec}} generate context-free representations. Contextual representations incorporate the phenomena possessed by Homonym words, and represent them according to their context. Capturing context information with contextual representations provided remarkable improvement in language understanding and developing state-of-the-art language models. Linguist J.R Firth summarized this principle as "You shall know a word by the company it keeps" (J.R Firth 1957).
% Learning to understand the context is one the most important goals for Language Models in understanding a language.

Learning Contextual representations can be done in a unidirectional or bidirectional way. OpenAI GPT \cite{GPT-1} learns the context in left-to-right direction and generates pretrained representations only from the context before a word. ELMo \cite{ELMo} learns contextualized word representations in a bidirectional way by parsing the sentence in a left-to-right and right-to-left manner and then concatenating the learned representations from both directions, making it shallowly bidirectional. ULMFiT \cite{ULMFiT} learns forward and backward contexts and takes average predictions of both forward and backward models after independently finetuning for both. ELMo and ULMFit are limited to shallow bidirectional learning due to the sequence to sequence technique of underlying RNN architectures. The introduction of Transformer \cite{transformer} enabled deep bidirectional learning, as it removes sequential dependency by reading all the words from a sentence at-once. BERT \cite{bert-model} exploits the Transformer's capability of reading sequences at-once to learn deep bidirectional context simultaneously, unlike the mechanism adopted by ELMo.

\subsection{Neural Language Models}

Conventional methods for language modeling used rule-based systems or probabilistic models, which had limitations of capturing contextual information in long sentences, and problems like running into "curse of dimensionality" as the vocabulary size increased. Deep Neural Network Models overcame these limitations with their complex and sophisticated architecture. Recurrent Neural Network shows improvement in capturing context information, but as input is fed serially, i.e., word by word, it makes them slower. In addition, RNNs face problems in maintaining the state over long sentences to capture the context. LSTM deals better with long sequences but is even slower than other Recurrent Neural Networks. Transformer Neural Network solves these problems by a pure self-attention approach, where a complete sequence is passed to the model simultaneously. Contextual word representations are taken from Neural Network's hidden states by feeding the word embedding through Neural Network. This idea made a major impact on creating Deep Contextual Word Representations. Researchers scaled up the idea to huge neural networks, creating state-of-the-art language models like ELMo, GPT, BERT, and XLNet \cite{xlnet-model}, by training on massive amount of unlabelled text. The study \cite{contextual-comparison} on deep contextual models shows that the upper layers of these models produce more context-specific representations.

\subsection{Pretraining} \label{subsec:pretraining}

The Pretraining process involves training a Neural Network, on a large amount of unlabelled text, in an unsupervised manner, using a language modeling objective to learn common language features, which can later be re-used to perform a supervised NLP task. In their Paper \cite{Semi-supervised-sequence-learning} on Semi-Supervised Sequence Learning, the authors presented two different approaches for using unlabelled data for training RNN. The first approach is the key task of a conventional language model, i.e., next-word prediction. The second approach is sequence autoencoding, where an input sequence is read into a vector and then predicted. Using these two techniques, a Neural Network can be pretrained to learn parameters in an unsupervised fashion, that can generalize well in the later supervised learning tasks. Pretraining is a one-time process, and weights learned can be used on many downstream NLP tasks, with simple finetuning, saving a lot of time and effort. This idea is an important technique that helped the current state-of-the-art models achieve better performance in many NLP tasks.

One of the major advantages of the pretraining process is the availability of training data. The unlabelled text is available in abundant amounts compared to scarcely available labeled data for specific tasks. OpenAI's GPT \cite{GPT-1} combined the idea of unsupervised pretraining with Transformer Network \cite{transformer} to gain state-of-the-art results on various NLP tasks. Results from GPT showed that the idea of pairing unsupervised pretraining with later supervised learning could yield better performance. The idea of unsupervised pretraining enables transfer learning for the pretrained models on NLP tasks, much like the idea of transfer learning that already exists in computer vision. Researchers exploited the idea in recent years, using different pretraining motives and underlying Neural Network architectures, and trained on a large amount of unlabelled data to create the state-of-the-art pretrained language models. Some of the well-known pretrained models are GPT pretrained on 800M tokens, BERT pretrained on 3.3B tokens, XLNet pretrained on 32.89B tokens, and GPT-2 pretrained on 40B tokens of the plain text corpus.

% TODO, read the 2 links below. read bert paper about pre-training process used by different other models as well as bert. Describe the pretraining process, purpose, output, advantages, transfer learning.

% https://arxiv.org/abs/1511.01432
% https://openai.com/blog/language-unsupervised/

\section{Invention of Transformer}\label{sec:transformer}

The introduction of Transformer in 2017 was a paradigm shift in the field of neural language modeling. Inspired by attention-mechanism \cite{attention-mechanism}, in their paper "Attention Is All You Need" \cite{transformer}, the authors proposed a new Neural Network architecture solely based on the attention-mechanism. The transformer's architecture provides the ability to read-in a whole sequence at-once, removing the recurrence and convolution. This ability of the Transformer's architecture also enables parallelism and exploiting the power of the GPUs. Reading complete sequence at-once also paves the way for better context learning, which is well exploited by models like BERT, to learn left-to-right and right-to-left context simultaneously.

%write about architecture of transformers in a paragraph, with focus on encoder. add the image for encoder's architecture at least.

% watch https://www.youtube.com/watch?v=S27pHKBEp30&ab_channel=SeattleAppliedDeepLearning&t=1725s

%Transfer learning - Empirical Advantages of Transformer over LSTM
%https://web.stanford.edu/class/cs224n/slides/Jacob_Devlin_BERT.pdf

\subsection{Transformer's Architecture}\label{subsec:transformer-architecture}
\begin{figure}
	\centering
	\includegraphics[width=1.0\linewidth]{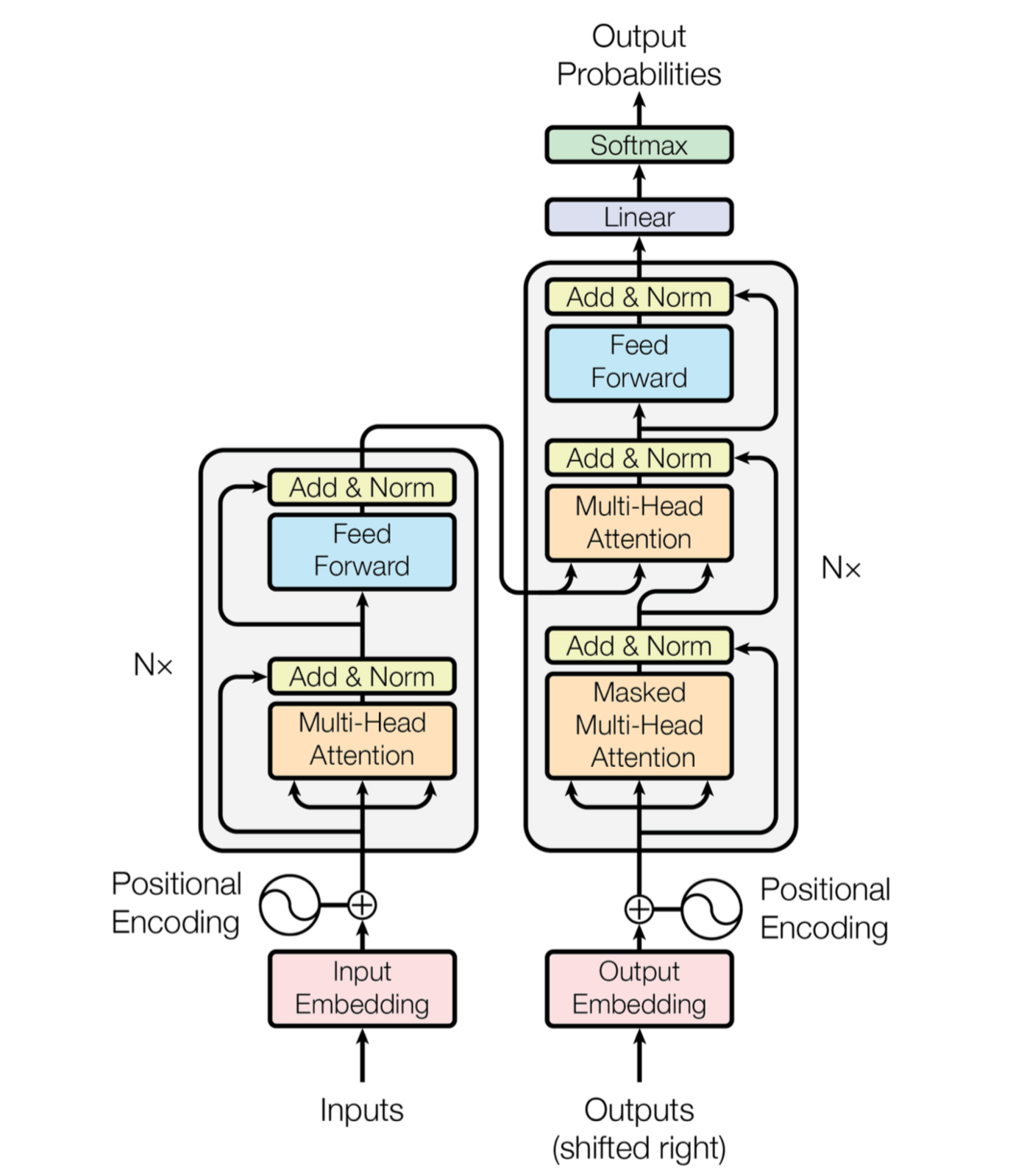}
	\caption{Transformer's Architecture. From \cite{transformer}}
% 	\caption{Transformer's Architecture. From ‘Attention Is All You Need’ \cite{transformer} by Vaswani et al.}
	\label{fig:transformerArchitecture}
\end{figure}

The Transformer is solely based on the Attention Mechanism instead of recursion and backpropagation. As shown in Figure-\ref{fig:transformerArchitecture},  the transformer model has 2 parts, an encoder and a decoder block for performing sequence to sequence tasks. The encoder-decoder architecture is similar to the previous sequence to sequence models like LSTM but differs in the mechanism, which is self-attention instead of recursions. The encoder is given the vectorized input sequence, along with the position encoding. It encodes the input sequence, which is then used by the decoder to make the predictions for all the individual tokens of the output sequence. The decoder is given the output sequence that is so far generated, along with the positional encoding. It combines it with the encoded input sequence to predict the next possible tokens. As described in Section-\ref{subsec:positional-encoding}, Positional Encoding is the mechanism needed to keep the order in which the sequences are fed into the transformer model.

Self-Attention Mechanism is explained in Section-\ref{subsec:self-attention-mechanism}, which is used as a sub-layer by both Encoder and Decoder of the Transformer Model. As shown on the left side of the Figure-\ref{fig:transformerArchitecture}, the Encoder is stacked with multiple layers (where Nx represents the number of layers). Each layer contains a Multi-Head Self-Attention sub-layer, followed by a fully connected feed-forward network. Both sub-layers are followed by a deep residual learning \cite{residual-learning} part and normalization. As shown on the right side of the Figure-\ref{fig:transformerArchitecture}, the Decoder is a stack of multiple layers, represented by Nx. Each decoder layer has one additional multi-head attention sub-layer along with the two sub-layers used by the encoder, which computes attention over the output it receives from the encoder. Applying residual connections and normalization is done by each individual sub-layer of the decoder. The decoder takes the output sequence predicted so far as input, along with the encoded input sequence from the encoder; hence the Multi-Head Attention sub-layer of the decoder is masked to prevent attending to the outputs that are not yet predicted in the output sequence. The decoder generates Output Probabilities for the next tokens; after applying the $Softmax$ function, and the highest probability is selected as the next token predicted by the Transformer Model in the output sequence.   %encoded input sequence, currently predicted output sequence, is passed to second attention sub-layer

\subsection{Self-Attention Mechanism}\label{subsec:self-attention-mechanism}
Authors of transformer argue that using the traditional RNN's recurrence over hidden states of the long input sequence results in loss of information over the long sequences. They propose to use the attention mechanism while predicting any particular output token instead of going through the recurrence mechanism. The attention mechanism makes the computation much faster as it excludes the need for multi-step backpropagation, as in RNN. The complete input sequence is fed to the encoder, and the decoder generates the probability of the next word of the output sequence in just one iteration. Getting rid of recurrence adds more parallelism and helps utilize the parallel computation power of GPUs.

Attention essentially means that the model should be able to figure out which words in the input sequence must be paid attention to when making a particular prediction. The decoder follows an addressing scheme to decide where to look at, in the input sequence. The encoder generates the Keys $(K)$ and Values $(V)$ from the encoded input sequence. The first Multi-Head Attention sub-layer of the Decoder generates Queries $(Q)$ from the output sequence generated so far. Keys, Values, and Queries are fed into the second Multi-Head Attention sub-layer of the Decoder.

\begin{figure}
	\centering
	\includegraphics[width=1.0\linewidth]{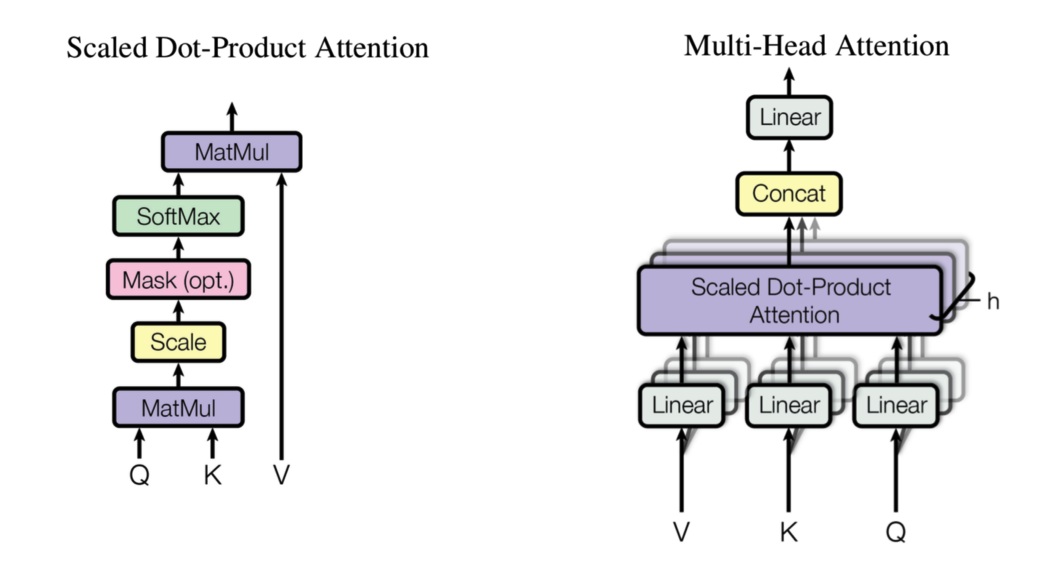}
	\caption{Transformer's Scaled Dot-Product Attention and Multi-Head Attention. From \cite{transformer}}
	\label{fig:transformerMultiHeadAttention}
\end{figure}

Figure-\ref{fig:transformerMultiHeadAttention} shows the Scaled Dot-Product Attention on the left and the Multi-Head Attention on the right side. The transformer uses one of the most commonly used attention mechanisms, i.e., Dot-Product Attention, and adds a scaling factor before taking softmax; hence the name "Scaled Dot-Product Attention". The authors selected Dot-Product due to its efficiency in time and memory consumption. Equation-\ref{equ:scaled-dot-product} \cite{transformer} shows the computation of the Scaled Dot Production Attention mechanism. A dot-product of Query (Q) is computed with all the Keys (K), both Q and K of dimensions $d_k$. A factor of $\frac{1}{\sqrt{d_k}}$ scales the result to avoid running into vanishing gradient problem when applying softmax to relatively large results from the dot product of Q and K. The result of the dot-product between Keys and Query is masked to avoid demolishing the autoregression by allowing each position to look only at the output generated so far by the decoder. The dot product yields maximum value when two vectors are fully aligned, and the dot product tends to get smaller with the increase of angle between the vectors. The dot-product yields higher results for the Key and Query vectors which are most aligned, and then softmax pushes the gap between the results from the dot products of different Keys and Query vectors, and the Key that yields the largest dot product with the Query is selected, which identifies the part which model should pay attention to when predicting next output token. The next token is selected by multiplying the result from softmax with the Values vector.

\begin{equation} \label{equ:scaled-dot-product}
\begin{split}
\text{Attention}(Q, K, V) & = \text{softmax} (\frac{Q K^T}{\sqrt{d_k}})V
\end{split}
\end{equation}

Multi-Head Attention is basically parallel Dot-Product Attentions, as shown in Figure-\ref{fig:transformerMultiHeadAttention}, where $h$ is the number of linear projections for Keys, Values, and Queries. These multiple attentions can be computed in parallel, and their results are concatenated to produce final results. Multiple attentions can learn different representations of Keys, Values, and Queries. In terms of language understanding, an attention-head can be considered as paying attention to a specific part of the language structure and semantic, e.g., personal pronouns. The modern language models use different combinations of these Multi-Head Attentions to achieve better performance on the Natural Language Processing tasks.

\subsection{Positional-Encoding} \label{subsec:positional-encoding}

Figure-\ref{fig:transformerArchitecture} shows that both encoder and decoder of the Transformer Model take Positional Encoding along with Input Embedding and Output Embedding, respectively. Since Transformer Model is purely based on the Attention Mechanism, unlike a recurrent network, there is no concept of time-step in the Transformer model; hence it cannot remember the order in which sequence was fed into the model. Positional Encoding is a mechanism to maintain the order of input since the Transformer is reading the whole sequence at-once. Relative positions of tokens in the input and output sequences are given to the encoder and decoder. Positional Encoding is the vector with information about distances between the words in the input/output sequence. After applying positional encoding on the input embedding, the word vector is generated with the positional information, i.e., context, so the positional encoding gives the notion of word context in the sentence. The Transformer \cite{transformer} uses a sin-cosine function to generate the Positional Encoding vector. The dimension of the positional encoding vector is the same as the dimension of input/output embedding, i.e., $d_{word\_embedding} = d_{positional\_encoding}$. The positional embedding vector is added into the input/output embedding, which injects the order of words without integrating into the Transformer model itself.

% write the detailed mechanism of how the positional encoding is generated

\section{BERT Model} \label{sec:bert-model}

BERT (short for Bidirectional Encoder Representations from Transformers) \cite{bert-model} is a revolutionary deep learning language model introduced in the last quarter of 2018. BERT uses representations learned in a bidirectional way by encoder part of the Transformer Neural Network, hence the name. BERT combines the strengths of the latest Language Modeling trends to produce state-of-the-art results on various Natural Language Processing tasks. It uses the recent Neural Network, i.e., Transformer \cite{transformer}, as its underlying architecture. BERT also takes advantage of Transfer Learning by Pretraining on the plain text in an unsupervised manner. The learning gained in the Pretraining process can later be used by simple Finetuning on downstream Natural Language Processing tasks. Transformer Neural Network enables BERT to learn deep bidirectional context. 
Bidirectional learning enables BERT to represent each word in a sequence according to its previous and next contexts, in contrast to unidirectional learning methods, which encode a word using only either previous or next context, depending on the direction in which they learned the context. GPT \cite{GPT-1} from OpenAI is a unidirectional Transformer Model. ELMo \cite{ELMo} concatenates LSTM trained in left-to-right and right-to-left separately, making it shallowly bidirectional. BERT, on the other hand, learns the bidirectional context simultaneously. 

Section-\ref{subsec:pre-training-bert} explains the pretraining process of BERT and how it achieved deep bidirectional context learning using the masked language modeling technique. BERT was pretrained on plain text from BookCorpus \cite{books-corpus} comprising 800 million words and text passages extracted from English Wikipedia, including 2500 million words. Pretraining is an expensive but one-time process. On cloud TPUs (Tensor Processing Units), pretraining BERT can take around 4 days. Authors released multiple variants for BERT Model, available on googleapis' bucket\footnote{\url{https://storage.googleapis.com/bert_models/2020_02_20/all_bert_models.zip}}. Pretrained BERT Models are now available for different natural languages, including a Multilingual BERT Model\footnote{\url{https://github.com/google-research/bert/blob/master/multilingual.md}}, which supports more than 100 natural languages.

\subsection{BERT's Architecture}\label{subsec:bert-architecture}

The encoder of the Transformer Neural Network explained in Section-\ref{subsec:transformer-architecture}, is used as the building block by BERT in its underlying architecture. BERT is a stack of multiple layers of Transformer Encoders. The paper \cite{bert-model} introduced two different configurations of pretrained BERT Models, i.e., \textbf{BERT$_{\textbf{BASE}}$} (L=12, H=768, A=12) and \textbf{BERT$_{\textbf{LARGE}}$} (L=24, H=1024, A=16), where L represents the number of layers, i.e., Transformer blocks, H is the size of hidden states, and A is the number of Attention Heads. The number of parameters learned by \textbf{BERT$_{\textbf{BASE}}$} and \textbf{BERT$_{\textbf{LARGE}}$} are 110M and 340M, respectively.

\begin{figure}
	\centering
	\includegraphics[width=.8\linewidth]{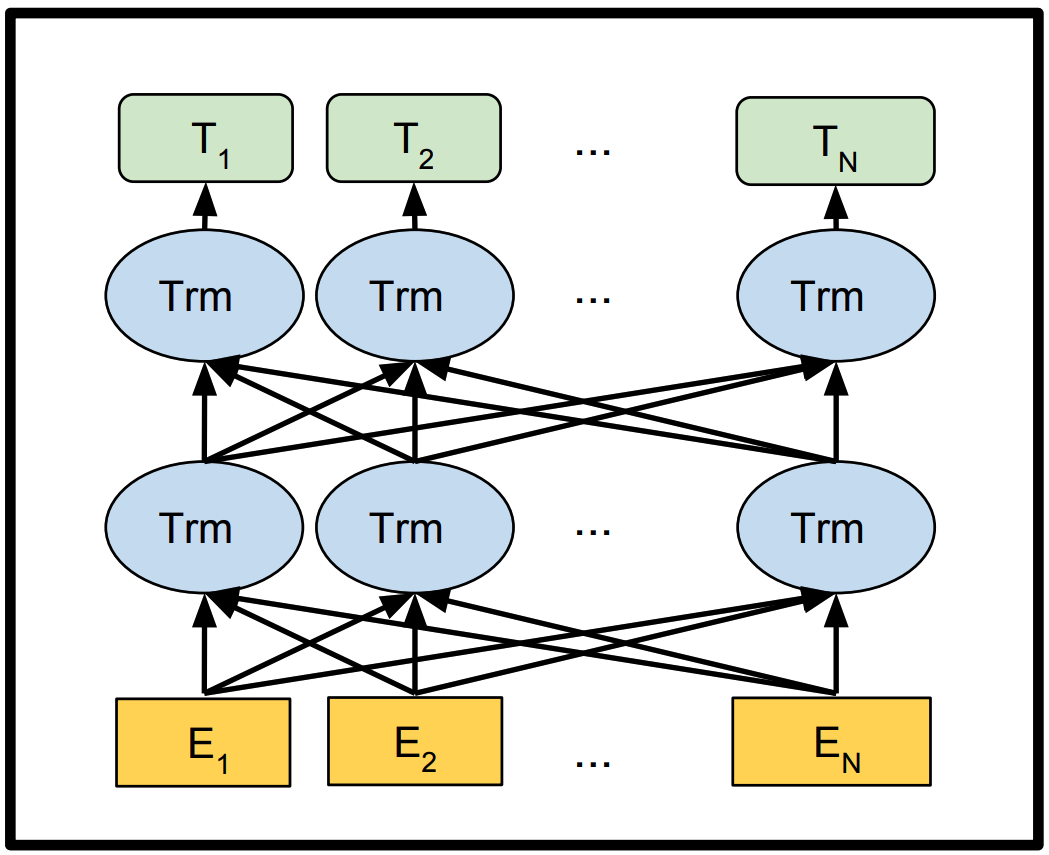}
	\caption{Pretraining architecture of BERT. From \cite{bert-model}}
	\label{fig:bert-pretraining-architecture}
\end{figure}

Figure-\ref{fig:bert-pretraining-architecture} shows the flow of information of words in the pretraining architecture of the BERT Model. Starting from the bottom layer of input embeddings ($E_1$, $E_1$, \dots, $E_N$),  information flows through all the encoder layers, with intermediate states (Trm, Trm, \dots, Trm) until the final encoded output ($T_1$, $T_1$, \dots, $T_N$). Intermediate states of the same size are generated at each layer by computing multi-headed attention on the previous layer's word representation. The direction of arrows in Figure-\ref{fig:bert-pretraining-architecture} shows that each word representation is learned by looking at its previous and next context.

\subsection{Pretraining BERT} \label{subsec:pre-training-bert}

The pretraining process is explained in Section-\ref{subsec:pretraining}. It involves unsupervised training of a Neural Network on a huge amount of plain text, with some pretraining Objectives. A Large Model with multiple Transformer Encoder layers, as explained in Section-\ref{subsec:bert-architecture}, is trained on a large corpus of plain text, using one million training steps, for a long time, and the outcome is the BERT Model. Pretraining BERT involves two main pretraining Objectives, i.e., \textbf{Masked Language Modeling} and \textbf{Next Sentence Prediction} explained in Section-\ref{subsec:masked-language-modelling} and Section-\ref{subsec:next-sentence-prediction}, respectively.

Pretraining BERT Model is a reasonably expensive task. English BERT is pretrained on the plain English text of 3.3 Billion tokens. Using the pretrained BERT Model released by the authors is suitable in most of the cases. However, further pretraining BERT with domain-specific corpora can help increase the performance in domain-specific tasks, as compared to general BERT model, which is pretrained on text from different domains.

BERT's authors provide the implementation for running Masked Language Modeling and Next Sentence Prediction tasks on arbitrary text to run pretraining on a BERT Model of our choice. Pretraining process involves two steps: 
\begin{enumerate}\bfseries
    \item Create Pretraining Data\footnote{\url{https://github.com/google-research/bert/blob/master/create_pretraining_data.py}}
    
    \textnormal{In the first step, data from plain text files are converted into TFRecord\footnote{TFRecord is a simple format from TensorFlow for storing sequence of binary records} file format for BERT.}
    \item Run Pretraining\footnote{\url{https://github.com/google-research/bert/blob/master/run_pretraining.py}}
    
    \textnormal{In the second step, BERT is pretrained on Masked Language Modeling and Next Sentence Prediction tasks using the TFRecord data generated in the first step. Refer to BERT's repository\footnote{\url{https://github.com/google-research/bert\#pre-training-with-bert}} for technical details on input parameters and pretraining recommendations from authors.}
    
\end{enumerate}

\subsection{Masked Language Modeling} \label{subsec:masked-language-modelling}
BERT takes the idea of Masked Language Modeling (MLM) from the Cloze task \cite{cloze-task}. It involves randomly masking some tokens from the input and then predict the masked words based on their context. MLM is the intuition behind the deep bidirectional context leaning of the BERT Model. The traditional language modeling is done in a unidirectional way, either left-to-right or right-to-left at a time. Masking the input helps to achieve it from both directions simultaneously without revealing the word to be predicted. BERT randomly replaces 15\% of the input tokens by \texttt{[MASK]} for prediction to learn enough context, yet avoiding overly expensive training. Given the example below, BERT aims to predict the masked word in the sentence by looking at their surrounding context.

Input Sentence: Animal abuse is a crime, and it is punishable by law.

\definecolor{light-gray}{gray}{0.95}
% \definecolor{masked-gray}{gray}{0.6}
\colorlet{NextBlue}{red!25!green!50!blue!75}

Masked Input Sentence:\colorbox{light-gray}{Animal} \colorbox{light-gray}{abuse} \colorbox{light-gray}{is} \colorbox{light-gray}{a} \colorbox{NextBlue!20}{\texttt{[MASK]}} \colorbox{light-gray}{,} \colorbox{light-gray}{and} \colorbox{light-gray}{it} \colorbox{light-gray}{is} \colorbox{light-gray}{punishable} \colorbox{light-gray}{by} \colorbox{NextBlue!20}{\texttt{[MASK]}} \colorbox{light-gray}{.}

The masked language modeling objective's problem is that masked tokens during the pretraining stage are never seen in the later finetuning stage. BERT makes the masked data biased to solve this problem. To make the data biased, when masking randomly selected 15\% input tokens, 80\% are actually replaced by \texttt{[MASK]}, 10\% are replaced with a random word, and the remaining 10\% are retained. 

\subsection{Next Sentence Prediction} \label{subsec:next-sentence-prediction}
The Next Sentence Prediction task involves pretraining sentence-pair representations to learn relationships between sentences. Given two sentences, A and B, with A appearing before B, the task is to predict if the sentence B actually follows sentence A or is it a randomly selected sentence from the Corpus. The input corpus of plain text have a defined structure, the input plain text file must have one sentence per line, and sentences must be in the correct order. Both pretraining objectives require a plain text format that is simple and easy to generate from plain text documents. Input representation of both pretraining objectives is explained in Section-\ref{subsec:berts-input-representation}.

\subsection{WordPiece Vocabulary and Tokenization}\label{subsec:wordpiece-vocab}

BERT uses WordPiece \cite{wordpiece-embedding} Vocabulary and Tokenization to generate embeddings. WordPiece is a subword segmentation algorithm, and its vocabulary contains around 30,000 tokens. A missing input word from vocabulary is greedily broken down into sub-words until all the sub-words are found in the vocabulary to handle the out-of-vocabulary problem. The authors provided the implementation\footnote{\url{https://github.com/google-research/bert/blob/master/tokenization.py}} of end-to-end tokenization and encoding to generate the input embeddings. Below is an example of Tokenization and Encoding of arbitrary legal text using BERT's WordPiece Tokenization.

Input Text: The Court had erred in overturning original decision

\definecolor{light-gray}{gray}{0.95}
Tokenized Input: \colorbox{light-gray}{the} \colorbox{light-gray}{court} \colorbox{light-gray}{had} \colorbox{light-gray}{er} \colorbox{light-gray}{\#\#red} \colorbox{light-gray}{in} \colorbox{light-gray}{over} \colorbox{light-gray}{\#\#turn} \colorbox{light-gray}{\#\#ing} \colorbox{light-gray}{original} \colorbox{light-gray}{decision}

Encoded Input: 
\colorbox{light-gray}{101} \colorbox{light-gray}{2006} \colorbox{light-gray}{13474}  \colorbox{light-gray}{1005} \colorbox{light-gray}{1055} \colorbox{light-gray}{4367} \colorbox{light-gray}{1010} \colorbox{light-gray}{1996} \colorbox{light-gray}{2212} \colorbox{light-gray}{2457} \colorbox{light-gray}{7219} \colorbox{light-gray}{1996} \colorbox{light-gray}{12087} \colorbox{light-gray}{1012} \colorbox{light-gray}{102}

where the first token in the Encoded Input, i.e., 101, represents the special token \texttt{[CLS]}, which BERT adds as the first token of a sequence. Ending Token 102 represents the special token \texttt{[SEP]}, added by BERT, to separate two sentences from a sentence-pair. These tokens are added when the input sequence is encoded with the BERT model.

% Write about tokenization and WordPiece embeddings used by Bert

\subsection{BERT's Input Representation} \label{subsec:berts-input-representation}

\begin{figure}
	\centering
	\includegraphics[width=.8\linewidth]{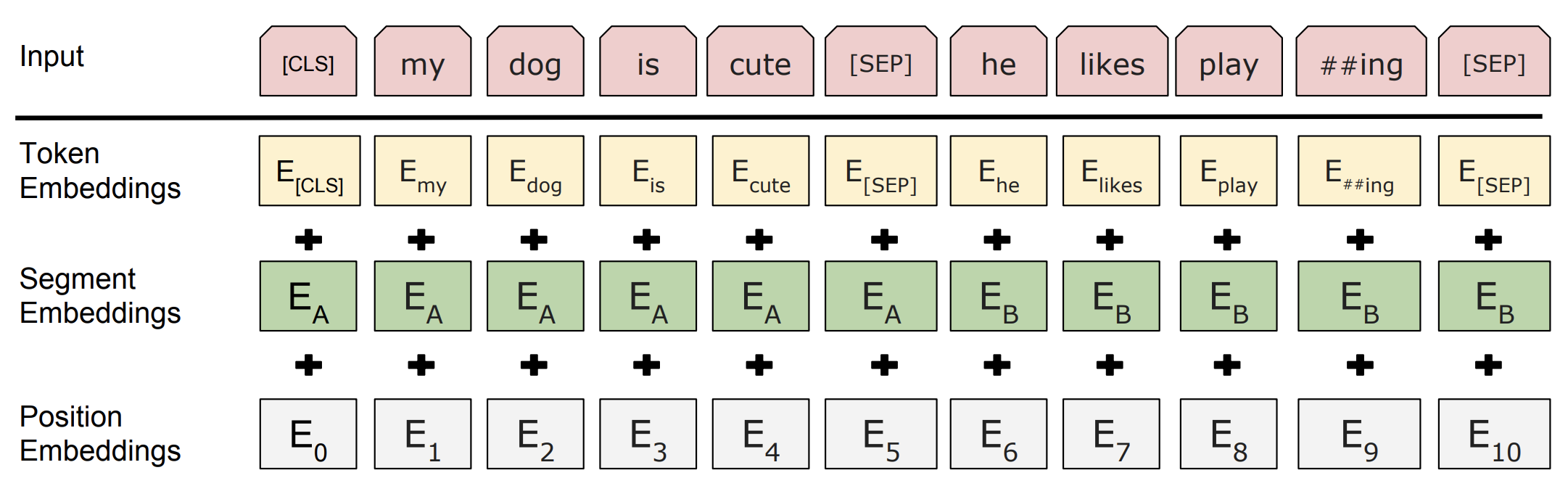}
	\caption{Input Representation of BERT. From \cite{bert-model}}
	\label{fig:bert-input-representation}
\end{figure}

BERT provides a single yet unambiguous input representation for both single sentence and sentence-pair tasks, hence easily adapting to a variety of NLP tasks. Figure-\ref{fig:bert-input-representation} shows the input representation of BERT Model. The maximum length of the input sequence can be up to 512 tokens. Token Embeddings, taken from the input text, Segment Embeddings distinguishing the sentences from sentence-pair, and Positional Encoding representing the context of words in the input sequence, add up to generate Input Embeddings. A single sentence input sequence ends with a special token \texttt{[SEP]}. Sentences are separated by the special character \texttt{[SEP]} in a sentence-pair. During the pretraining step, 15\% of the tokens are masked from the input sequence, hence the input representation satisfying the requirements of both pretraining Objectives, i.e., Masked Language Modeling, and Next Sentence Prediction. 
During the finetuning step, the input representation uses the final hidden state of the first token, i.e., \texttt{[CLS]}, for aggregate sequence representation of classification tasks. Similarly, it fits to the need for sentence-pair tasks such as Language Inference in the finetuning stage. The ingenuity of BERT's input representation design makes it portable between pretraining and finetuning stages, on various tasks, without any significant task-specific changes.

\subsection{Finetuning BERT}

The pretrained BERT model can be easily finetuned on various downstream NLP tasks. Finetuning is comparatively much less expensive and can be done in a matter of a few minutes to a couple of hours, depending on available resources and tasks being performed. Authors recommended best finetuning parameters for a variety of downstream NLP tasks, trained for 4 epochs:
\begin{itemize}
    \item batch sizes: 8, 16, 32, 64, 128
    \item learning rates: 3e-4, 1e-4, 5e-5, 3e-5
\end{itemize}

\begin{figure}
	\centering
	\includegraphics[width=.8\linewidth]{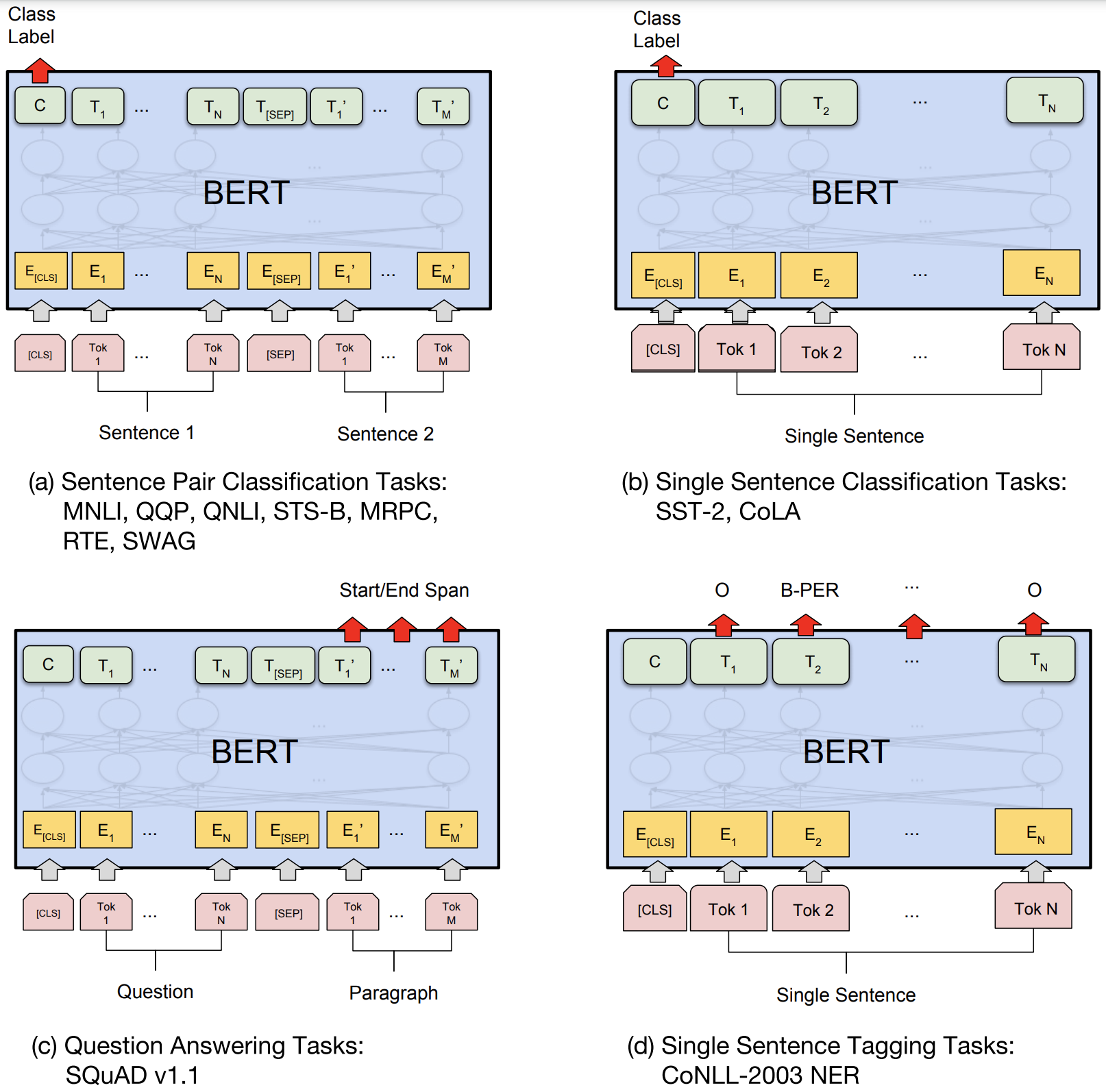}
	\caption{Finetuning BERT on different Tasks. From \cite{bert-model}}
	\label{fig:fine-tuning-bert}
\end{figure}

All the parameters are tuned in the Finetuning stage with the same learning rate selected for a task. The top-right part of Figure-\ref{fig:fine-tuning-bert} shows finetuning configurations for a sequence classification task. Input Representations, as discussed in Section-\ref{subsec:berts-input-representation}, are easily adapted to the sequence classification task. Pretrained BERT layers are trained simultaneously on the input embedding fed to give an output representation. The hidden state of the special token at the beginning of the input sequence, i.e., \texttt{[CLS]}, is fed to the classification layer as aggregated sequence representation. The classification layer is added in the finetuning stage. It has a dimension of $\mathbf{LxH}$, where L is the number of classes/labels of the sequence classification data, and H is the hidden size of the pretrained BERT model. $Softmax$ is applied at the end to compute probabilities for output labels.

For sentence-pair classification tasks, such as Natural Language Inference, the input representation is quite similar. The only change compared to the single sentence classification task is that two sentences are packed together in the input sequence, using a special token, i.e., \texttt{[SEP]}, to separate both sentences, as shown in the top-left part of Figure-\ref{fig:fine-tuning-bert}. Special token \texttt{[CLS]} is used for feeding aggregated sequence representation to the classification layer, similar to the single sentence classification task.

The finetuning setting for a token classification task such as Named Entity Recognition is shown on the bottom-right part of Figure-\ref{fig:fine-tuning-bert}. In token classification, every token's final hidden output is fed to the classification layer instead of just the first token, i.e., \texttt{[CLS]} as in the case of the sequence classification task, and a label is predicted for every token. For tokens, broken down by WordPiece tokenizer into multiple parts, the prediction is made only for the first part.

The bottom-left part of the Figure-\ref{fig:fine-tuning-bert} shows the finetuning setting for the Question-Answering task. In the Question-Answering task, a question and a context paragraph containing the answer to the question is given. The model predicts the beginning and ending tokens from the paragraph that possibly mark the span of the answer in the paragraph. Like the sentence-pair classification task, the Question and the context paragraph are packed together in the input sequence, with a special token, i.e., \texttt{[SEP]} separating both from each other. Two new parameters, i.e., "start vector" and "end vector", are learned in this task during the finetuning. Final probabilities for starting and ending tokens are computed by taking softmax over each output token's product with "start vector" and "end vector", respectively.

 \section{XLNet Model} \label{sec:xlnet-model}

XLNet \cite{xlnet-model} is an autoregressive pretraining method that enables bidirectional context learning using Permutation Language Modeling as a pretraining Objective. Like BERT \cite{bert-model}, XLNet also uses Transformer \cite{transformer} as its underlying Neural Network for its training. Modern Neural Language Models use pretraining in their first stage to train themselves on a large amount of plain text of a natural language, in an unsupervised way. Two of the popular pretraining objectives used are autoencoding and autoregressive (AR) language modeling. BERT uses autoencoding-based pretraining, whereas XLNet uses autoregressive language modeling. Authors of XLNet argue that using autoencoding-based pretraining, BERT corrupts its input by masking\footnote{Masked Language Modeling, as explained in Section-\ref{subsec:masked-language-modelling}} it during the pretraining stage. Special tokens, masking the words from input in the pretraining stage, are not part of real data during the finetuning stage on the downstream NLP tasks, making both stages inconsistent, resulting in the pretrain-finetune discrepancy problem. 

The two most successful pretraining objectives, i.e., autoencoding and autoregressive language modeling, have their own pro and cons. BERT is based on denoising autoencoding, which maximizes the likelihood of a masked token $x_t$ from input sequence x = $[x_0, x_1, \dots, x_t, \dots, x_T]$ based on all the tokens from input sequence x except $x_t$ itself. Autoencoding (AE) enables BERT learning bidirectional context, but input masking leads to the pretrain-finetune discrepancy. Also, it assumes independence between masked inputs, which does not always hold in natural languages. Autoregressive (AR) language modeling aims to predict a token $x_t$ based on the tokens that appear before it, i.e., $[x_0, x_1, \dots, x_{t-1}]$ or after it, i.e., $[x_{t+1}, x_{t+2}, \dots, x_T]$, depending on the direction. AR language Modeling does not rely on corrupted input, but it can only encode unidirectional context, in either forward or backward direction, lacking simultaneous bidirectional context learning. XLNet exploits the strength of both these objectives while avoiding their limitations using its ingenious Permutation Language Modeling pretraining objective. Using Permutation Language Modeling, XLNet overcame the gap between autoregressive language modeling and effective bidirectional pretraining. By combining the benefits of both types of objectives and avoiding their drawbacks, XLNet became the new state-of-the-art Language Model, outperforming BERT on 20 Natural Language Processing tasks.

\subsection{Transformer-XL}\label{subsec:transformer-xl}

The novel Transformer Neural Network Model \cite{transformer}, has a lot of potential, and helps to learn long term dependencies, enabling simultaneous bidirectional context learning. However, it has a certain limitation on the length of the input sequence. In the beginning of the year 2019, Transformer-XL \cite{transformer-xl} overcame this limitation and enabled the attention mechanism beyond the fixed-length context.

Transformer-XL can learn long-term dependencies that are 80\% longer than Recurrent Neural Networks and 450\% longer than the Transformer.  It achieves better performance on both long and short sequences in both evaluation measures and computation time. Transformer-XL achieves this using a new positional encoding scheme combined with a segment-level recurrence mechanism. State-of-the-art Language Model XLNet incorporates the ideas from Transformer-XL and is named after it.

\subsection{XLNet's Architecture}\label{subsec:xlnet-architecture}

The major contribution from XLNet is not its architecture, rather a superior pretraining method that adapts the benefits of different pretraining objectives. It borrows the ideas of segment recurrence mechanism and the novel relative encoding scheme proposed by Transformer-XL \cite{transformer-xl}, as explained in Section-\ref{subsec:transformer-xl}. Underlying XLNet architecture is based on multiple Transformer layers, with recurrence. Multiple variants of different sizes and configurations of pretrained XLNet Models are released. The pretrained XLNet Models include \textbf{XLNet$_{\textbf{BASE}}$} (L=12, H=768, A=12) and \textbf{XLNet$_{\textbf{LARGE}}$} (L=24, H=1024, A=16) where L represents the number of layers, i.e., Transformer blocks, H is the size of hidden states, and A is the number of Attention Heads. The number of parameters learned by \textbf{XLNet$_{\textbf{BASE}}$} and \textbf{XLNet$_{\textbf{LARGE}}$} are $\sim$110M and $\sim$340M, respectively. The size and configurations of both XLNet Models are comparable to BERT, although \textbf{XLNet$_{\textbf{LARGE}}$} is pretrained on much larger pretraining data than \textbf{BERT$_{\textbf{LARGE}}$}. However, to give a fair comparison, authors of XLNet also released another variant of \textbf{XLNet$_{\textbf{LARGE}}$}, namely \textbf{XLNet-Large-wikibooks}, which is pretrained on the same data, i.e., WikiPedia+BooksCorpus, like \textbf{BERT$_{\textbf{LARGE}}$}.

%first write about the building block and size and configuration and different available xlnet configurations

\subsection{Pretraining XLNet} \label{subsec:pretraining-xlnet} 
% and objectives, and pre-training statistics
% write about pre-training statistics, and time.

Pretraining data for XLNet includes BooksCorpus \cite{books-corpus} and text from English Wikipedia, on which BERT was pretrained. In addition, it also includes Giga5 \cite{giga5}, ClueWeb 2012-B (an extension of \cite{callan2009clueweb09}), and Common Crawl \cite{common-crawl}. XLNet uses SentencePiece \cite{sentence-piece} tokenization on the input pretraining data after excluding the low-quality documents from raw corpora. The total number of tokens (after applying SentencePiece tokenization) included in pretraining data for XLNet are 32.89B, including Wikipedia (2.78B), BooksCorpus (1.09B), Giga5 (4.75B), ClueWeb (4.30B), and Common Crawl (19.97B). Largest XLNet Model, i.e., \textbf{XLNet$_{\textbf{LARGE}}$}, reuses all pretraining hyper-parameters as in \textbf{BERT$_{\textbf{LARGE}}$} and pretrains on Wikipedia+BookCorpus, making it equal in size and configuration. This model is further augmented by pretraining on all the remaining pretraining datasets. Pretraining XLNet Model took five and a half days on 512 TPU v3 chips for 500K training steps, using sequence length as 512 tokens and batch size as 8192. The finetuning process for XLNet is fairly similar to that of BERT. 

\subsection{Permutation Language Modeling} \label{subsec:permutation-lm}

\begin{figure}
	\centering
	\includegraphics[width=.8\linewidth]{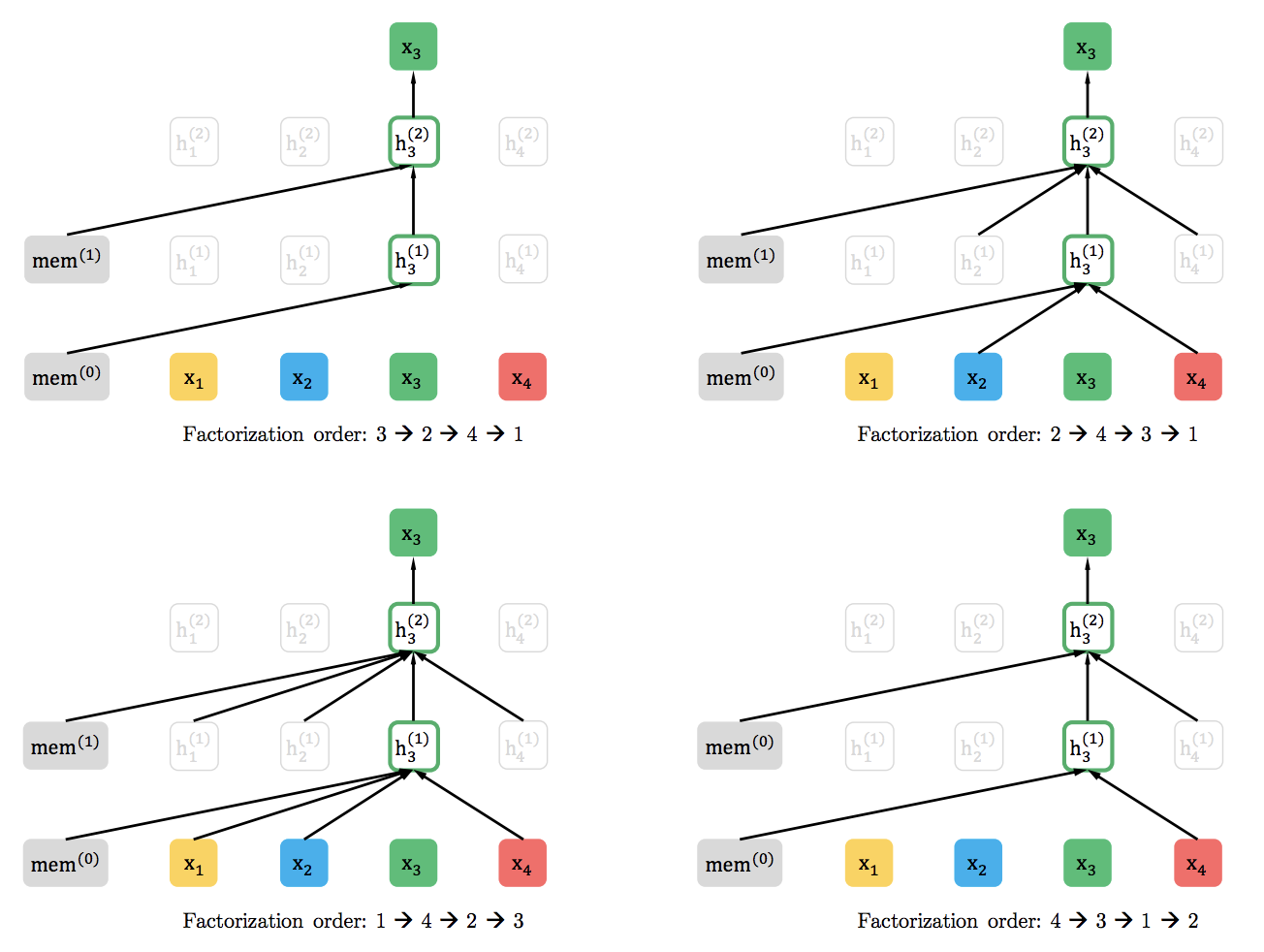}
	\caption{Permutation Language Modeling for predicting $x_3$ from the different permutations of the same input sequence. From \cite{xlnet-model}}
	\label{fig:xlnet-permutation-lm}
\end{figure}

Traditional Autoregressive Language Modeling cannot learn simultaneous bidirectional context. To use the advantages of AR Language Modeling, along with the deep bidirectional context learning, XLNet \cite{xlnet-model} uses the novel idea of Permutation Language Modeling, inspired by NADE \cite{neural-ar-dist-est-permutatin-lm}, as its pretraining Objective. XLNet aims to improve pretraining using Permutation LM to predict all tokens in random order. The basic idea behind Permutation LM is Permutations. Given a sequence x containing T number of tokens, there are T! possible factorization orders, e.g., x =  $[x_1, x_2, x_3]$, then 3! i.e., 6 possible permutations for the sequence are \{$[x_1, x_2, x_3]$, $[x_1, x_3, x_2]$, $[x_2, x_1, x_3]$, $[x_2, x_3, x_1]$, $[x_3, x_1, x_2]$, $[x_3, x_2, x_1]$\}. An illustration of Permutation Language Modeling is shown in Figure-\ref{fig:xlnet-permutation-lm}. XLNet shares the parameters across all the permuted factorization orders to learn the context from all positions for a word in the sequence. It aims to maximize the log-likelihood of a sequence from the possible permutations of the factorization order, to capture the context information from both directions simultaneously. The mathematical formulation given by authors for the Permutation Language Modeling is shown in Equation-\ref{eqn:permutation-lm} \cite{xlnet-model}.

\begin{equation} \label{eqn:permutation-lm}
\underset{\theta}{\mathrm{max}} \ \ \  \mathbb{E}_{\mathbf{z}\sim \mathcal{Z}_T} \Bigg[\sum_{t=1}^T \log p_\theta (x_{z_t}\ |\ \mathbf{x}_{\mathbf{z}_{<t}}) \Bigg].
\end{equation}

where $\mathcal{Z}_T$ is a set of all possible permutations of the sequence of length T, $x_{z_t}$ is the $t^{th}$ element, and $\mathbf{x}_{\mathbf{z}_{<t}}$ represents the tokens before it, in the permutation $\mathbf{z}\in\mathcal{Z}_T$ of the input sequence.

% % https://www.kdnuggets.com/2019/09/bert-roberta-distilbert-xlnet-one-use.html
% % To improve the training, XLNet introduces permutation language modeling, where all tokens are predicted but in random order.

% % https://cloud4scieng.org/modeling-natural-language-with-transformers-bert-roberta-and-xlnet/

% % https://medium.com/@zxiao2015/understanding-language-using-xlnet-with-autoregressive-pre-training-9c86e5bea443
% % How does XLNet differ from conventional AR and AE (BERT)

\subsection{Two-Stream Self-Attention}\label{two-stream-self-attention}
% %optional. write about it is it seems vital to the understanding of XLNet model.
XLNet reparameterizes the Transformer-XL architecture to use the Permutation Language Modeling objective. This is necessary in order to make the objective work with traditional Transformer/Transformer-XL architecture, to remove the target ambiguity. The Transformer-XL \cite{transformer-xl} looks at the entire input sequence simultaneously, and XLNet does not mask any tokens like BERT \cite{bert-model}. Hence it modifies the Transformer-XL to only look at the hidden representation of the tokens preceding the target to be predicted and the positional information of the target itself. To maintain the information about the order of input sequence, positional information is embedded using relative encoding. 

\begin{figure}
	\centering
	\includegraphics[width=.8\linewidth]{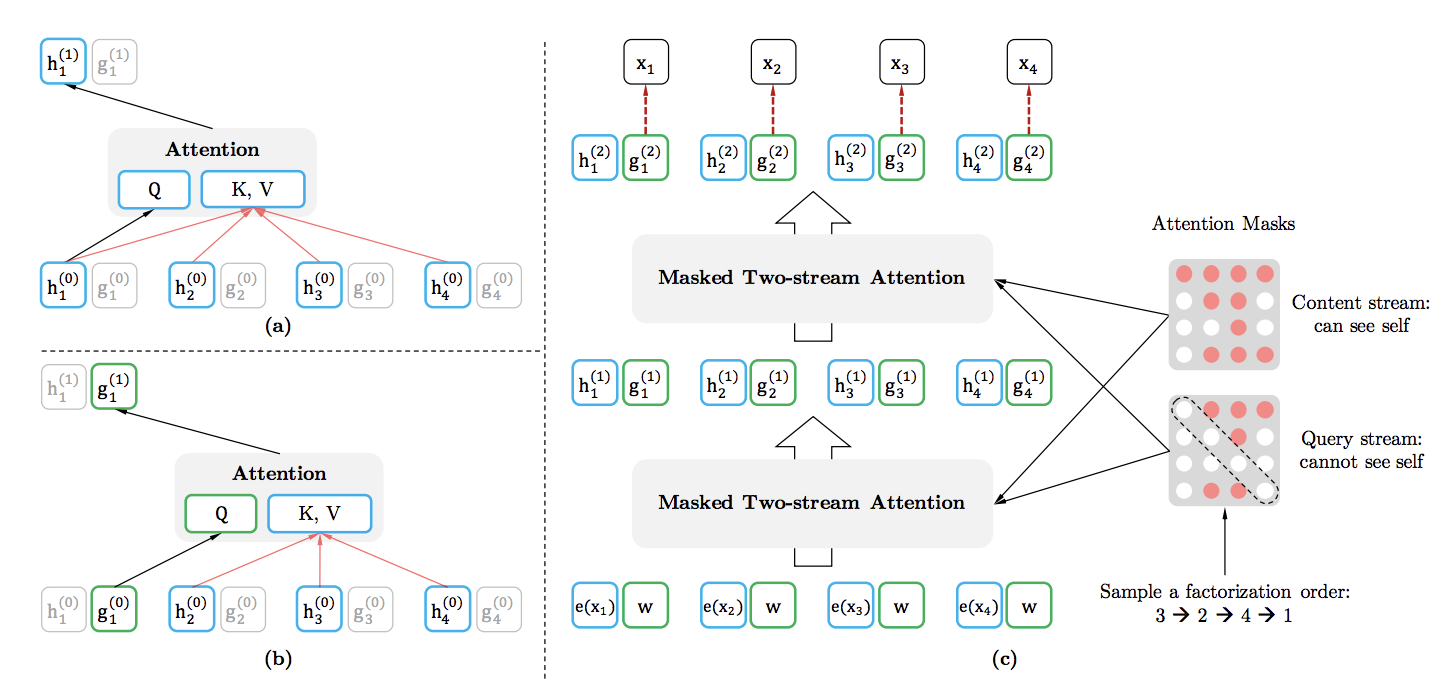}
	\caption{Two-Stream Self-Attention. From \cite{xlnet-model}}
	\label{fig:2-stream-self-attention}
\end{figure}
Two-Stream Self-Attention involves two types of attention. The first one, content stream representation, is the same as the standard self-attention in the Transformer architecture. It examines both content and the positional information of the token to be predicted. The second attention, query stream representation, replaces the input masking, i.e., \texttt{[MASK]}, as explained in Section-\ref{subsec:masked-language-modelling}. It learns to predict the target token using context and the target's positional information, but not the content of the target. If token $x_i$ is to be predicted, the subsequent layers can only access its positional information and the content of tokens preceding it, but not the content of $x_i$ itself. The attention is computed by changing the Query (Q), Key (K), and Value (V) vectors according to this modification. Figure-\ref{fig:2-stream-self-attention} shows the content stream and query stream self-attentions. 
%this figure can be removed if required

% \input{Master Thesis - Report/chapters/related-work.tex}
\chapter{Related Work}\label{chap:related-work}
% write about at least 3 or 4 related research studies and compare them to our own

\section{Evaluation of Deep Neural Language Models on Hate Speech Detection}

Capturing long-term dependencies and a bidirectional context in the input sequence are key factors for improvements in deep neural language modeling. In their paper \cite{hate-speech-sameval-task6} on detecting hate speech using Deep Learning Language Models, authors combined a Convolutional Neural Network (CNN) and bidirectional GRU (2 concatenated GRUs, trained in each direction, i.e., left-to-right and right-to-left) to build the proposed C-BiGRU Model, capturing long-term dependencies to distinguish offensive tweets from non-offensive tweets. C-BiGRU was used in another paper \cite{hate-speech-sameval-task12} participating in SemEval 2020 workshop, taking part in English, Turkish and Danish competitions, achieving F1 scores of 90.88\%, 76.76\%, and 76.70\%, respectively.

\section{Legal Area Classification} \label{legal-area-classification}
In a comparative study \cite{legal-area-classification-singapore-supreme-court} of different Text Classifiers on Judgments from the Supreme Court of Singapore, authors applied different machine learning approaches to classify the judgments in different legal areas. It gives a comparison between state-of-the-art Neural Language Models and Traditional Statistical Models. Their experiments tested different models, including Topic Models, Word Embedding Feature Models, and Pretrained Language Models. The authors concluded that while data limitation affects all classifiers, the state-of-the-art Neural Language Models like BERT are much less sensitive to data scarcity and can perform well on as few legal judgments as 588.

The study compares different types of models on legal judgment classification. In contrast to their work, we examine two different Neural Language Models in the Legal Domain, i.e., BERT \cite{bert-model}, and XLNet \cite{xlnet-model}, pretrained with different Pretraining Objectives, i.e., autoencoding and regressive language modeling, respectively. At the time of writing this paper, there is no comparison between these models already available on legal judgment classification.

\section{German Legal Decision Corpora}\label{subsec:german-law-corpora}

A recent paper on Design and Implementation of German Legal Decision Corpora \cite{icaart21} presents two German legal text corpora and experiments on these corpora. One of the corpora consists of German courts' decisions, whereas the second one is a subset of the first corpus, with the conclusion, definition, and subsumption components of German legal writing style, annotated by a domain expert. Based on the first corpus, the paper provides an experiment on the classification of decision type. The corpus contains decisions of various types, including types of resolutions, and judgments, etc., with imbalanced decision classes. To create balance, authors categorized all types of resolutions into a class $resolution$, and all types of judgments into a class $judgment$, and the remaining are labeled as $others$. They illustrated the corpus's usefulness by conducting a comparison between Logistic Regression (LR) and Support Vector Classification (SVC), in which SVC slightly outperformed LR. The second corpus is labeled with the components of Urteilsstil (judgment style). The paper shows results from experiments performed using LR and SVC, with unigram and tf-idf features. The goal is to train a model on Urteilsstil (judgment style) and use it on Gutachtenstil (appraisal style) to prove that the model can withstand a domain transfer, for detecting the common components between the two legal writing styles. All the combinations of classifiers and features outperformed the baseline, which is created by a decision stump that always predicts the majority class, showing that meaningful features are selected for domain transfer.

In contrast, to conduct experiments on Natural Language Processing in the Legal domain, we evaluate Deep Neural Language Models using US courts' decisions.

There has been a lot of research done on performance evaluation of BERT Model \cite{bert-model} on domain-specific corpora, including BioBERT Model \cite{LM-NLP:BioBERT} pretrained in the biomedical domain, SciBERT Model \cite{LM-NLP:SciBERT} pretrained on scientific datasets, and LEGAL-BERT Model \cite{muppets-legal-bert} pretrained on Legal Data. Besides the domain-specific adaptation, BERT is also retrained by caselli et al. to create a language variety-oriented model, i.e., HateBERT \cite{caselli2021hatebert}.

\section{HateBERT}\label{subsec:hatebert}

In their paper on Retraining BERT for Abusive Language Detection in English, authors released HateBERT \cite{caselli2021hatebert}, an abuse-inclined version of the BERT model. HateBERT is prepared by further pretraining the BERT\textsubscript{$\mathrm{base}$-$\mathrm{uncased}$} model on a total of 43,379,350 tokens from the RAL-E (Reddit Abusive Language English) dataset. The RAL-E dataset is also part of the contribution, which is prepared by extracting a publicly available Reddit Comments collection from communities banned for hosting hateful, offensive, and abusive content. Authors performed experiments with multiple datasets, including OffensEval 2019 \cite{sameval-2019}, AbusEval \cite{caselli-etal-2020-feel}, and HatEval \cite{basile-etal-2019-semeval}. HateBERT outperforms the general BERT model on all the datasets and is more robust in improvements on negative and positive classes from hate speech data. Authors conclude that further pretraining is an effective technique to adapt BERT to other language varieties. HateBERT obtains robust representations for the offensive phenomenon against which it is finetuned. It adapts better to the language variety and polarity, as compared to its corresponding general BERT model. 

In comparison, instead of language variety, we choose the domain of Law for evaluation of domain adaptation abilities of the BERT Model. 

\section{BioBERT}\label{subsec:rw-biobert}

To develop BioBERT \cite{LM-NLP:BioBERT}, authors used data from archives of biomedical and life science literature, comprising a Corpus 18B words in total, including 4.5B words from PubMed\footnote{\url{https://catalog.data.gov/dataset/pubmed}} and 13.5B words Corpus from PMC\footnote{\url{https://www.ncbi.nlm.nih.gov/pmc/}}. BioBERT is an outcome of their investigation on how the BERT model can be adapted to the medical domain. The authors built the model BioBERT by initializing it with the weights learned from pretrained BERT\textsubscript{$\mathrm{BASE}$} Model. Authors released a model trained on PubMed and PMC corpora, using 470K training steps, which until this release was their best performing model. Later, they released another model, which was trained only on PubMed corpus using 1M training steps, which performed better than their first model. Authors found that BioBERT largely outperforms BERT in biomedical text mining tasks. 

In comparison to their work in the biomedical domain, we propose a Legal-BERT Model specializing in the Legal Domain. In addition to pretraining a general pretrained BERT model on legal corpora, we also explore the benefits of adding Legal vocabulary in the general WordPiece vocabulary used by BERT, comprising around 30K tokens. We initialized our Legal-BERT Model with pretrained BERT\textsubscript{$\mathrm{MEDIUM}$} Model and pretrained it further for 1M training steps on legal data. We compare our Legal-BERT Model's performance with the general BERT Model in the Legal Domain to analyze the effects of Legal Domain adaptation.

\section{SciBERT}

SciBERT \cite{LM-NLP:SciBERT} is a pretrained BERT Model on a large amount of scientific text. It is pretrained from scratch, using the same configuration and size as BERT\textsubscript{$\mathrm{BASE}$} instead of initializing it with weights of an existing pretrained BERT Model. The pretraining data is taken from Semantic Scholar Corpus\footnote{\url{https://www.semanticscholar.org/}}, which includes Biomedical (82\%) and Computer Science (18\%) papers building a corpus of 3.17B tokens. Along with the SciBERT Model, authors also constructed a $\mathrm{SCIVOCAB}$, i.e., the WordPiece vocabulary generated from their scientific corpus. Keeping the size of $\mathrm{SCIVOCAB}$ the same as BERT\textsubscript{$\mathrm{BASE}$}, authors compared the evaluation of SciBERT and BERT\textsubscript{$\mathrm{BASE}$}. They concluded that SciBERT significantly outperformed BERT\textsubscript{$\mathrm{BASE}$} in the scientific domain. They performed several NLP tasks, ranging in Computer Science, Biomedical, and multidomain tasks, where SciBERT achieved state-of-the-art results in 8 out of 12 tasks.

In comparison to SciBERT, our Legal-BERT Model, as discussed in Section-\ref{subsec:preparing-legal-bert}, specializes in Legal Domain. There are two major differences besides the domain expertise between SciBERT and Legal-BERT. The first difference is the BERT Model's size, i.e., SciBERT has the size and configurations of BERT\textsubscript{$\mathrm{BASE}$}, whereas our Legal-BERT has BERT\textsubscript{$\mathrm{MEDIUM}$}. To analyze the effects on performance, Legal-BERT is compared with pretrained general BERT\textsubscript{$\mathrm{MEDIUM}$} as released by the BERT's authors. The second difference is that SciBERT was pretrained from scratch, whereas our Legal-BERT is initialized with BERT\textsubscript{$\mathrm{MEDIUM}$} and then further pretrained on Legal text.

The difference between SCIVOCAB and our work in analyzing the effects of Legal Vocabulary on the BERT Model is that SCIVOCAB was constructed from the scientific corpus, overlapping original WordPiece Vocabulary\footnote{the original WordPiece vocabulary that comes with pretrained BERT Model, constructed from general multi-domain English text} by 42\%. In contrast, we add new Vocabulary in the original WordPiece vocabulary, which is specific to the Legal Domain, i.e., legal terminologies. The newly added Legal Vocabulary is not broken down into multiple parts, contrary to WordPiece Vocabulary created using SentencePiece\footnote{\url{https://github.com/google/sentencepiece}} library. This increases the size of the original WordPiece vocabulary, and we analyze the effects on performance in evaluation measures and computation time on legal tasks.

% https://www.researchgate.net/profile/Jelena_Mitrovic9

%https://www.researchgate.net/publication/327285631_Automated_Legal_Research_for_German_Law

% https://www.researchgate.net/publication/345395991_nlpUP_at_SemEval-2019_Task_6_A_Deep_Neural_Language_Model_for_Offensive_Language_Detection

% https://www.researchgate.net/publication/332057279_nlpUP_at_SemEval-2019_Task_6_A_Deep_Neural_Language_Model_for_Offensive_Language_Detection

% https://www.researchgate.net/publication/343558319_NLP_Passau_at_SemEval-2020_Task_12_Multilingual_Neural_Network_for_Offensive_Language_Detection_in_English_Danish_and_Turkish

% https://www.researchgate.net/publication/344878570_HateBERT_Retraining_BERT_for_Abusive_Language_Detection_in_English

%  (https://www.researchgate.net/publication/327285631_Automated_Legal_Research_for_German_Law), and then proceed with this heading

\section{LEGAL-BERT: The Muppets straight out of Law School}\label{sec:muppets-legal-bert}

In the last quarter of 2020, a LEGAL-BERT Model was released by researchers in their paper \cite{muppets-legal-bert} during the writing of our paper. The authors developed a family of Legal BERT Models. They prepared a pretrained BERT Model from scratch on Legal text (named LEGAL-BERT-SC), with configurations and size of the BERT\textsubscript{$\mathrm{BASE}$} model. The authors initialized the second Legal BERT Model (named LEGAL-BERT-FP) with pretrained BERT\textsubscript{$\mathrm{BASE}$} and added further pretraining on legal text for 1M training steps. Authors also released a Legal BERT Model of smaller configuration and size, i.e., LEGAL-BERT-SMALL with 6 layers, 512 hidden units, and 8 attention heads. Authors collected legal text of 12GB from multiple legal sub-domains for pretraining Legal BERT Models. Pretraining data collected contained legal documents from EU Legislation (16.5\%), UK Legislation (12.2\%), European Court of Justice (ECJ) cases (5.2\%), European Court of Human Rights (ECHR) cases (4.3\%), US court cases, from Case Law Project (27.8\%), and US Contracts (34.0\%). Within the Further Pretrained Legal BERT Models, i.e., LEGAL-BERT-FP, authors developed legal models specializing in legal sub-fields, e.g., US Contracts, ECJ, ECHR. The authors concluded that a Legal variant of BERT always outperformed pretrained BERT\textsubscript{$\mathrm{BASE}$} on all legal datasets. The LEGAL-BERT-SMALL compares well to Legal variants of BERT Models, despite being small in size. Within the LEGAL-BERT-FP family, sub-domain specific models adapt faster in sub-domain tasks than a variant pretrained on the complete collection of legal corpora. 

The fundamental differences between Legal BERT released by authors of the paper "LEGAL-BERT: The Muppets straight out of Law School" \cite{muppets-legal-bert}, and our Legal-BERT Model, are the size and configurations, and the pretraining data. Our Model is pretrained on Legal text from both publicly accessible as well as Research Scholar's data of US Cases from the Case Law Project, whereas they used legal text from various subfields within the Legal Domain from publicly accessible resources. Our Legal Model has the size and configurations of a relatively smaller BERT model, i.e., BERT\textsubscript{$\mathrm{MEDIUM}$}, and is pretrained on Legal text from Case Law Project for 1M training steps. Besides the pretraining strategy for adapting BERT to the Legal Domain, we also consider exploiting the tokenization and encoding input by introducing legal terminologies in WordPiece Vocabulary.

Authors of the paper "LEGAL-BERT: The Muppets straight out of Law School" released their Legal BERT Model\footnote{\url{https://huggingface.co/nlpaueb/legal-bert-base-uncased}} that is pretrained on legal text from various legal sub-domains. We evaluated their model on legal tasks and compared its performance with our Legal-BERT Models. Comparison is given in the Chapter-\ref{chap:results}.

% \subsection{Effects of Legal Vocabulary on BERT fo German Legal Language}

%write one related work involving xlnet. find a study on xlnet in law domain.
% \afterpage{\blankpage}
\chapter{Methods}\label{chap:methods}

%our source code: use the variables below. as defined in thesis.tex. change once to reflect everywhere.

% \SourceCodeRepo # for using the url to source code repository
% \RepoRootPath # for the next path to the root directory of the repository's code.
% \GenPreTrainData # for path to the generate_pretraining_data script.

% Describe the method/software/tool/algorithm you have developed here

\section{Datasets} \label{sec:datasets}

% Datasource and statistics of the data taken. recognition and credits to "case.law"
Legal Datasets used in the pretraining and evaluation process of the Transformer-based Models are taken from the CASELAW Access Project\footnote{\url{https://case.law/}} of Harvard Law School. The Legal Data is composed of the Judgments given by the State and Federal Jurisdictions of the Unites-States, on Legal Cases. The CASELAW Access Project's publicly available Legal data is from four Jurisdictions of the states of Arkansas, Illinois, New Mexico, and North Carolina, comprising 358,819 Legal Cases. We acquired further Legal data after applying for access to the researcher's data\footnote{\url{https://case.law/bulk/download/\#researcher}}. Researcher Data includes the Judgements for Legal Cases from all the remaining states and federal Jurisdictions of the United-States. The total Legal Data contains 6,725,065 unique Legal Cases, including Public and Researchers Data. Datasets for Legal Cases are available in \texttt{XML} and \texttt{JSON} formats. We acquired the JSON format as it fits well with our simple data manipulation and usage requirements. An example of a Legal Case from  CASELAW project is given below:

\begin{lstlisting}[language=json,firstnumber=1]
{
  "id": 18630,
  "decision_date": "1997-12-17",
  "citations": [ { "cite": "124 N.M. 498" } ],
  "jurisdiction": {
    "name_long": "New Mexico",
    "id": 52
  },
  "casebody": {
    "data": {
      "judges": [ "APODACA and ARMIJO, JJ., concur." ],
      "attorneys": [
        "Maria Garcia Geer, Geer, Wissel & Levy, P.A., Albuquerque, for plaintiff-appellant.",
        "Michael E. Knight, Knight & Nagel, P.A., Albuquerque, for defendant-appellee."
      ],
      "opinions": [
        {
          "type": "majority",
          "text": "OPINION\nPICKARD, Judge. This case requires us to consider the appropriate relief when a seller wishes to declare forfeiture of a large down payment upon default of a real estate contract. Caye Buckingham (Buyer) sued James Ryan (Seller) for damages resulting from .......",
          "author": "PICKARD, Judge."
        }
      ]
    }
  }
}
\end{lstlisting}

The Opinion text shown in the example above is truncated to save space; in reality, the Legal Cases' Judgments are much longer, comprising multiple pages. The data for the Legal Cases also includes other information, but the example shows only the relevant fields for sufficient understanding of the Legal Data used.

\subsection{Pretraining Dataset}\label{subsec:pre-training-dataset}

The Pretraining process and its benefits are discussed in Section-\ref{subsec:pretraining}. BERT is pretrained on a large amount of English text from different domains. The details on the pretraining process for BERT are discussed in Section-\ref{subsec:pre-training-bert}. The pretraining Objectives used by BERT, i.e., Masked Language Modeling and Next Sentence Prediction, are discussed in Section-\ref{subsec:masked-language-modelling} and Section-\ref{subsec:next-sentence-prediction}, respectively. Pretraining BERT requires plain unlabelled text in a specific format. The input file with unlabelled plain text for pretraining must have one sentence per line, and sentences must be in sequence as they appear in the document. A blank line in the input text file indicates the end of a document and the beginning of a new one. 

The Legal Data acquired as described in Section-\ref{sec:datasets} contains Opinions from Judges on the Legal Cases from all the state and federal Jurisdictions of the United-States. The text from these Legal Opinions is parsed to generate the pretraining unlabelled data, adherent to BERT's pretraining format requirements. The text from legal opinions is cleaned and pre-processed before splitting it into sentences. NLTK\footnote{\url{https://www.nltk.org/}} is used for sentence segmentation to create sequential sentences from the much longer legal opinions. To avoid raw, low-quality text, each sentence is then filtered by a set of intuitive rules. Short and low-quality sentences are excluded from the output pretraining data. As per the pretraining text format, a blank line is used as a delimiter to separate multiple documents, as the next sentence prediction objective requires the sentences to follow the actual order in the data. A new document starts when either a Legal Opinion ends, and the next one starts, or one or more sentences get discarded due to the text's low quality. To ensure the quality of documents in the pretraining data, the documents containing fewer than ten sentences are discarded. We provide a simple script\footnote{\url{\SourceCodeRepo\RepoRootPath\GenPreTrainData}} for generating pretraining-format adherent data in plain text files, using the Legal Data in \texttt{JSON} Format acquired from CASELAW Project\footnote{\url{https://case.law/bulk/download/}}. It provides easy pagination on input legal data files and file rotation on output files to generate the pretraining data in an optimal way. The total pretraining data generated from the Legal Cases data from the CASELAW Project constitute about 200 text files, with each file growing on average around 43MB in size.

%write about the statistics now. discuss the optimization strategy, just briefly, to facilitate the pretraining process, describe the optimization in more detail in the preparation of Legal-BERT.

%The process of preparing the pretraining data. Write about the structure of the pretraining data. follow the script for pre-pairing the pretraining data to include the details.

%write statistics of time and amount of data. and the optimization reasons behind the way data is split.

\subsection{Legal Opinions Classification Dataset}\label{subsec:classification-dataset}

%TODO, move the data example to index?
An example of Legal Case Data from CASELAW Access Project is shown in Section-\ref{sec:datasets}. It shows opinion from the Judge on the Legal Cases, classified as type \texttt{majority}. Similarly, a Judge can dissent from the majority decision of the Court, and in this case, the Opinion is marked as \texttt{dissent}. The Opinions in the Legal Cases Data are quite long, often up to multiple pages. However, the limitations of input sequence length for BERT restricts using such long opinions. To solve the sequence length limitation problem, the legal opinions are first summarized using the gensim\footnote{\url{https://radimrehurek.com/gensim/}} summarizer. The gensim summarizer is based on the popular TextRank algorithm and is well suited for summarizing long text into important sentences. TextRank is similar to the famous PageRank algorithm, except that it considers text instead of pages.

\begin{table}[H]
\centering
\begin{tabular}{|>{\arraybackslash}m{12.5cm}|>{\centering\arraybackslash}m{1.7cm}|}
\hline
\textbf{Opinion} & \textbf{Label} \\
\hline
\textnormal{Defendant Eric Galaz appeals from an order revoking his probation based on a violation of a condition of probation prohibiting him from possessing at any time, \dots\dots} & majority \\
\hline
I concur in every respect with Justice Minzner s dissenting opinion I agree that the State did not meet its burden at the trial level of establishing facts supporting the \dots\dots & dissent \\
\hline
\end{tabular}
\caption{Legal Opinions Classification Data}
\label{tab:classification-data}
\end{table}
%TODO, move the classification data example to index?
Table-\ref{tab:classification-data} shows an example of the classification data, with truncated legal opinions. On average, the summarization reduces the legal opinions to an average length of around 150 words. The Classification dataset includes about 20K legal opinions, involving approximately 10K for each type of Opinion.

\subsection{Named-Entity-Recognition Dataset}\label{subsec:ner-dataset}

The Legal Opinions' text from the data acquired from CaseLaw, as described in Section-\ref{sec:datasets}, is used to prepare Named Entity Recognition Dataset. We used off-the-shelf NLP toolkits to label the Legal data with Named Entities to create silver standard corpora. These corpora were later merged and corrected in a semi-automatic way to a certain extent. 

\begin{figure}
	\centering
	\includegraphics[width=0.9\linewidth]{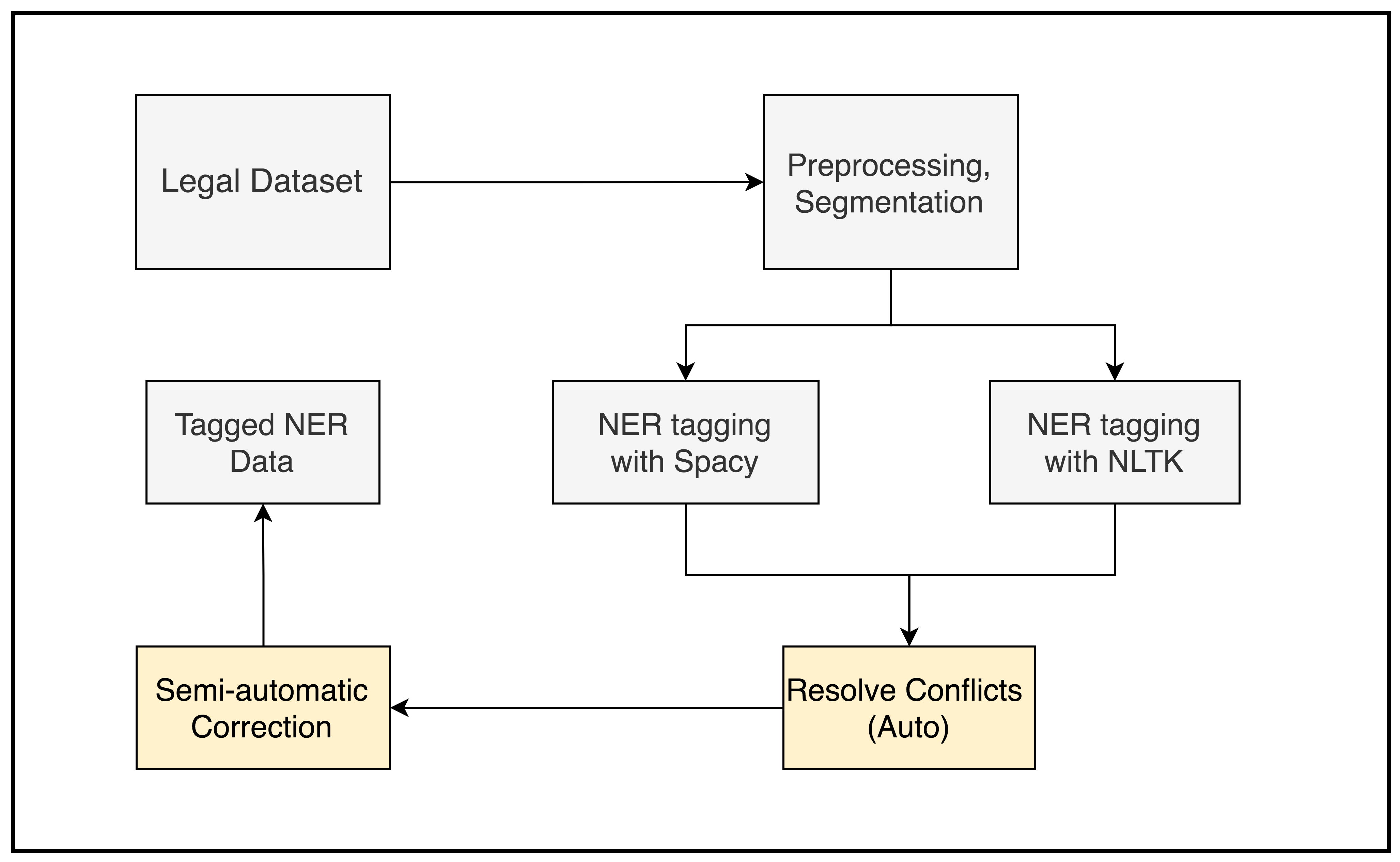}
	\caption{Preparing NER Dataset}
	\label{fig:preparing-ner-data}
\end{figure}

Figure-\ref{fig:preparing-ner-data} shows the complete process for preparing Named-Entity-Recognition data from the Legal Opinions Data. In the first step, text from Legal Opinions is preprocessed and segmented using the Natural Language Toolkit, i.e., NLTK \cite{loper-bird-2002-nltk}. A plain text file is generated from the Legal Opinions, in a format that is similar to the pretraining data, except the blank line to separate the documents/opinions. In the next step, the Legal Data is annotated in an automated way, using two off-the-shelf NLP toolkits, i.e., NLTK and spaCy\footnote{\url{https://spacy.io/}}. NLTK and spaCy provide a simple and handy implementation for multiple NLP tasks. NER module of NLTK is trained on the Automatic Content Extraction (ACE) corpus \cite{Doddington2004TheAC} with Maximum Entropy Model \cite{pinto_et_al:OASIcs:2016:6008}. NLTK achieves an F1-Score of 89\% on the CoNLL NER shared task \cite{ConLL-Task}. To label Legal data with Named Entities using spaCy, we use the English multi-task CNN pretrained on OntoNotes \cite{inbook}, i.e., $en\_core\_web\_sm$\footnote{\url{https://spacy.io/models/en\#en_core_web_sm}}, with an F1-Score of 85.43\%. Once the Legal text is labeled with Named Entities, both Silver standard corpora are merged in an automated way. The automated merging is done using an intuitive mapping between the NER tags of both toolkits, as both toolkits have their own schemes for Named Entity tags, e.g., $ORGANIZATION_{NLTK}$ = $ORG_{SPACY}$, and a simple set of Rules as defined below:

\begin{enumerate}
    \item Retain the Named Entity (NE) label for a token if both toolkits agree on the same type of Named Entity.
    \item Keep the NE label instead of the Non-NE label, i.e., \texttt{O} for a token, if only one of the toolkit labels it as a Named Entity.
    \item If both toolkits label a token as different Named Entities, keep both, and resolve later during the manual correction. 
\end{enumerate}

The cases where toolkits conflicted by labeling a token with different Named Entities were uncommon.

\begin{figure}
	\centering
	\includegraphics[width=0.9\linewidth]{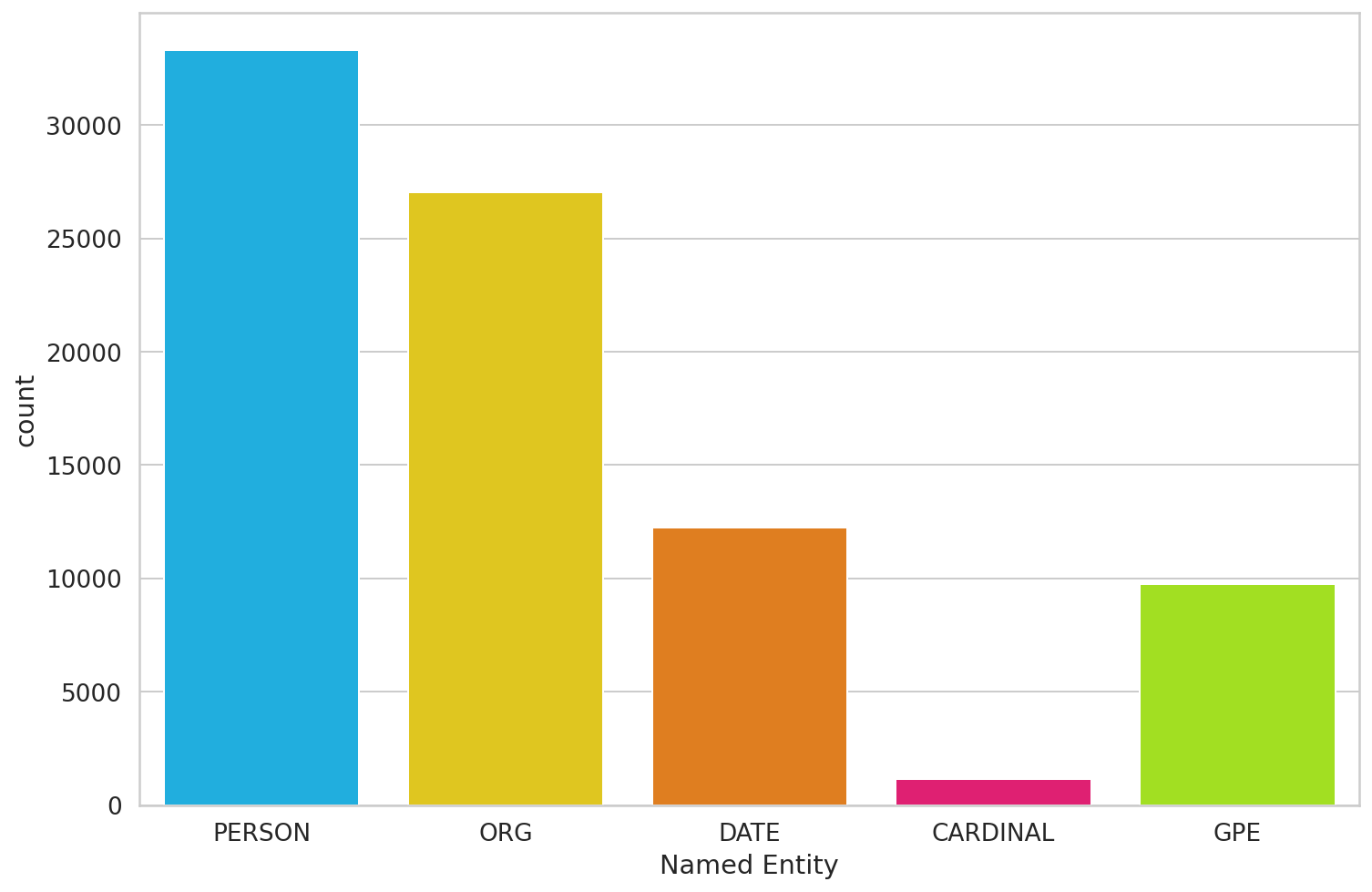}
	\caption{Named Entities Population in NER Dataset}
	\label{fig:ner-population}
\end{figure}

Once the Silver Standard Corpora are merged, we pass it through a recurrent Semi-automatic correction phase. In this phase, first, the disagreements between toolkits are resolved. Afterward, the evident anomalies are observed from the merged and resolved labeled NER data, e.g., \textit{month day, year} not labeled as \texttt{DATE}, or Camel-Cased Name preceded by [President, Sir, Dr.] not labeled as \texttt{PERSON}. These repetitive patterns are observed and corrected using \texttt{regex} in an automated way throughout the Dataset. This process is repeated for all Named Entities types to correct the possible errors in labeling, to get the final Legal Data labeled with Named Entities. The final NER Dataset contains around 59K sentences of Legal text. Figure-\ref{fig:ner-population} shows a bar chart of the population of Named Entities in the prepared NER Dataset.

\subsection{Legal Vocabulary}\label{subsec:legal-vocab}

To prepare Vocab-BERT Model, we prepared a Legal Vocabulary defining 1233 legal terms. The legal terms were extracted by crawling the Legal Dictionary\footnote{\url{https://dictionary.law.com/}} using BeautifulSOUP\footnote{\url{https://www.crummy.com/software/BeautifulSoup/bs4/doc/}}. The Legal Vocabulary\footnote{\url{\SourceCodeRepo\RepoRootPath\LegalVocabulary}} prepared is also released along with the Legal Opinions Classification Dataset. The legal-term-frequencies are also added for the legal terms based on the Classification Dataset. For a legal term defined in the vocabulary, legal-term-frequency is defined as the number of Legal Cases from the Classification Dataset, using the legal term. These statistics are necessary to determine the legal term's eligibility for being added in the Vocab-BERT Model, as discussed in more detail in Section-\ref{subsec:preparing-bert-vocab}.

\section{Transformer-Based Language Models} \label{sec:preparing-models}

The research's main focus is the performance comparison of the Transformer Based Language Models in the Legal Domain. Table-\ref{tab:transformer-models} shows the list of employed Models, with their sizes, including the pretrained Models as released by their authors and the ones modified to adapt the Legal Domain.

\begin{table}[H]
\centering
\begin{tabular}{|>{\arraybackslash}m{3.5cm}|>{\arraybackslash}m{2.5cm}|>{\arraybackslash}m{2.5cm}|}
\hline
\textbf{Model Name}  & \textbf{(L, H, A)}  \\
\hline
BERT-Base-Cased & (12, 768, 12) \\
\hline
XLNet-Base-Cased & (12, 768, 12) \\
\hline
BERT-Medium & (8, 512, 8) \\
\hline
Legal-BERT & (8, 512, 8) \\
\hline
Vocab-BERT & (8, 512, 8) \\
\hline
Legal-Vocab-BERT & (8, 512, 8) \\
\hline
\end{tabular}
\caption{Transformer Models for Comparison}
\label{tab:transformer-models}
\end{table}

Where L represents the number of layers, H represents the size of the hidden states, and A is the number of Attention Heads used in a Model. \textbf{BERT-Base-Cased}, as described in Section-\ref{sec:bert-model}, and \textbf{XLNet-Base-Cased}, as described in Section-\ref{sec:xlnet-model}, are comparable in size and configurations. Both Models use the transformer Neural Network as the underlying architecture and use the benefits of pretraining and transfer learning techniques. The major difference between both Transformer-based Models is their pretraining objectives, i.e., BERT being autoencoder and XLNet being autoregressive. BERT-Base-Cased and XLNet-Base-Cased are used as released by authors to evaluate their performance on NLP tasks in Legal Domain, with simple finetuning.

\textbf{BERT-Medium} is a relatively smaller version of the BERT model compared to BERT-Base-Cased. We choose BERT-Medium because it is comparable to the BERT-Base-Cased Model in size and performance, compared to other smaller versions. The main benefit of using BERT-Medium is the ease in the pretraining process as BERT-Base-Cased requires much more memory and pretraining time than BERT-Medium. For this reason, BERT-Medium is for adaptation in the Legal Domain. The three Models prepared during the research, i.e., Legal-BERT, Vocab-BERT, and Legal-Vocab-BERT, are initialized with BERT-Medium Model and inherit the same size and configurations.

\textbf{Legal-BERT} is initialized with BERT-Medium Model and further pretrained on Legal Data, as described in Section-\ref{subsec:preparing-legal-bert}. \textbf{Vocab-BERT} is BERT-Medium Model fed with Legal Vocabulary, as described in Section-\ref{subsec:preparing-bert-vocab}. \textbf{Legal-Vocab-BERT} is a BERT-Medium Model that combines the powers of both Legal-BERT and Vocab-BERT by adding legal vocabulary to Legal-BERT, as discussed in Section-\ref{subsec:preparing-legal-bert-vocab}.

Our implementation and experimental setup for performing downstream NLP tasks uses the pytorch implementation of the Transformer Models. \textbf{Legal-BERT} is initialized using the tensorflow checkpoint of \textbf{BERT-Medium} Model. The output of the pretraining process is the Legal-BERT Model produced as a tensorflow checkpoint. These Models are first converted into pytorch versions using transformers-cli\footnote{\url{https://huggingface.co/transformers/converting_tensorflow_models.html}} before using on the downstream NLP tasks. Similarly, pytorch versions of \textbf{Vocab-BERT} and \textbf{Legal-Vocab-BERT} are created to be compared on the NLP tasks on legal data.

\subsection{Preparing Legal-BERT} \label{subsec:preparing-legal-bert}

BERT-Medium Model is additionally pretrained on 3B tokens of English Legal text from the CASELAW Access Project for 1M training steps to prepare the Legal-BERT Model. The pretraining process for BERT is discussed in Section-\ref{subsec:pre-training-bert}. BERT is already pretrained on a large amount of unlabelled English text. To adapt BERT to the Legal Domain, we run additional pretraining of the BERT-Medium Model on Legal Data. BERT requires the pretraining data in a particular format. Preparation of Legal Data for pretraining is discussed in Section-\ref{subsec:pre-training-dataset}.

\begin{figure}
	\centering
	\includegraphics[width=0.9\linewidth]{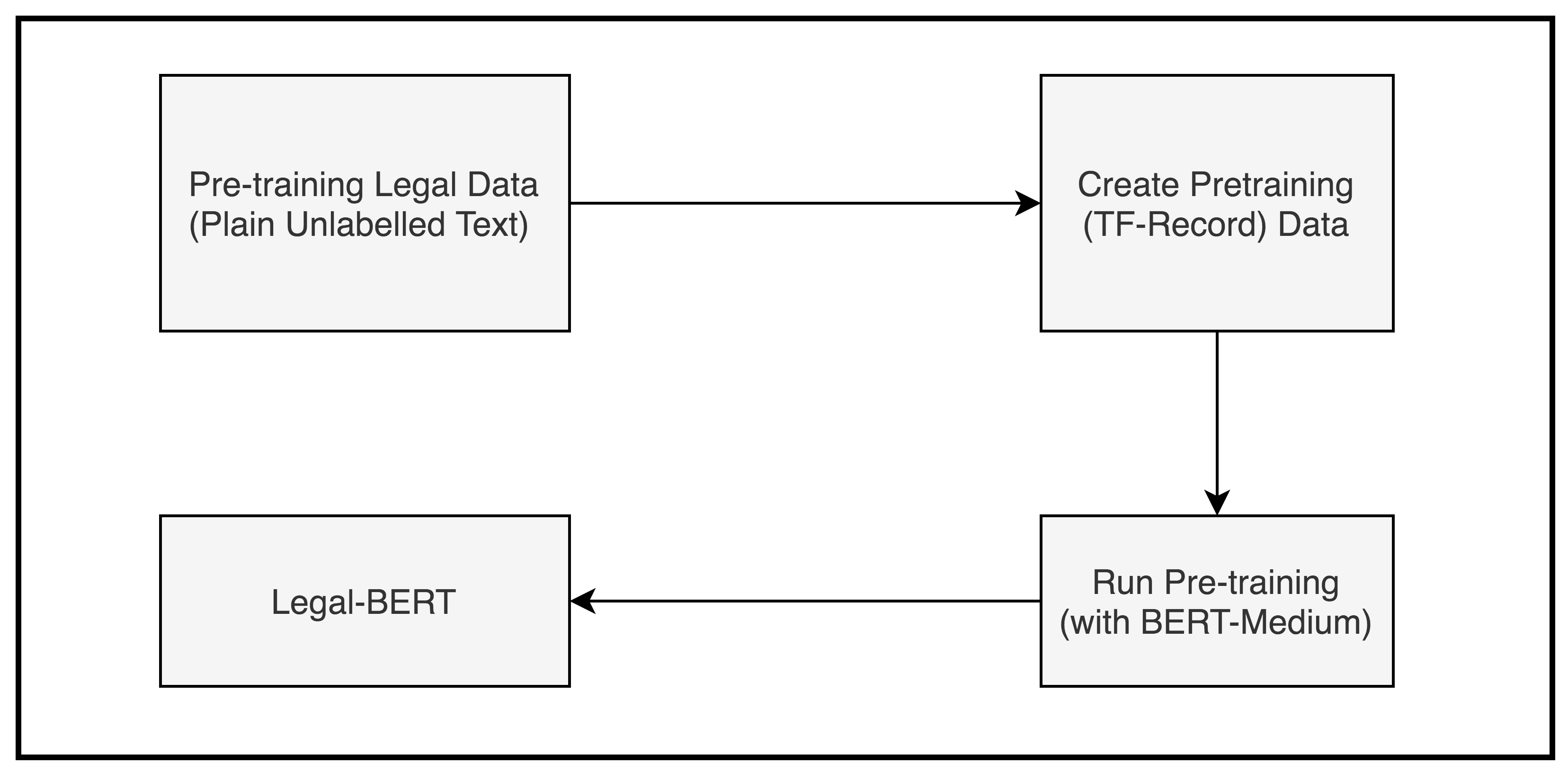}
	\caption{Preparation of Legal-BERT Model}
	\label{fig:preparing-legal-bert}
\end{figure}

Figure-\ref{fig:preparing-legal-bert} shows the preparation of the Legal-BERT Model. The pretraining process involves two major steps, i.e., "Create Pretraining Data" and "Run Pretraining". In the first step, i.e., \textbf{Create Pretraining Data}, the unlabelled text, adherent to pretraining format requirements, is converted into TensorFlow-Record (TF-Record) format. This step's output is the TF-Record files, created in a compatible format with the Size and Configuration of the BERT-Medium Model. The second step, i.e., \textbf{Run Pretraining}, takes the TF-Record Data as input and pretrains the BERT-Medium Model for 1Million training steps. Both steps, i.e., "Create Pretraining Data" and "Run Pretraining", are performed using bash scripts, facilitated by the python code released by the BERT's authors for running the pretraining process for BERT Model. The output of the second step is the Legal-BERT Model. Legal-BERT was pretrained using GPU "Tesla K80" with 11441MiB Memory. The pretraining process, including "Create Pretraining Data" and "Run Pretraining" steps, took approximately eighteen days on the GPU. Legal-BERT achieved $\approx$70\% and $\approx$93\% accuracies on pretraining objectives, i.e., masked language modeling and next sentence prediction, respectively.

Pretraining BERT is an expensive process, both time-wise and resources-wise. Running out of memory is a commonly faced problem in the pretraining process. Following measures and design decisions are taken to avoid such problems: 

\begin{enumerate}
    \item BERT-Medium Model is used to prepare Legal-BERT as it is relatively smaller, much faster than BERT-Base Model, and requires less memory, yet a good candidate for comparison.
    \item Input pretraining text is fragmented into multiple files to avoid loading all the legal data (3B tokens) in the memory at-once.
    \item During the "Create Pretraining Data" process, files are loaded and produced in a batch of 10 files at a time.
    \item The "Run Pretraining" process uses a file glob to read one file at a time from all the TF-Record files.
\end{enumerate}

\begin{table}[H]
\centering
\begin{tabular}{|>{\arraybackslash}m{4.5cm}|>{\arraybackslash}m{2.5cm}|>{\arraybackslash}m{2.5cm}|}
\hline
\textbf{Parameter} & \textbf{Value} \\
\hline
Number of layers (L) & 8 \\
\hline
Hidden size (H) & 512 \\
\hline
Attention heads (A) & 8 \\
\hline
do\_lower\_case & False \\
\hline
train\_batch\_size & 16 \\
\hline
num\_train\_steps & 1000000 \\
\hline
num\_warmup\_steps & 20000 \\
\hline
learning\_rate & 2e-5 \\
\hline
max\_seq\_length & 128 \\
\hline
max\_predictions\_per\_seq & 20 \\
\hline
masked\_lm\_prob & 0.15 \\
\hline
random\_seed & 12345 \\
\hline
dupe\_factor & 5 \\
\hline
\end{tabular}
\caption{Pretraining Hyperparameters}
\label{tab:pretraining-hparams}
\end{table}

Table-\ref{tab:pretraining-hparams} shows the Hyperparameters used for pretraining of the Legal-BERT Model. As Legal-BERT is initialized with the pretrained BERT-Medium Model, hence it inherits its size and configurations. \textbf{do\_lower\_case} is set to \texttt{False} to make the Legal-BERT Model case sensitive. Maximum sequence length, i.e., \textbf{max\_seq\_length}, is used as 128 tokens because, on average, the sentences are segmented around the same length limit in the pretraining data. \textbf{max\_predictions\_per\_seq} shows the maximum number of masked-token predictions BERT should make in a sentence. Masked Language Modeling Probability, i.e., \textbf{masked\_lm\_prob}, is the product of \textbf{max\_seq\_length} and \textbf{max\_predictions\_per\_seq} and is set manually, following the BERT's authors' recommendations.

To verify the impact of maximum sequence length, i.e., \textbf{max\_seq\_length} parameter during the pretraining process, we also pretrained BERT-Medium on the same amount of Legal data and hyperparameters, except increasing the maximum sequence length to 256. The pretraining took more than twice the time taken by using the maximum sequence length as 128. Despite the difference in the maximum sequence length, Legal-BERT$_{256}$ did not show any significant improvements over Legal-BERT$_{128}$ on the downstream NLP tasks. We discard the Legal-BERT$_{256}$ from further analysis as it shows similar performance on the downstream NLP tasks and takes more than twice as much time for the pretraining process.

%write about the pretraining procedure. create pretraining data and run pretraining using bash scripts. write about the inputs and outputs of these processes, and the parameters used. follow scripts.

%write about used data's statistics and trainign time on gpu at chair. and results of pretraining process? 

%converting the model into pytorch model as the next steps in the pipeline use the pytorch implementation of these models from transformer's library of hugging face.

\subsection{Preparing Vocab-BERT} \label{subsec:preparing-bert-vocab}

BERT uses WordPiece Vocabulary to create embeddings, as described in Section-\ref{subsec:wordpiece-vocab}. WordPiece Vocabulary included with BERT Model contains around 30K tokens, prepared from the general English text. BERT-Medium Model is fed with additional vocabulary specific to the Legal Domain to prepare Vocab-BERT Model to adapt to the Legal Domain. The preparation of Legal Vocabulary added to Vocab-BERT Model is discussed in Section-\ref{subsec:legal-vocab}. 

\begin{figure}
	\centering
	\includegraphics[width=0.9\linewidth]{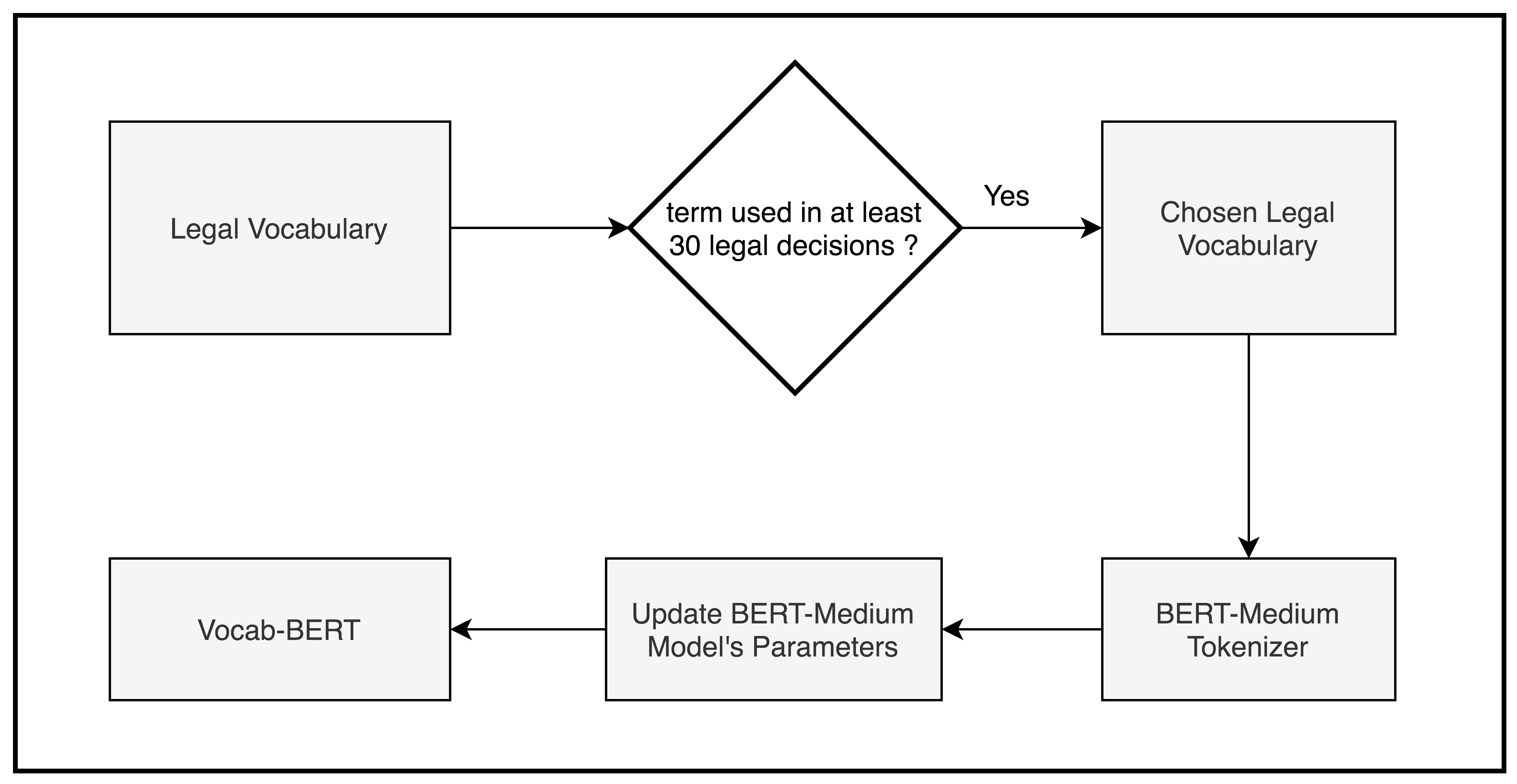}
	\caption{Preparation of Vocab-BERT Model}
	\label{fig:preparing-vocab-bert}
\end{figure}

Figure-\ref{fig:preparing-vocab-bert} shows the preparation of the Vocab-BERT Model. In the first step, Legal Vocabulary is filtered based on its term-frequency in the Legal Data. The term-frequency for a legal term is the number of Legal Opinions using the term in Opinions Classification Dataset. We keep a threshold of 30 Legal Opinions out of the total $\approx$20K population. To add Legal Vocabulary to BERT-Medium for preparing Vocab-BERT Model, a set of legal terms is created by choosing the legal terms with term-frequency equal to or more than 30. It is important to do such filtering as adding more words in the Vocabulary means the size of token embeddings of the BERT Model increases and makes it computationally more expensive. Intuitively, adding new vocabulary that is not used in our target Data would not contribute towards the improvement in BERT's performance on the target NLP tasks. Once the set of legal vocabulary is prepared, it is added to BERT-Medium Model's Tokenizer. The new vocabulary is added in the Tokenizer without splitting as per the WordPiece tokenization mechanism. A legal term is added in the BERT-Medium Model's WordPiece vocabulary if it is not already included. Eventually, 555 new legal terms are added to the WordPiece vocabulary of Vocab-BERT. In the next step, we update the Parameter, i.e., vocab\_size of the BERT-Medium Model. Finally, the vocabulary size and configurations are verified to create the resultant Vocab-BERT Medium Model.

The newly added legal vocabulary in the Vocab-BERT Model affects the tokenization and embeddings generated from the Legal Data. For a legal term that does not already exist in the BERT's WordPiece vocabulary, the BERT tokenizer would greedily break it down until it finds all the tokens in the vocabulary space. However, the same legal-term encountered by Vocab-BERT would be found in its vocabulary and treated as a single token. Following is an example of tokenization and embeddings generated by BERT-Medium and Vocab-BERT, for a legal-term added in Vocab-BERT Model, e.g., "impermissible":

\newpage

\textbf{Input Text}: Ethnic discrimination is impermissible by law.

\textbf{Tokenization}:
\definecolor{light-gray}{gray}{0.95}

\textnormal{BERT-Medium:} \colorbox{light-gray}{Ethnic} \colorbox{light-gray}{discrimination} \colorbox{light-gray}{is} \colorbox{NextBlue!20}{imp} \colorbox{NextBlue!20}{\#\#er} \colorbox{NextBlue!20}{\#\#missible} \colorbox{light-gray}{by} \colorbox{light-gray}{law} \colorbox{light-gray}{.} 

\textnormal{Vocab-BERT-Medium:} \colorbox{light-gray}{Ethnic} \colorbox{light-gray}{discrimination} \colorbox{light-gray}{is} \colorbox{NextBlue!20}{impermissible} \colorbox{light-gray}{by} \colorbox{light-gray}{law} \colorbox{light-gray}{.} 

\textbf{Embeddings}:

\textnormal{BERT-Medium:}
\colorbox{light-gray}{101} \colorbox{light-gray}{27673} \colorbox{light-gray}{9480} \colorbox{light-gray}{1110} \colorbox{NextBlue!20}{24034} \colorbox{NextBlue!20}{1200} \colorbox{NextBlue!20}{27119} \colorbox{light-gray}{1118} \colorbox{light-gray}{1644} \colorbox{light-gray}{119} \colorbox{light-gray}{102}

\textnormal{Vocab-BERT-Medium: }
\colorbox{light-gray}{101} \colorbox{light-gray}{27673} \colorbox{light-gray}{9480} \colorbox{light-gray}{1110} \colorbox{NextBlue!20}{29622} \colorbox{light-gray}{1118} \colorbox{light-gray}{1644} \colorbox{light-gray}{119} \colorbox{light-gray}{102}

%feeding the vocabulary to BERT-Medium, after conversion into pytorch version.

%write down the number of legal terms actually added, the changes made by the new vocabulary in the size of bert model

%must show an example of tokenization via wordpiece to show the difference between before and after adding legal vocabulary in wordpiece vocabulary.

\subsection{Preparing Legal-Vocab-BERT} \label{subsec:preparing-legal-bert-vocab}

\begin{figure}
	\centering
	\includegraphics[width=0.9\linewidth]{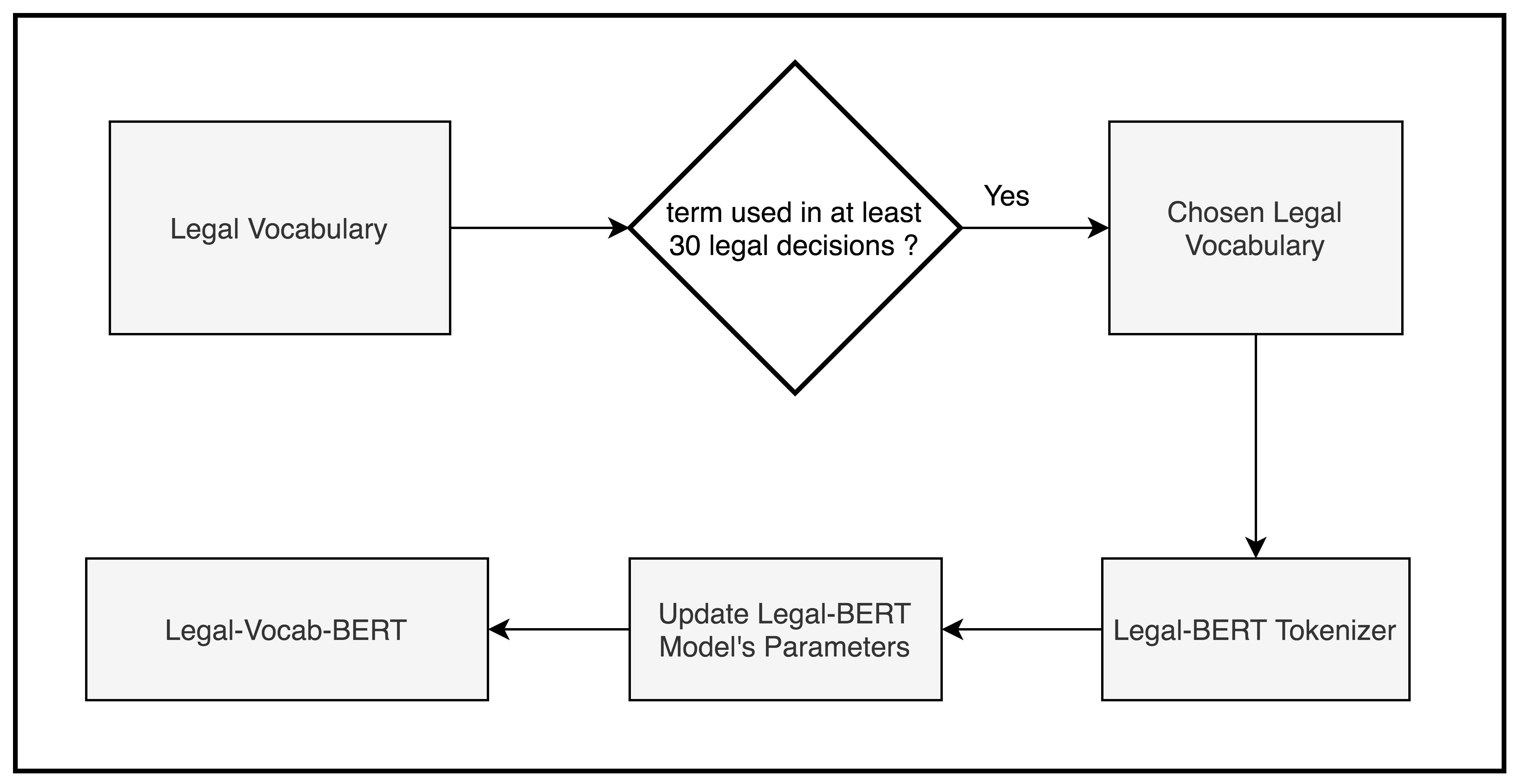}
	\caption{Preparation of Legal-Vocab-BERT Model}
	\label{fig:preparing-legal-vocab-bert}
\end{figure}

Legal-Vocab-BERT is created by combining the mechanisms used to create Legal-BERT and Vocab-BERT as described in Section-\ref{subsec:preparing-legal-bert} and Section-\ref{subsec:preparing-bert-vocab}, respectively. It is equipped with additional pretraining on legal data and additional legal vocabulary. Preparation of Legal-Vocab-BERT is quite similar to the preparation of Vocab-BERT, except that instead of feeding legal vocabulary to BERT-Medium Model, we feed it to Legal-BERT Model prepared by adding pretraining on the Legal Data. After analyzing the impacts of pretraining BERT on legal data, i.e., Legal-BERT, and adding legal vocabulary to BERT, i.e., Vocab-BERT, we combine both the techniques to analyze the combined impact on BERT's adaptation to the Legal Domain.

Figure-\ref{fig:preparing-legal-vocab-bert} shows the preparation of Legal-Vocab-BERT Model. The legal term-frequency filter, as discussed in Section-\ref{subsec:preparing-bert-vocab} for Vocab-BERT, is also applicable to Legal-Vocab-BERT. We prepare a set of legal terms from the Legal Vocabulary (\ref{subsec:legal-vocab}) prevalent in the Legal Opinions Classification Dataset. These legal terms are added in Legal-BERT (\ref{subsec:preparing-legal-bert}) Tokenizer. Next, the vocab\_size of the Legal-BERT Model is updated according to the number of legal-terms added in vocabulary. Finally, vocabulary size and model size are verified to produce the Legal-Vocab-BERT Model.

% write down about the changes needed to combine both strategies to prepare a legal-bert-vocab Model.
%include new picture and describe briefly.

\chapter{Experimentation}\label{chap:experimentation}

\section{Experimental Setup}

% Describe the experimental setup, the used datasets/parameters and the experimental results achieved

Table-\ref{tab:transformer-models} shows six different Transformer \cite{transformer} based Models that are evaluated on Natural Language Processing tasks in the Legal Domain. The pretrained Models XLNet$_{BASE}$, BERT$_{BASE}$, and BERT$_{MEDIUM}$, are used as released by their authors. Legal-BERT, Vocab-BERT, and Legal-Vocab-BERT Models are initialized with BERT$_{MEDIUM}$ and adapted to the Legal Domain. The transformers library from huggingface\footnote{\url{https://huggingface.co/transformers/}} is used to evaluate the Models on NLP tasks. It provides an implementation for most transformer-based models on downstream NLP tasks. BERT$_{BASE}$ and XLNet$_{BASE}$ Models are used from the huggingface repository\footnote{\url{https://huggingface.co/models}} of the Transformer Models. The remaining Models, i.e., BERT$_{MEDIUM}$, Legal-BERT, Vocab-BERT, and Legal-Vocab-BERT, are first converted from tensorflow checkpoints to pytorch Models before evaluation on the NLP tasks, as we use the pytorch implementation of transformer-based models provided by huggingface. All the models were evaluated on the chosen downstream NLP tasks using Tesla T4 GPU on Google Colaboratory\footnote{\url{https://colab.research.google.com/}}.

Experiments are performed multiple times for all the Models on the NLP tasks, on a randomly distributed train, validation, and test Data. The results are taken as an average of ten readings of performance measures for each model and task. The data is split randomly in each assessment to test the model's consistency on any random combination of the input data.

%look into the notebooks for nlp tasks, to get more details on the experimental setup, and reasons behind the parameters and finetuning details.

\section{NLP Tasks in Legal Domain}\label{sec:running-nlp-tasks}

We evaluate all the models on two different NLP tasks:
\begin{enumerate}
    \item Sequence Classification Task of Legal Opinions Classification.
    \item Token Classification Task of Named-Entity-Recognition on Legal Data.
\end{enumerate}

All the models used in the research are evaluated using the same Datasets and finetuning procedure. Following comparisons between different transformer-based Models are made to answer the research questions defined in Section-\ref{sec:research-questions}:
\begin{enumerate}
    \item XLNet$_{BASE}$ is compared with BERT$_{BASE}$ to evaluate the bidirectional Transformer Models with different pretraining objectives.
    \item BERT$_{MEDIUM}$ is compared with Vocab-BERT to analyze the impact of additional Legal Vocabulary on BERT in the Legal Domain.
    \item BERT$_{MEDIUM}$ is compared with Legal-BERT to analyze the impact of additional pretraining on Legal Data on BERT's performance on NLP tasks in the Legal Domain.
    \item Legal-Vocab-BERT, Legal-BERT, and Vocab-BERT are compared with pretrained Models BERT$_{BASE}$ and BERT$_{MEDIUM}$ to analyze how well they adapted to the Legal Domain.
\end{enumerate}

\subsection{Legal Opinions Classification Task}\label{subsec:classification-task}
%describe the pipeline

All the Transformer Models described in Section-\ref{sec:preparing-models} are evaluated on the Legal Opinions Classification Task. The Dataset for this task contains the Opinions from Judges on the Legal Cases from the Jurisdiction of the State of New Mexico of the United-States. The Opinions from Judges can be either dissenting or agree with the majority's decision of the Court, hence labeled as \texttt{dissent} or \texttt{majority}, respectively. Detail on Legal Opinions Dataset is given in Section-\ref{subsec:classification-dataset}. The total number of Opinions in the Classification Dataset are 19,927 and almost evenly divided between dissent and majority type Opinions. Table-\ref{tab:opinions-classification-population} shows the population of Opinions in the Classification Dataset.

\begin{figure}
	\centering
	\includegraphics[width=0.9\linewidth]{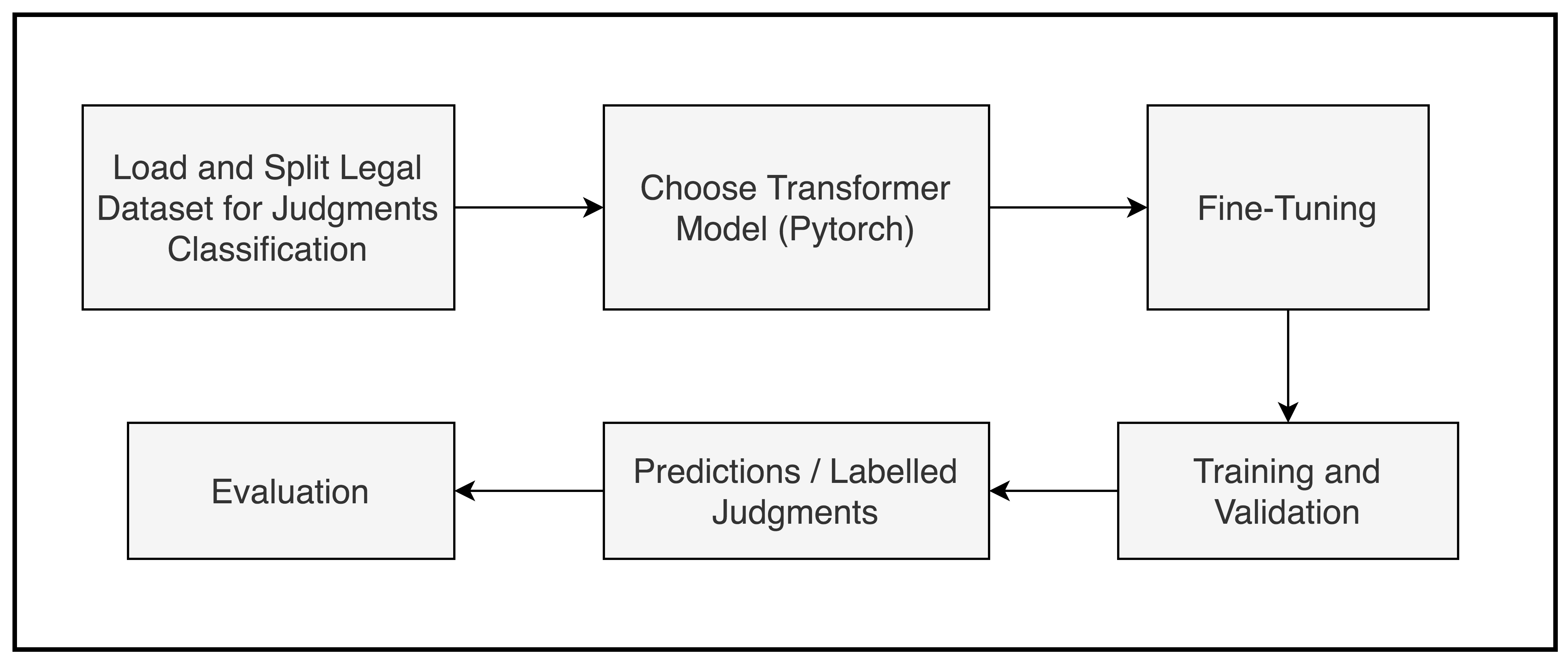}
	\caption{Pipeline for Opinion Classification Task}
	\label{fig:opinion-classification-pipeline}
\end{figure}

\begin{table}[H]
\centering
\begin{tabular}{|>{\arraybackslash}m{3.5cm}|>{\arraybackslash}m{5cm}|}
\hline
\textbf{Opinion Type} & \textbf{Number of Opinions} \\
\hline
dissent & 9995 \\
\hline
majority & 9932 \\
\hline
\end{tabular}
\caption{Opinions Population in Classification Dataset}
\label{tab:opinions-classification-population}
\end{table}

Figure-\ref{fig:opinion-classification-pipeline} shows the pipeline for evaluation of our Models on the Opinions Classification Task. The Opinions Classification Dataset is already preprocessed and summarized to fit the needs of our Models under observation, as described in Section-\ref{subsec:classification-dataset}. The classification data is split into Train (70\%), Validation (15\%), and Test (15\%) datasets. Next in the pipeline is selection of the transformer model for finetuning and evaluation. The finetuning process and hyperparameters are kept the same for each model for fair comparison. An Opinion Classifier is fashioned from the Transformer Model's Sequence Classifier by modifying the fully connected output layer for binary classification. The Model under observation is trained and validated for 10 epochs on the training and validation datasets. Visual comparisons between training and validation accuracies for the models used are available on our repository\footnote{\url{https://gitlab.com/malik.zohaib90/nlp-on-legal-datasets/-/tree/master/nlp/statistics/classification}}. After training and validation, the best performing model is evaluated on its predictions on the Test data. The details on evaluation measures for all the models are given in Section-\ref{subsec:classification-results}.

\begin{table}[H]
\centering
\begin{tabular}{|>{\arraybackslash}m{4.5cm}|>{\arraybackslash}m{2.5cm}|>{\arraybackslash}m{2.5cm}|}
\hline
\textbf{Parameter} & \textbf{Value} \\
\hline
max\_seq\_length & 256 \\
\hline
train\_batch\_size & 16 \\
\hline
num\_train\_steps & 8720 \\
\hline
num\_warmup\_steps & 0 \\
\hline
learning\_rate & 2e-5 \\
\hline
drop\_out & 0.3 \\
\hline
activation\_function & gelu \\
\hline
\end{tabular}
\caption{Finetuning Hyperparameters for Opinions Classification}
\label{tab:finetuning-hparams-op}
\end{table}

Table-\ref{tab:finetuning-hparams-op} shows the finetuning hyperparameters used in the Opinions Classification task for all the models. The maximum sequence length is kept at 256 for the sequence classification task as all the opinions from the Opinions Classification Data are concise to around 150 tokens on average. The batch size is selected as 16. This is the best possible batch size in combination with the maximum sequence length of 256. Given the amount of data, and maximum sequence length, increasing the batch size would result in out-of-memory problem, whereas decreasing it would slow down the training progress significantly. The maximum number of training steps are approximately 8720, based on the size of training data, batch size, and epochs. The dropout probability for the fully connected output layer is set to 0.3. The activation function is used as the default one, i.e., "gelu" for all the models, with a learning rate of 2e-5.

All the Models are trained for 10 epochs to facilitate the analysis of the impact of the pretraining methods and domain adaptation techniques on their computational time. As a stochastic optimization method, for finetuning our models, Adam optimizer, "an algorithm for first-order gradient-based optimization of stochastic objective functions" \cite{adam-optimizer} is used. We use the adam optimizer's implementation provided by huggingface\footnote{\url{https://huggingface.co/transformers/main_classes/optimizer_schedules.html\#adamw-pytorch}}.

\subsection{Named-Entity-Recognition Task} \label{subsec:ner-task}
%describe the pipeline

The Transformer Models described in Section-\ref{sec:preparing-models} are evaluated on Named-Entity-Recognition (NER) Task. The preparation of NER dataset on Legal data is described in Section-\ref{subsec:ner-dataset}. Table-\ref{tab:ner-population} shows the population of Named Entity tags in our NER labeled Legal Data. We use the transformers library from Huggingface for performing NER task using our models. It provides both pytorch and tensorflow implementations for Natural Language Processing with Transformer Models. We use the implementation from Stefan Schweter\footnote{\url{https://github.com/stefan-it/transformers}} (a forked version of transformers library) to facilitate the Named Entity Recognition task on our Legal Data. 

We modified the forked version of transformers library by Stefan Schweter to add classification report for individual Named Entity type. The modified version is available in our repository\footnote{\url{https://gitlab.com/malik.zohaib90/nlp-on-legal-datasets/-/tree/master/resources/transformers_stefan-it}}. Similarly, the transformers library at its version 3.3.0, was tweaked to get around an unhandled exception, which prevented successfully running NER task with XLNet Model. The modified transformers library is available in our resources on the source code repository\footnote{\url{https://gitlab.com/malik.zohaib90/nlp-on-legal-datasets/-/tree/master/resources/transformers_modified}}.

\begin{table}[H]
\centering
\begin{tabular}{|>{\arraybackslash}m{6cm}|>{\arraybackslash}m{4cm}|}
\hline
\textbf{Named-Entity (NE) Type} & \textbf{NE Population} \\
\hline
PERSON & 33293 \\
\hline
ORG & 27050 \\
\hline
DATE & 12252 \\
\hline
GPE & 9753 \\
\hline
CARDINAL & 1151 \\
\hline
\end{tabular}
\caption{Population of NER Dataset}
\label{tab:ner-population}
\end{table}

% The Dataset for this task containing Opinions from Judges on the Legal Cases from the Jurisdiction of State of New Mexico of United States.

\begin{figure}
	\centering
	\includegraphics[width=0.9\linewidth]{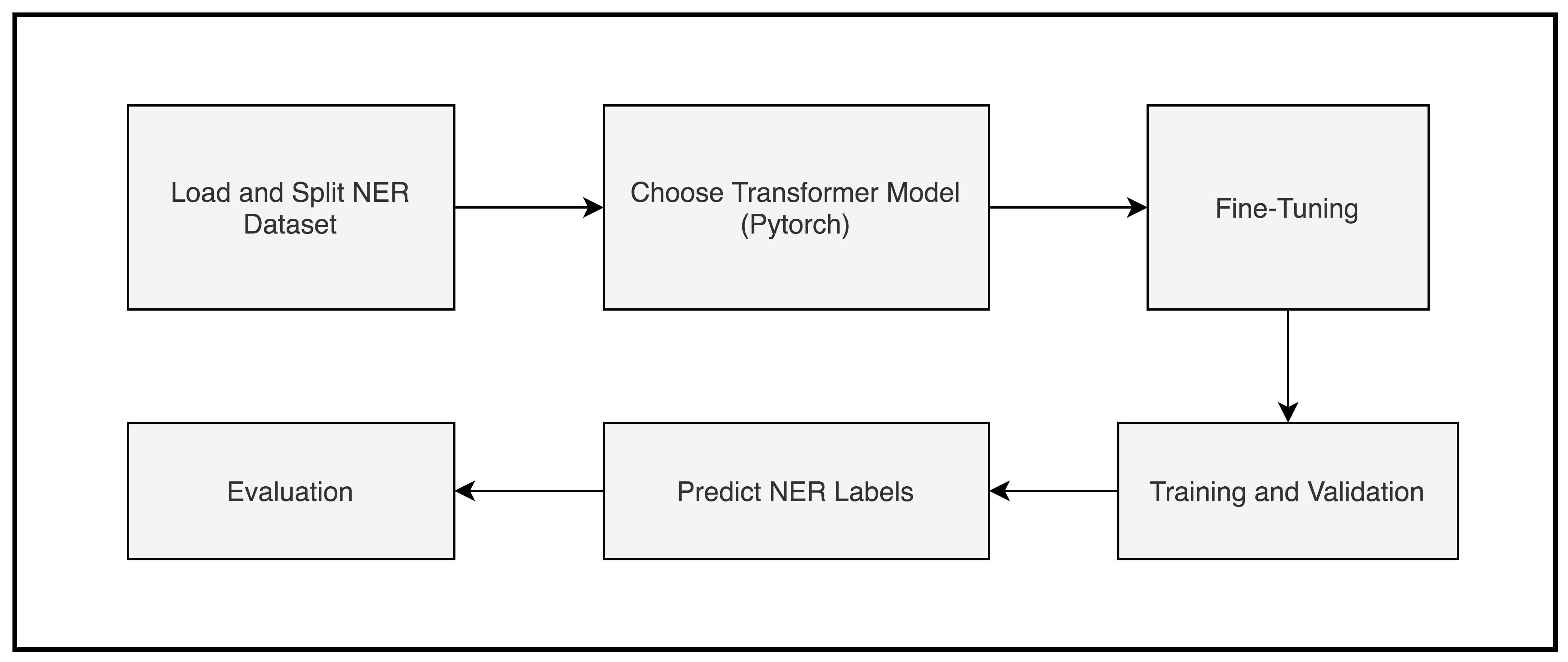}
	\caption{Pipeline for Named Entity Recognition Task}
	\label{fig:ner-pipeline}
\end{figure}

Figure-\ref{fig:ner-pipeline} shows the pipeline for running the Named Entity Recognition task on our Legal Data using the Transformer Models from Table-\ref{tab:transformer-models}. In the first step, the NER tagged legal data is split into Train (70\%), Test (20\%), and Dev (10\%) datasets. In the next step, the model under observation is chosen from the list of Transformer models involved in the comparison. Finetuning parameters are defined for training the Model. The selected model is used to run the NER task using a utility from the forked version of transformers library from stefan-it\footnote{\url{https://github.com/stefan-it/transformers/blob/master/examples/token-classification/run_ner.py}}. The model is trained for five epochs on the NER Training dataset. Finally, the model is evaluated on the predicted Named Entity labels of the Test Dataset. The details on results of NER task for all the models are given in Section-\ref{subsec:ner-results}.

\begin{table}[H]
\centering
\begin{tabular}{|>{\arraybackslash}m{4.5cm}|>{\arraybackslash}m{2.5cm}|>{\arraybackslash}m{2.5cm}|}
\hline
\textbf{Parameter} & \textbf{Value} \\
\hline
max\_seq\_length & 128 \\
\hline
train\_batch\_size & 32 \\
\hline
num\_train\_steps & 6415 \\
\hline
num\_warmup\_steps & 0 \\
\hline
learning\_rate & 5e-5 \\
\hline
weight\_decay & 0 \\
\hline
do\_train & True \\
\hline
do\_eval & True \\
\hline
do\_predict & True \\
\hline
\end{tabular}
\caption{Finetuning Hyperparameters for Named Entity Recognition}
\label{tab:finetuning-hparams-ner}
\end{table}

Table-\ref{tab:finetuning-hparams-ner} shows the Hyperparameters for finetuning transformer models on the Named-Entity-Recognition task. Since the average length for the sentences tagged for Named Entities in our NER Dataset is less than 128 tokens, the  \textbf{max\_seq\_length} is selected as 128. We also verified the impact by performing the NER task using the maximum sequence length as 256, and it didn't show any notable improvements in the evaluation. However, using the maximum sequence length as 256 requires a lot more memory and significantly slows down the models' computation on the NLP task. The \textbf{train\_batch\_size} is set to 32 for the Named-Entity-Recognition task for all the models. As we have decreased the maximum sequence length to half for the NER task, as compared to Opinion Classification task-\ref{subsec:classification-task}, we can increase the batch size to double without worrying about the memory usage. The learning-rate is set to 5e-5, as per the recommendations from the Models' authors. The training and validation are run for five epochs for all the models on training and validation NER data. Given the amount of training dataset, five epochs, and batch size, the model is trained for 6,415 update steps. Since the NER token classification utility from the transformer runs the consecutive stages of training, validation, and testing, we keep all three parameters, i.e., do\_train, do\_eval, and do\_predict, to perform training, validation, and testing with it in a single execution.

%mark the chapter as experiments above, split the chapter after this for results.
\afterpage{\blankpage}
\chapter{Results}\label{chap:results}

\section{Legal Opinions Classification Results}\label{subsec:classification-results}

The transformer Models used in the research for comparison, from Table-\ref{tab:transformer-models}, are evaluated for the Sequence Classification task of Legal Opinions using evaluation measures like accuracy, precision, recall, and f-measure. The population for majority and dissent type of Opinions is fairly even, as shown in Table-\ref{tab:opinions-classification-population}.

\begin{table}[H]
\centering
\begin{tabular}{|>{\arraybackslash}m{4cm}|>{\arraybackslash}m{4cm}|>{\arraybackslash}m{4cm}|}
\hline
\textbf{Model} & \textbf{Accuracy} & \textbf{Training Time} \\
\hline
BERT-Base-Cased & 94.1471 \% & 2 hrs \\
\hline
XLNet-Base-Cased & 94.2140 \% & 3 hrs \\
\hline
BERT-Medium & 93.9130 \% & 40 mins \\
\hline
Legal-BERT & 95.0836 \% & 41 mins \\
\hline
Vocab-BERT & 94.3478 \% & 52 mins \\
\hline
Legal-Vocab-BERT & \textbf{95.2842} \% & 51 mins \\
\hline
\end{tabular}
\caption{Opinions Classification Results}
\label{tab:opinions-classification-results}
\end{table}

Table-\ref{tab:opinions-classification-results} shows the results for all the Models used for evaluation on the Opinions Classification Task. The accuracy and training time for each Model is average from ten observations with the same finetuning hyperparameters and dataset. The training, validation, and test data are randomly split for each evaluation.

\begin{figure}
	\centering
	\includegraphics[width=0.9\linewidth]{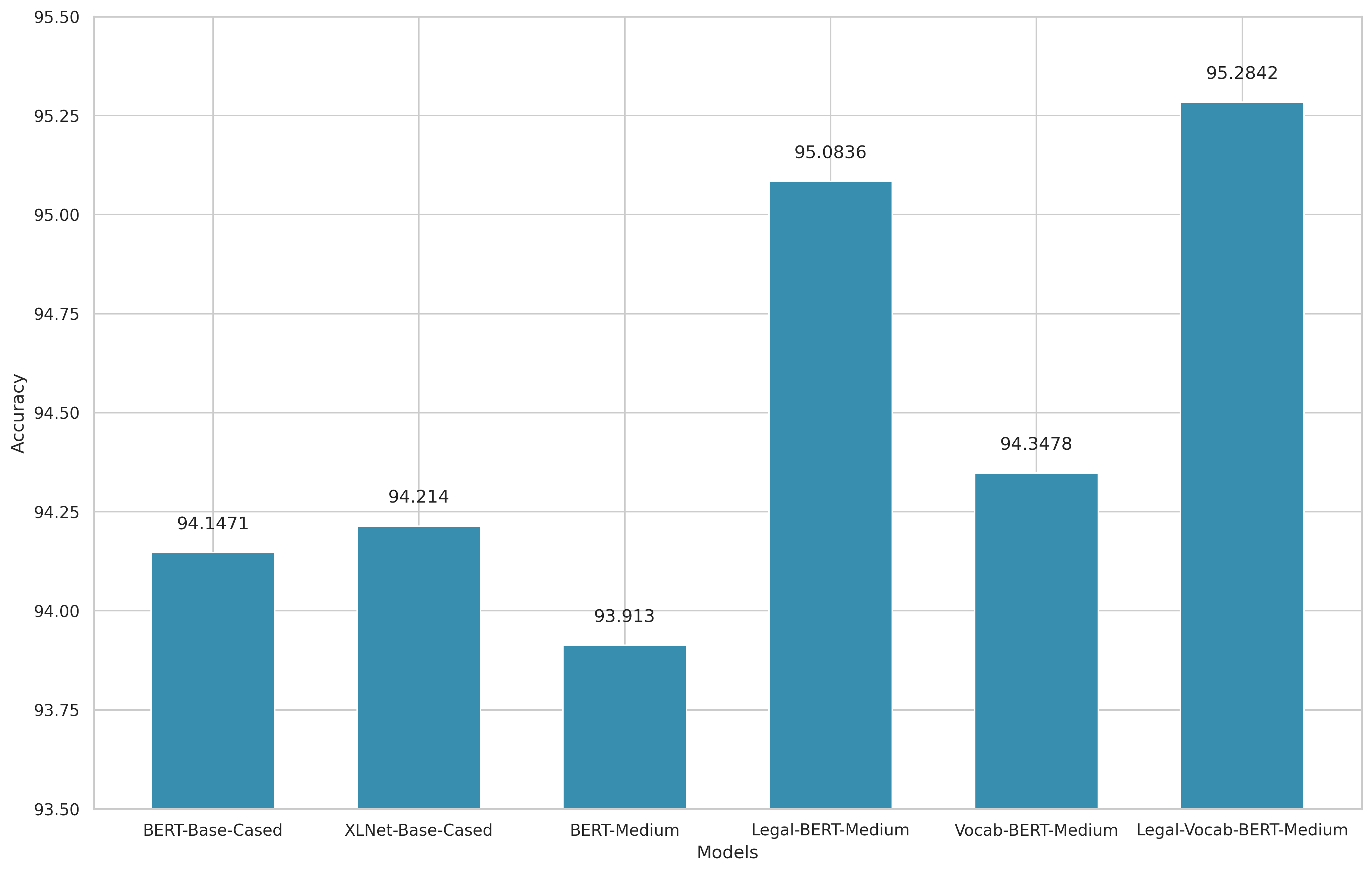}
	\caption{Legal Opinions Classification Results}
	\label{fig:op-classification-accuracies-barchart}
\end{figure}

Figure-\ref{fig:op-classification-accuracies-barchart} shows a comparison of accuracies for different Models on the Opinions Classification Task, using a bar chart. Legal-BERT, Vocab-BERT, and Legal-Vocab-BERT are adapted to the Legal Domain. All the models used in comparison are described in Section-\ref{sec:preparing-models}.

\begin{table}[H]
\centering
\begin{tabular}{|>{\arraybackslash}m{2.5cm} >{\arraybackslash}m{2.5cm} >{\arraybackslash}m{2.5cm} >{\arraybackslash}m{2.5cm} >{\arraybackslash}m{2.5cm}|}
\hline
& \textbf{precision} & \textbf{recall} & \textbf{f1-score} & \textbf{support} \\
\hline
majority & 0.9193 & 0.9656 & 0.9419 & 1452 \\
dissent & 0.9659 & 0.9200 & 0.9424 & 1538 \\
& & & & \\
accuracy & & & 0.9421 & 2990 \\
macro avg & 0.9426 & 0.9428 & 0.9421 & 2990 \\
weighted avg & 0.9433 & 0.9421 & 0.9421 & 2990 \\
\hline
\end{tabular}
\caption{Opinions Classification Report for XLNet-Base-Cased Model}
\label{tab:xlnet-opinions-classification-report}
\end{table}

Table-\ref{tab:xlnet-opinions-classification-report} shows the classification report with different evaluation metrics for the Opinion Classification task using the XLNet-Base-Cased Model.

\begin{table}[H]
\centering
\begin{tabular}{|>{\arraybackslash}m{2.5cm} >{\arraybackslash}m{2.5cm} >{\arraybackslash}m{2.5cm} >{\arraybackslash}m{2.5cm} >{\arraybackslash}m{2.5cm}|}
\hline
& \textbf{precision} & \textbf{recall} & \textbf{f1-score} & \textbf{support} \\
\hline
majority & 0.9468 & 0.9346 & 0.9407 & 1484 \\
dissent & 0.9364 & 0.9482 & 0.9423 & 1506 \\
& & & & \\
accuracy & & & 0.9415 & 2990 \\
macro avg & 0.9416 & 0.9414 & 0.9415 & 2990 \\
weighted avg & 0.9415 & 0.9415 & 0.9415 & 2990 \\
\hline
\end{tabular}
\caption{Opinions Classification Report for BERT-Base-Cased Model}
\label{tab:bert-base-opinions-classification-report}
\end{table}

Table-\ref{tab:bert-base-opinions-classification-report} shows the classification report with different evaluation metrics for the Opinion Classification task using the BERT-Base-Cased Model.

\begin{table}[H]
\centering
\begin{tabular}{|>{\arraybackslash}m{2.5cm} >{\arraybackslash}m{2.5cm} >{\arraybackslash}m{2.5cm} >{\arraybackslash}m{2.5cm} >{\arraybackslash}m{2.5cm}|}
\hline
& \textbf{precision} & \textbf{recall} & \textbf{f1-score} & \textbf{support} \\
\hline
majority & 0.9393 & 0.9380 & 0.9386 & 1484 \\
dissent & 0.9390 & 0.9402 & 0.9396 & 1506 \\
& & & & \\
accuracy & & & 0.9391 & 2990 \\
macro avg & 0.9391 & 0.9391 & 0.9391 & 2990 \\
weighted avg & 0.9391 & 0.9391 & 0.9391 & 2990 \\
\hline
\end{tabular}
\caption{Opinions Classification Report for BERT-Medium Model}
\label{tab:bert-medium-opinions-classification-report}
\end{table}

Table-\ref{tab:bert-medium-opinions-classification-report} shows the classification report with different evaluation metrics for the Opinion Classification task using the BERT-Medium Model.

\begin{table}[H]
\centering
\begin{tabular}{|>{\arraybackslash}m{2.5cm} >{\arraybackslash}m{2.5cm} >{\arraybackslash}m{2.5cm} >{\arraybackslash}m{2.5cm} >{\arraybackslash}m{2.5cm}|}
\hline
& \textbf{precision} & \textbf{recall} & \textbf{f1-score} & \textbf{support} \\
\hline
majority & 0.9407 & 0.9616 & 0.9510 & 1484 \\
dissent & 0.9613 & 0.9402 & 0.9507 & 1506 \\
& & & & \\
accuracy & & & 0.9508 & 2990 \\
macro avg & 0.9510 & 0.9509 & 0.9508 & 2990 \\
weighted avg & 0.9511 & 0.9508 & 0.9508 & 2990 \\
\hline
\end{tabular}
\caption{Opinions Classification Report for Legal-BERT Model}
\label{tab:legal-bert-opinions-classification-report}
\end{table}

Table-\ref{tab:legal-bert-opinions-classification-report} shows the classification report with different evaluation metrics for the Opinion Classification task using the Legal-BERT Model.

\begin{table}[H]
\centering
\begin{tabular}{|>{\arraybackslash}m{2.5cm} >{\arraybackslash}m{2.5cm} >{\arraybackslash}m{2.5cm} >{\arraybackslash}m{2.5cm} >{\arraybackslash}m{2.5cm}|}
\hline
& \textbf{precision} & \textbf{recall} & \textbf{f1-score} & \textbf{support} \\
\hline
majority & 0.9398 & 0.9468 & 0.9433 & 1484 \\
dissent & 0.9472 & 0.9402 & 0.9437 & 1506 \\
& & & & \\
accuracy & & & 0.9435 & 2990 \\
macro avg & 0.9435 & 0.9435 & 0.9435 & 2990 \\
weighted avg & 0.9435 & 0.9435 & 0.9435 & 2990 \\
\hline
\end{tabular}
\caption{Opinions Classification Report for Vocab-BERT Model}
\label{tab:vocab-bert-opinions-classification-report}
\end{table}

Table-\ref{tab:vocab-bert-opinions-classification-report} shows the classification report with different evaluation metrics for the Opinion Classification task using the Vocab-BERT Model.

\begin{table}[H]
\centering
\begin{tabular}{|>{\arraybackslash}m{2.5cm} >{\arraybackslash}m{2.5cm} >{\arraybackslash}m{2.5cm} >{\arraybackslash}m{2.5cm} >{\arraybackslash}m{2.5cm}|}
\hline
& \textbf{precision} & \textbf{recall} & \textbf{f1-score} & \textbf{support} \\
\hline
majority & 0.9450 & 0.9609 & 0.9529 & 1484 \\
dissent & 0.9608 & 0.9449 & 0.9528 & 1506 \\
& & & & \\
accuracy & & & 0.9528 & 2990 \\
macro avg & 0.9529 & 0.9529 & 0.9528 & 2990 \\
weighted avg & 0.9530 & 0.9528 & 0.9528 & 2990 \\
\hline
\end{tabular}
\caption{Opinions Classification Report for Legal-Vocab-BERT Model}
\label{tab:legal-vocab-bert-opinions-classification-report}
\end{table}

Table-\ref{tab:legal-vocab-bert-opinions-classification-report} shows the classification report with different evaluation metrics for the Opinion Classification task using the  Legal-Vocab-BERT Model.

\begin{figure}
	\centering
	\includegraphics[width=0.9\linewidth]{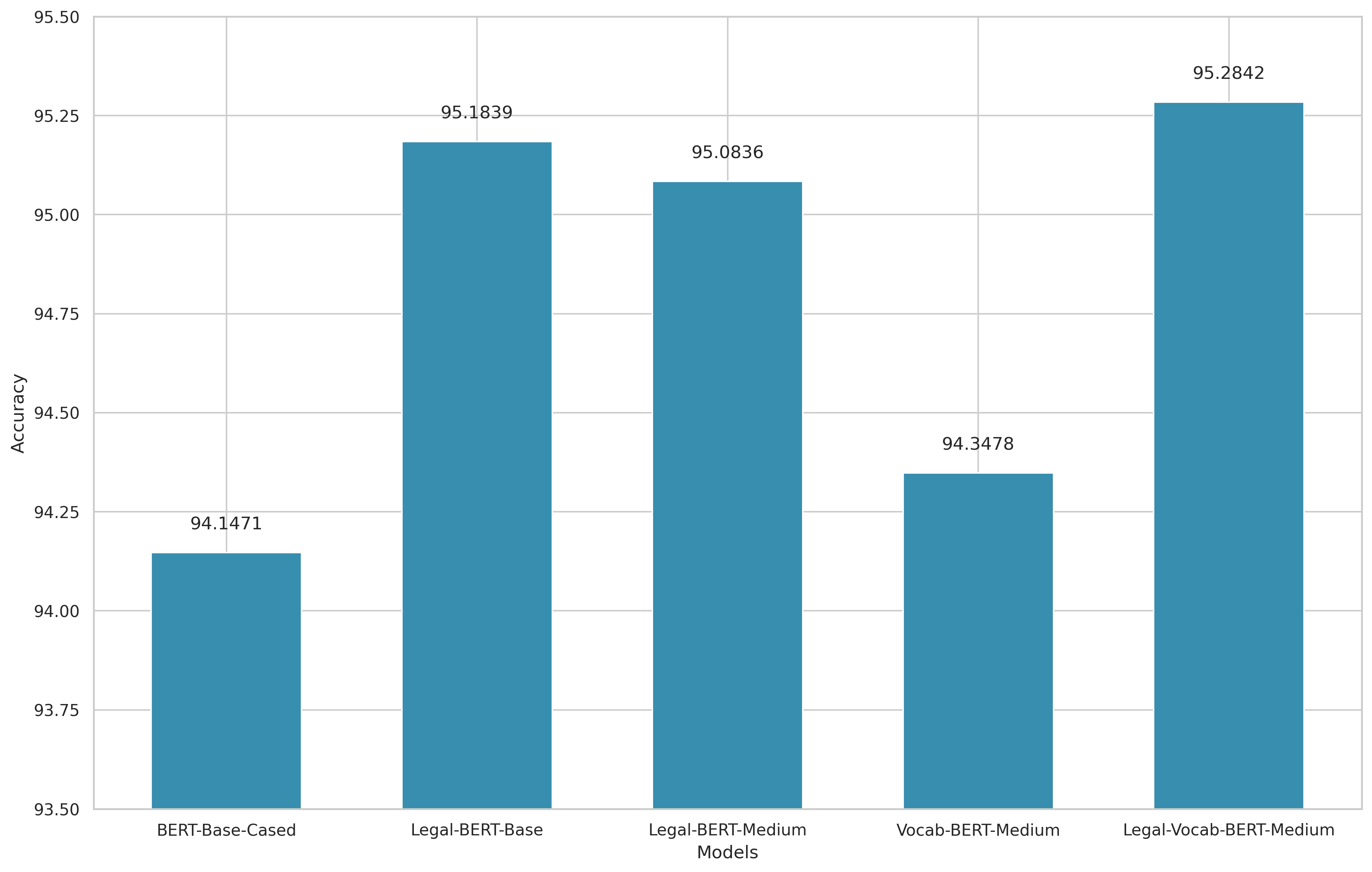}
	\caption{Opinions Classification comparison with Legal-BERT-Base from Chalkidis et al.}
	\label{fig:op-classification-accuracies-barchart-with-muppert-bert}
\end{figure}

Our Legal-BERT, Vocab-BERT, and Legal-Vocab-BERT Models have the same size and configurations as BERT-Medium Model. A Legal-BERT Model recently released by Chalkidis et al. \cite{muppets-legal-bert}, is discussed in the Section-\ref{sec:muppets-legal-bert}. They pretrained the BERT Model of size and configurations same as BERT-Base, on legal data from different sub-domains of law. We compare their Legal-BERT's performance with our Models adept in the Legal Domain. Figure-\ref{fig:op-classification-accuracies-barchart-with-muppert-bert} shows a comparison between our family of BERT models specializing in the Legal Domain, with the \textbf{Legal-BERT-Base} model from Chalkidis et al., on the Opinions Classification Data. BERT-Base-Cased Model in the bar chart is the pretrained general language model. 

The detailed results on the classification task, along with the confusion matrices, can be found in our repository\footnote{\url{https://gitlab.com/malik.zohaib90/nlp-on-legal-datasets/-/blob/master/nlp/results/Classification_Results.md}}. The repository also contains the readings from multiple observations for all the models on the Classification task \footnote{\url{https://gitlab.com/malik.zohaib90/nlp-on-legal-datasets/-/blob/master/nlp/results/best_and_average_scores.txt}}.

%write about the muppet legal-bert model.

%link to the confusion matrices for all the models.

\section{Named-Entity-Recognition Results}\label{subsec:ner-results}

We compare all the transformer Models from Table-\ref{tab:transformer-models} on the token classification task of Named-Entity-Recognition. The population of different Named Entities in the NER Dataset is shown in Table-\ref{tab:ner-population}. Since the distribution of Named Entity tags is not even, and the number of Non-Named-Entity tokens is much higher as compared to the Named Entities, making it largely imbalanced, hence we evaluate the Models using f1-score.

\begin{table}[H]
\centering
\begin{tabular}{|>{\arraybackslash}m{4cm}|>{\arraybackslash}m{4cm}|>{\arraybackslash}m{4cm}|}
\hline
\textbf{Model} & \textbf{F1-Score} & \textbf{Training Time} \\
\hline
BERT-Base-Cased & \textbf{85.6865} \% & 1 hr 18 mins \\
\hline
XLNet-Base-Cased & 85.2921 \% & 1 hr 45 mins \\
\hline
BERT-Medium & 82.8686 \% & 30 mins \\
\hline
Legal-BERT & 82.6935 \% & 29.5 mins \\
\hline
Vocab-BERT & 83.1251 \% & 2 hrs 5 mins \\
\hline
Legal-Vocab-BERT & 82.6127 \% & 2 hrs 6 mins \\
\hline
\end{tabular}
\caption{Named-Entity-Recognition Results}
\label{tab:ner-results}
\end{table}

%f1-score in the table are calculated on the predictions of on the NE tags.

Table-\ref{tab:ner-results} shows the results for all the Models used for evaluation on Named-Entity-Recognition Task on the Test Dataset. The F1-Score and Training Time for each Model is the average of ten observations with the same finetuning hyperparameters and Dataset. The F1-Score is calculated only from the predicted Named-Entity (NE) labels, i.e., excluding predictions of Non-NE tokens.

\begin{figure}
	\centering
	\includegraphics[width=0.9\linewidth]{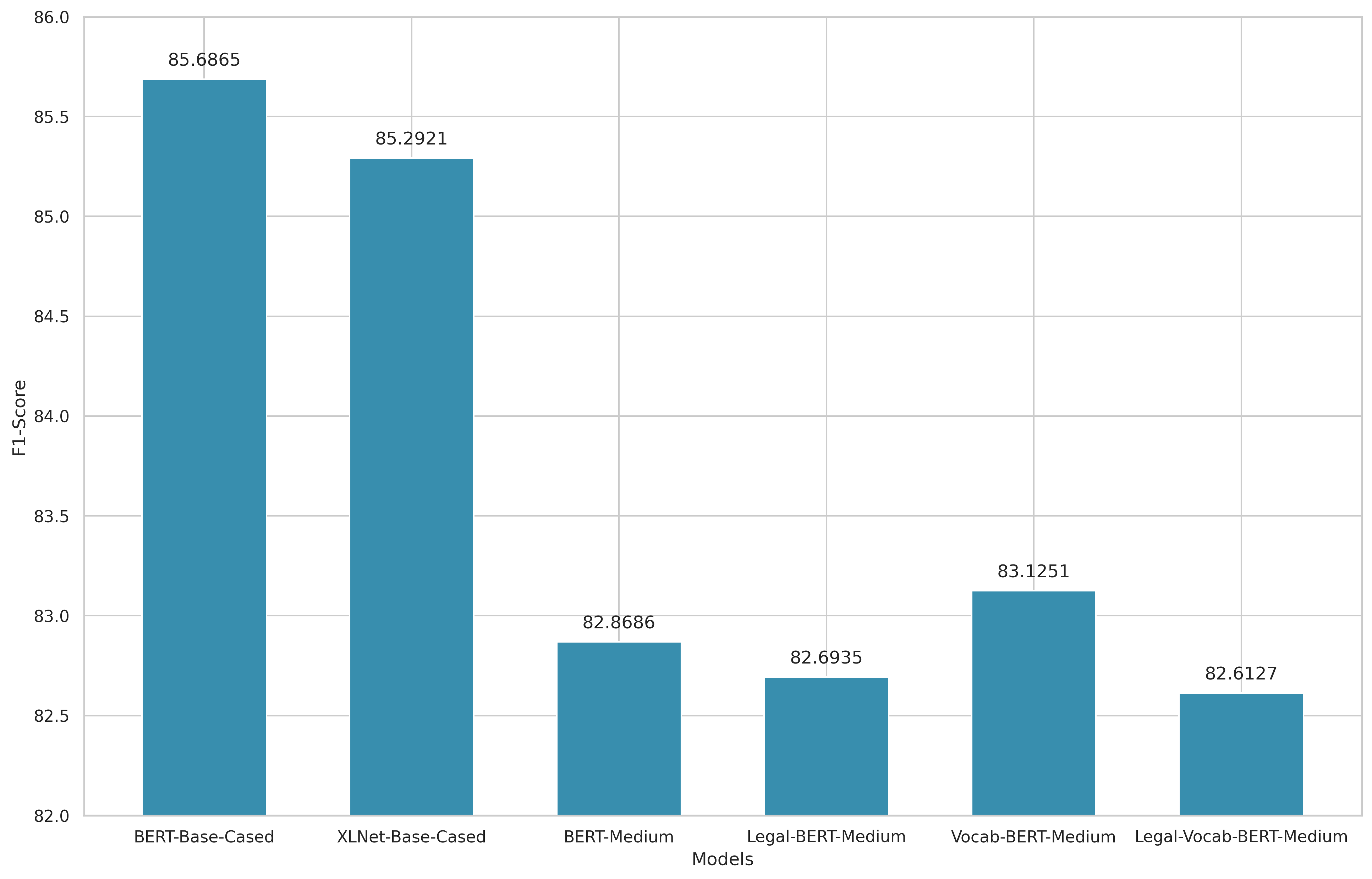}
	\caption{Named-Entity-Recognition Results}
	\label{fig:ner-fscores-barchart}
\end{figure}

Figure-\ref{fig:ner-fscores-barchart} shows a comparison of f1-scores for different Models on Named-Entity-Recognition Task, using a bar chart. Legal-BERT, Vocab-BERT, and Legal-Vocab-BERT are adapted to the Legal Domain. All the models used in comparison are described in Section-\ref{sec:preparing-models}.

\begin{table}[H]
\centering
\begin{tabular}{|>{\arraybackslash}m{2.5cm}|>{\arraybackslash}m{4cm}|>{\arraybackslash}m{4cm}|}
\hline
& \textbf{Validation Dataset} & \textbf{Test Dataset} \\
\hline
loss & 0.0494 & 0.0596 \\
\hline
accuracy & 0.9909 & 0.9897 \\
\hline
precision & 0.8416 & 0.8353 \\
\hline
recall & 0.8416 & 0.8713 \\
\hline
f1\_score & 0.8534 & 0.8529 \\
\hline
\end{tabular}
\caption{NER Results for XLNet-Base-Cased Model}
\label{tab:xlnet-ner-result}
\end{table}

Table-\ref{tab:xlnet-ner-result} shows the results on Validation and Test Data with different evaluation metrics for the Named-Entity-Recognition task using the XLNet-Base-Cased Model.

\begin{table}[H]
\centering
\begin{tabular}{|>{\arraybackslash}m{2.5cm}|>{\arraybackslash}m{4cm}|>{\arraybackslash}m{4cm}|}
\hline
& \textbf{Validation Dataset} & \textbf{Test Dataset} \\
\hline
loss & 0.0492 & 0.0595 \\
\hline
accuracy & 0.9911 & 0.9898 \\
\hline
precision & 0.8448 & 0.8447 \\
\hline
recall & 0.8619 & 0.8694 \\
\hline
f1\_score & 0.8532 & 0.8569 \\
\hline
\end{tabular}
\caption{NER Results for BERT-Base-Cased Model}
\label{tab:bert-base-ner-result}
\end{table}

Table-\ref{tab:bert-base-ner-result} shows the results on Validation and Test Data with different evaluation metrics for the Named-Entity-Recognition task using the BERT-Base-Cased Model.

\begin{table}[H]
\centering
\begin{tabular}{|>{\arraybackslash}m{2.5cm}|>{\arraybackslash}m{4cm}|>{\arraybackslash}m{4cm}|}
\hline
& \textbf{Validation Dataset} & \textbf{Test Dataset} \\
\hline
loss & 0.0660 & 0.0826 \\
\hline
accuracy & 0.9863 & 0.9835 \\
\hline
precision & 0.8290 & 0.8178 \\
\hline
recall & 0.8279 & 0.8399 \\
\hline
f1\_score & 0.8285 & 0.8287 \\
\hline
\end{tabular}
\caption{NER Results for BERT-Medium Model}
\label{tab:bert-medium-ner-result}
\end{table}

Table-\ref{tab:bert-medium-ner-result} shows the results on Validation and Test Data with different evaluation metrics for the Named-Entity-Recognition task using the BERT-Medium Model.

\begin{table}[H]
\centering
\begin{tabular}{|>{\arraybackslash}m{2.5cm}|>{\arraybackslash}m{4cm}|>{\arraybackslash}m{4cm}|}
\hline
& \textbf{Validation Dataset} & \textbf{Test Dataset} \\
\hline
loss & 0.0657 & 0.0838 \\
\hline
accuracy & 0.9861 & 0.9835 \\
\hline
precision & 0.8156 & 0.8176 \\
\hline
recall & 0.8253 & 0.8365 \\
\hline
f1\_score & 0.8204 & 0.8269 \\
\hline
\end{tabular}
\caption{NER Results for Legal-BERT Model}
\label{tab:legal-bert-ner-result}
\end{table}

Table-\ref{tab:legal-bert-ner-result} shows the results on Validation and Test Data with different evaluation metrics for the Named-Entity-Recognition task using the Legal-BERT Model.

\begin{table}[H]
\centering
\begin{tabular}{|>{\arraybackslash}m{2.5cm}|>{\arraybackslash}m{4cm}|>{\arraybackslash}m{4cm}|}
\hline
& \textbf{Validation Dataset} & \textbf{Test Dataset} \\
\hline
loss & 0.0648 & 0.0786 \\
\hline
accuracy & 0.9863 & 0.9838 \\
\hline
precision & 0.8298 & 0.8222 \\
\hline
recall & 0.8286 & 0.8405 \\
\hline
f1\_score & 0.8292 & 0.8313 \\
\hline
\end{tabular}
\caption{NER Results for Vocab-BERT Model}
\label{tab:vocab-bert-ner-result}
\end{table}

Table-\ref{tab:vocab-bert-ner-result} shows the results on Validation and Test Data with different evaluation metrics for the Named-Entity-Recognition task using the Vocab-BERT Model.

\begin{table}[H]
\centering
\begin{tabular}{|>{\arraybackslash}m{2.5cm}|>{\arraybackslash}m{4cm}|>{\arraybackslash}m{4cm}|}
\hline
& \textbf{Validation Dataset} & \textbf{Test Dataset} \\
\hline
loss & 0.0658 & 0.0822 \\
\hline
accuracy & 0.9860 & 0.9838 \\
\hline
precision & 0.8140 & 0.8133 \\
\hline
recall & 0.8218 & 0.8394 \\
\hline
f1\_score & 0.8179 & 0.8261 \\
\hline
\end{tabular}
\caption{NER Results for Legal-Vocab-BERT Model}
\label{tab:legal-vocab-bert-ner-result}
\end{table}

Table-\ref{tab:legal-vocab-bert-ner-result} shows the results on Validation and Test Data with different evaluation metrics for the Named-Entity-Recognition task using the Legal-Vocab-BERT Model.

\begin{figure}
	\centering
	\includegraphics[width=0.9\linewidth]{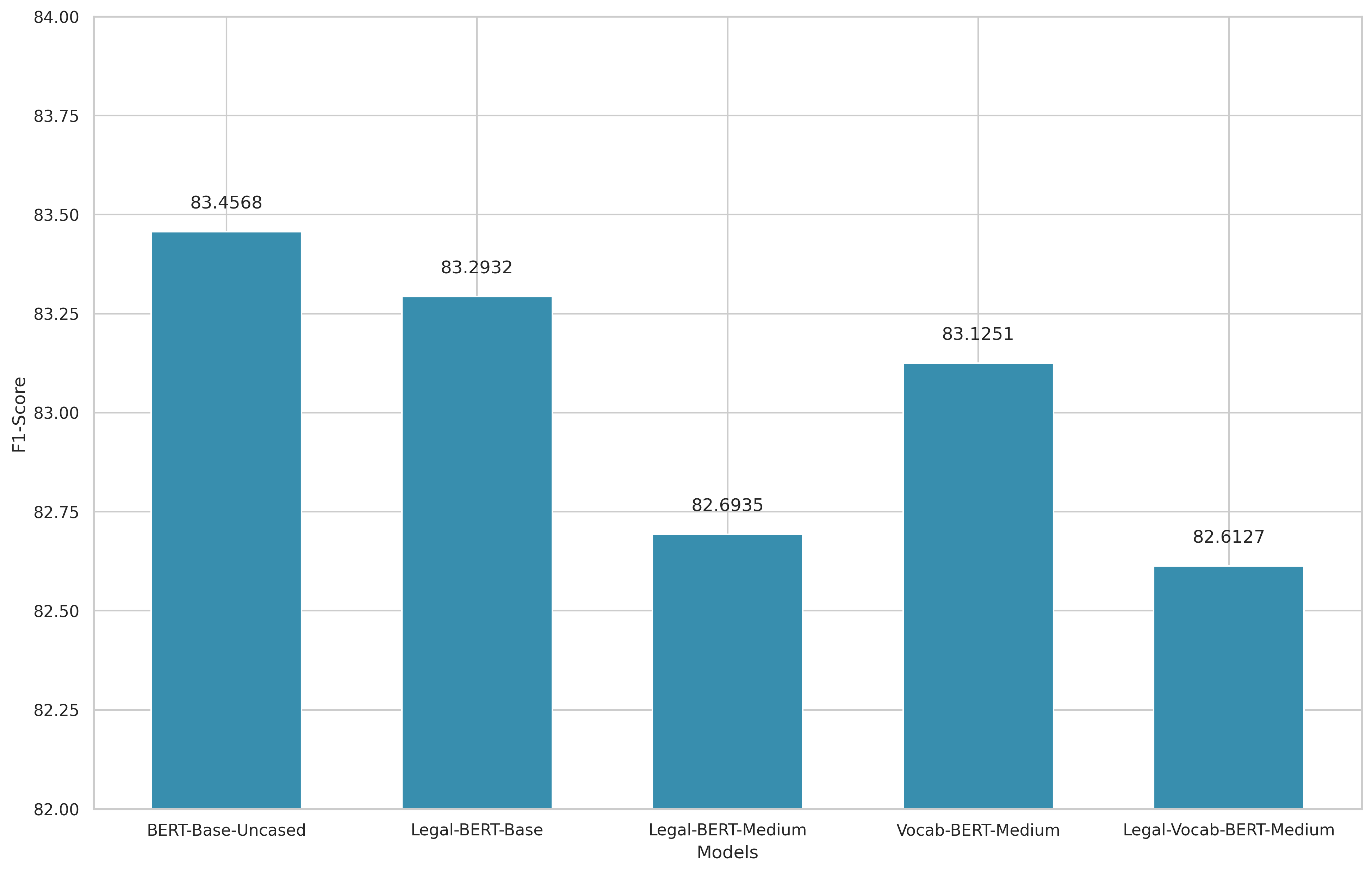}
	\caption{NER comparison with Legal-BERT-Base from Chalkidis et al.}.
	\label{fig:ner-fscores-with-muppets-barchart}
\end{figure}

Just like the Opinion Classification Task, we also evaluated the Legal-BERT-Base Model released by Chalkidis et al. \cite{muppets-legal-bert} on Named-Entity-Recognition Task. Figure-\ref{fig:ner-fscores-with-muppets-barchart} shows a comparison between our family of BERT models specializing in the Legal Domain, with the \textbf{Legal-BERT-Base} model from Chalkidis et al., on the NER Task. BERT-Base-Uncased Model in the bar chart is the pretrained general language model. 

The detailed classification report NER task can be found on our repository\footnote{\url{https://gitlab.com/malik.zohaib90/nlp-on-legal-datasets/-/blob/master/nlp/results/NER_Results.md}}. The repository also contains the readings from multiple observations for all the models on the NER task\footnote{\url{https://gitlab.com/malik.zohaib90/nlp-on-legal-datasets/-/blob/master/nlp/results/best_and_average_scores.txt}}.

\afterpage{\blankpage}
\chapter{Discussion}\label{chap:discussion}

% Discuss the results. What is the outcome of your experimetns?

Based on the experimentation discussed in Chapter-\ref{chap:experimentation} and results shown in Chapter-\ref{chap:results}, the findings and outcomes are discussed in this Chapter that answers all research questions from Section-\ref{sec:research-questions}. Comparisons between different transformer-based models, as listed in Table-\ref{tab:transformer-models}, are discussed. These models include both pretrained models as released by their authors and the models that we customized to adapt to the Legal Domain. All the models are tested on the same datasets, finetuning parameters, and the NLP tasks in the Legal Domain, as discussed in Section-\ref{sec:running-nlp-tasks}. The results from both NLP tasks, i.e., Opinion Classification and NER, are taken as an average from ten observations by randomly splitting the train, validation, and test data.

\section{Autoencoder vs Autoregressive Models}\label{sec:autoencoder-vs-autoregressive}

\textbf{Research Question 1:} Which of the Transformer based Language Models such as BERT and XLNet perform better in Legal Domain?

The transformer-based pretrained models with different pretraining objectives, i.e., BERT-Base-Cased Model (autoencoder) and XLNet-Base-Cased Model (autoregressive) are compared on the Legal Opinions Classification task and Named-Entity-Recognition task on Legal Data. 

\textbf{Legal Opinions Classification Task:}  The results from multiple observations show that BERT Model is more stable in performance over the randomly distributed data. XLNet, on the other hand, shows a variation of $\pm0.8\%$ on randomly split training data. Nonetheless, XLNet Model outperforms BERT Model on the average results taken from multiple observations.

\textbf{NER Task:} The Named-Entity-Recognition task's results, using BERT-Base-Cased and XLNet-Base-Cased Models, show that BERT outperforms XLNet Model on the NER task by a slight margin.

\textbf{Computational Performance:} Given the same amount of training data and size and configurations of the models, XLNet Model takes longer than BERT in performing both Sequence and Token Classification tasks. Besides the computational time, by increasing the input data size, e.g., batch\_size, maximum\_sequence\_length, XLNet Model is likely to run into memory problems before the BERT Model.

% xlnet can run out of memory before bert does, verified on ner task using 256 sequence length 

\section{Impact of Legal Vocabulary}\label{sec:impacts-of-legal-vocab}

\textbf{Research Question 2: }  How does Legal Vocabulary affect the performance of BERT on Legal Tech?

Vocab-BERT Model, discussed in Section-\ref{subsec:preparing-bert-vocab}, is compared with pretrained BERT-Medium Model on the Legal Opinions Classification task and Named-Entity-Recognition task on Legal Data. Vocab-BERT is prepared by feeding the Legal Vocabulary to BERT-Medium Model; hence both models share the same size and configurations. The only difference is that WordPiece Vocabulary of Vocab-BERT includes additional Legal Terms from the Legal Data.

Results from the \textbf{Opinion Classification Task} shows that Vocab-BERT consistently shows improvement in performance over the pretrained BERT-Medium Model. Vocab-BERT also performs slightly better than BERT-Medium Model on \textbf{Named-Entity-Recognition Task}.

An example of tokenization and embedding using BERT-Medium and Vocab-BERT is given in Section-\ref{subsec:preparing-bert-vocab}.  The additional legal vocabulary, which is prevalent in our target Data, provides the Vocab-BERT with a notable edge over the BERT-Medium Model. The improvement comes from the fact that Vocab-BERT can essentially interpret a longer input sequence than the general BERT model. Vocab-BERT shows a slight improvement on NER task with general Named Entities, as opposed to Legal-BERT. It exhibits the potential for improvements for token classes more specific to Law, e.g., Legal Entities.

\textbf{Computational Performance:} Adding more vocabulary to the existing WordPiece vocabulary increases the vocab\_size of the BERT Model. The growing size of vocabulary results in slowing down the Model's training. Vocab-BERT's average training time on Opinion Classification Task showed an increase of approximately 10 minutes when vocabulary is increased by 555 new tokens. The impact on the NER task is even more prominent. Hence it is crucial to add the vocabulary that would contribute to the performance improvements on the target NLP task. It is a necessary trade-off between accuracy and computational performance.

\section{Benefits of Domain Specific Pretraining}\label{sec:benefits-of-domain-specific-pretraining}

\textbf{Research Question 3: }  Is additional pretraining with Legal data on top of an existing pretrained general language model of BERT worthwhile for optimizing NLP performance in Legal Tech?

Legal-BERT Model, discussed in Section-\ref{subsec:preparing-legal-bert}, is compared with the pretrained BERT-Medium Model on the Legal Opinions Classification task and Named-Entity-Recognition task on Legal Data. Legal-BERT is prepared by running additional pretraining on Legal Data by initializing it with BERT-Medium Model; hence both models share the same size, configurations, and WordPiece vocabulary.

\textbf{Opinions Classification Task:} Table-\ref{tab:opinions-classification-results} shows the results of all the models used on Opinions Classification Task. Legal-BERT outperforms the BERT-Medium Model by 1.17\% on the sequence classification task. Results show that additional pretraining of a smaller BERT Model, i.e., BERT-Medium, on Legal Data, makes it well adapted to Legal Domain than a larger general language model like BERT-Base Model, which is bigger and computationally more expensive.

\textbf{NER Task: } Legal-BERT didn't show any significant improvements on Named-Entity-Recognition Task over the BERT-Medium Model. This behavior is also verified with the Legal-BERT-Base Model, released by Chalkidis et al. \cite{muppets-legal-bert}, as shown in Figure-\ref{fig:ner-fscores-with-muppets-barchart}. Since NER data tags are generic, additional pretraining on legal data did not gain an advantage on the task. Legal-BERT specializes in the Legal Domain, hence susceptible to better performance on data labeled with legal terms, e.g., Legal Entities, Crimes, Legal Areas, for a token classification task.

\textbf{Computational Performance:} The pretraining process adjusts the model's weights and does not change the size and configuration of the model's architecture. Results show that additional pretraining does not affect the training time of the BERT Model. With the improvements gained from pretraining on legal data, an architecturally smaller BERT model can surpass a larger model that is pretrained on text from the general language on a downstream Legal Task. As a result, our Legal-BERT Model, with the size and configuration of BERT-Medium, outperforms BERT-Base Model and is computationally much faster at the same time, as shown in the Opinions Classification Task's results in the Table-\ref{tab:opinions-classification-results}.

\section{Combining Legal Vocabulary and Pretraining}\label{sec:combining-legal-vocabulary-pretraining}

Vocab-BERT and Legal-BERT showed promising results on the Legal Tasks. It is evident from the Opinions Classification results that additional pretraining on Legal Data yields better performance than additional legal vocabulary. Legal-Vocab-BERT described in Section-\ref{subsec:preparing-legal-bert-vocab} is prepared by adding legal vocabulary to Legal-BERT, combining the benefits gained by both techniques. Results from the Legal Opinions Classification task show that Legal-Vocab-BERT is pre-eminent among the family of BERT models adapted to the Legal Domain, as shown in Figure-\ref{fig:op-classification-accuracies-barchart-with-muppert-bert}. Since the Legal-Vocab-BERT model includes the same amount of additional legal vocabulary as Vocab-BERT, its computational time is comparable to Vocab-BERT Model, yet exceeding in evaluation measures. 

\subsubsection{Competing with Legal-BERT from Chalkidis et al.} Legal-BERT-Base Model is discussed in Section-\ref{sec:muppets-legal-bert}, which was recently released by Chalkidis et al. \cite{muppets-legal-bert}. Authors of the Legal-BERT-Base Model pretrained a BERT model of the size and configurations of the BERT-Base Model on legal text from different legal areas. We compared their Model to our Legal-BERT model, which has the size and configurations of BERT-Medium, a relatively smaller model. When compared on Legal Opinions Classification task, our Legal-BERT Model competes well, with an accuracy of 95.08\%, as compared to Legal-BERT-Base with an accuracy of 95.18\%. However, our Legal-Vocab-BERT Model, having the size and configurations of BERT-Medium model, outperforms Legal-BERT-Base Model from Chalkidis et al., on Legal Opinions Classification Task.

The reason behind the better performance of our Legal-BERT and Legal-Vocab-BERT Models compared to the Legal-BERT-Base model from Chalkidis et al., given the size differences, is the idea of "Sub-Domain pretraining". A Model pretrained within a sub-domain of Law outperforms a Model pretrained on text from different sub-domains of Law when performing an NLP task within the given sub-domain. As explained in Section-\ref{sec:muppets-legal-bert}, the Legal-BERT-Base model is pretrained on legal text from several sub-domains of Law, like legislation, court cases, and contracts. In contrast, our Legal-BERT and Legal-Vocab-BERT Models are pretrained on legal data from the US court cases. The Legal-BERT-Base model prepared by Chalkidis et al. contains 37.3\% (4.3 GB out of total 11.5 GB) of the pretraining text from court cases, whereas our Legal-BERT and Legal-Vocab-BERT Models are pretrained on 100\% text from US court cases (8.6 GB). The rationale of sub-domain expertise is also proved by Chalkidis et al. in their paper, where multiple variants of the Legal-BERT model specializing in a sub-domain are compared. The authors release multiple variants of the BERT Models specializing in different sub-domains of Law; however, they did not release a model specializing in US court cases.

\afterpage{\blankpage}
\chapter{Conclusion}\label{chap:conclusion}

% Summarize the thesis and provide a outlook on future work.  

The recent developments in Language Modeling offer a lot of improvements in the field of Natural Language Processing. The invention of the Transformer Neural Network, together with unsupervised pretraining and transfer learning, paved the way for BERT and XLNet Models to achieve state-of-the-art results on various NLP tasks. While these models provide excellent results, further efforts can make them adapt better to a specific domain. We prepared variants of the BERT Model, specializing in the Legal Domain. During the research, we compared these models on multiple NLP tasks in the Legal Domain. We reached the following conclusions based on these models' evaluation, including the pretrained models, as released by their authors and the models specializing in the Legal Domain prepared during the research.

\subsubsection{BERT vs XLNet}

We compared BERT-Base-Cased and XLNet-Base-Cased Models on Legal Opinions Classification task and NER task on legal data from US court cases. Results show that XLNet outperforms BERT on the sequence classification task of Legal Opinions. However, it is observed that BERT is more stable on the task through multiple experiments. XLNet, on the other hand, deviates by $\pm0.8\%$ in accuracy over randomly distributed training and test datasets. Results from the token classification task show that BERT performed better on the NER tagged Legal data compared to XLNet Model. Besides the evaluation metrics, BERT is faster than XLNet in the training process, and XLNet is more likely to run into memory issues than BERT when the input data size is increased.

\subsubsection{Pretraining BERT on Legal Domain}

% We prepared Legal-BERT Model by pretraining BERT-Medium Model additionally on legal text for 1M training steps. 
Legal-BERT showed promising results in the Legal Opinions Classification task, achieving an improvement of 1.17\% over the BERT-Medium Model, which is a general language model. The additional pretraining did not affect the computational performance of BERT Model on the downstream NLP tasks. Our Legal-BERT outperformed BERT-Base Model on the Legal Opinions task despite being smaller in size. It shows that Legal-BERT has adapted better in the Legal Domain than its larger, and computationally expensive, general language variant, i.e., BERT-Base. The NER task results depicted that additional pretraining on BERT did not show any improvements on the Task. A token classification task, specific to Law, e.g., Legal Entities Data, will be better suited for the evaluation.

\subsubsection{Adding Legal Vocabulary in BERT}

Additional Legal Vocabulary helps in achieving improvements on the Legal Tasks. The results from both Legal Opinions Classification and NER tasks showed that Vocab-BERT achieved improvements over the BERT-Medium Model. Vocabulary selection is crucial and must be task-specific, as additional vocabulary slows down BERT's computational performance due to increased vocabulary size. Intuitively, non-existent or less prevalent vocabulary would not positively affect the Model's performance on the target task. Combining the techniques of additional pretraining and legal vocabulary produces better results than models enabled with individual techniques. The Legal-Vocab-BERT model, a combination of Legal-BERT and Vocab-BERT models, produced the best results on the Legal Opinions Classification Task.

Overall, the research questions are answered in the paper by evaluating all the models prepared and used in the research. However, the work opens a path to some additional possibilities for the Model's better adaptation in the Legal Domain. Research shows that additional pretraining on Legal data from US Court Cases makes it better adapted to the Legal Opinions Classification task. A variant of Legal-BERT Model can be prepared by pretraining it on the Legal Data from US Court Cases from scratch. Comparing it with our Legal-BERT Model will answer if the missing learning from other domains positively impacts the Model's performance.

For the preparation of Vocab-BERT Model, additional Legal Vocabulary is included in the existing WordPiece vocabulary of the BERT Model. Although it helps in improving the accuracy, it slows down the computational performance. A variant of Vocab-BERT can be prepared by creating the WordPiece vocabulary from the Legal Data, by keeping the same vocabulary size as the WordPiece vocabulary that comes with pretrained BERT. WordPiece vocabulary can be initialized by the Legal Vocabulary prepared, and then continuing with the remaining vocabulary generated from Legal text to combine both techniques. Comparing it with our Vocab-BERT will show which of the variants perform better on the Legal Tasks.

Legal Opinions Classification's results showed that Legal-BERT outperformed BERT-Base Model, despite being smaller in size and configurations. In the future, with sufficient resources, variants of larger Legal-BERT models can be prepared by pretraining on the Legal Data from US Court Cases. Evaluation of larger Legal-BERT variants will indicate if we can push the accuracy further on the Legal Task. 

After the successful results of BERT's domain adaptation, it would be interesting to create similar variants of the XLNet Model. Comparing them with Legal-BERT, Vocab-BERT, and Legal-Vocab-BERT models would show which model have better domain adaptation capabilities.

It is evident from the research that domain adaptation techniques help in improving results within the Legal Domain. It still has lots of possibilities to enhance these models' understanding of the language of Law even better.

%future work:

% 1: pre-train bert from scratch on case law legal text, instead of initializing from a general english language bert model.

% 2: create wordpiece legal vocabulary on legal datasets being used and use that instead of WordPiece vocabulary+Special Legal Vocabulary we created. 
% Text taken from: https://arxiv.org/pdf/1903.10676.pdf      Vocabulary BERT uses WordPiece (Wu et al. , 2016) for unsupervised tokenization of the input text. The vocabulary is built such that it contains the most frequently used words or subword units. We refer to the original vocabulary released with BERT as BASEVOCAB. We construct SCIVOCAB, a new WordPiece vocabulary on our scientific corpus using the SentencePiece library. We produce both cased and uncased vocabularies and set the vocabulary size to 30K to match the size of BASEVOCAB. The resulting token overlap between BASEVOCAB and SCIVOCAB is 42%, illustrating a substantial difference in frequently used words between scientific and general domain texts.

% 3: User a larger bert model to push f1 score further if possible.

% write acknowledgements at the end? The research data from case.law. The GPU from university, google colab gpus.

% understanding points for paper: judgment and opinion means the same. Legal-BERT implies BERT additionally pretrained on Legal Data. Vocab-BERT implies the vocabulary used to feed bert is legal vocabulary. Legal-Vocab-BERT implies Legal-BERT fed with Legal Vocabulary.

\afterpage{\blankpage}

% -- Appendix (optional)
\begin{appendices}
    % !TeX spellcheck = en_US
% !TeX encoding = UTF-8
\chapter{Abbreviations}

\begin{table}[H]
\begin{tabular}{ll}
BERT & Bidirectional Encoder Representations from Transformers \\
Transformer-XL & Transformer - (Extra Long) \\
XLNet & Extra Long Network (Named after Transformer-XL) \\
NLP & Natural Language Processing \\
RNN & Recurrent Neural Network \\
LSTM & Long Short Term Memory \\
CNN & Convolutional Neural Network \\
GPT & Generative Pre-trained Transformer \\
ELMo & Embeddings from Language Models \\
ULMFiT & Universal Language Model Fine-tuning \\
GPU & Graphics Processing Unit \\
TPUs & Tensor Processing Units \\
LM & Language Modeling \\
MLM & Masked Language Modeling \\
AE & Autoencoding \\
AR & Autoregressive \\
NADE & Neural  Autoregressive  Distribution  Estimation \\
XML & Extensible Markup Language \\
JSON & JavaScript Object Notation \\
NE & Named-Entity \\
Non-NE & Non Named-Entity \\
NER & Named-Entity-Recognition \\
NLTK & Natural Language Toolkit \\
ACE & Automatic Content Extraction \\
MiB & Mebibyte \\
TF-Record & TensorFlow Record \\
GELU & Gaussian Error Linear Unit \\
% LR & Logistic Regression \\
% SVC & Support Vector Classification \\
% tf-idf & tf-idf
\end{tabular}
\end{table}

\chapter{Code}
The Source Code contains implementation for all the datasets and legal vocabulary preparation, running pretraining with BERT, feeding BERT with legal vocabulary, and running Token and Sequence Classification tasks, with all the models used in research. Below is the link to Source Code Repository:

\url{https://gitlab.com/malik.zohaib90/nlp-on-legal-datasets}

The Models prepared during the research are available at: 

\url{https://gitlab.com/malik.zohaib90/nlp-on-legal-datasets/-/tree/master/models}

\chapter{Dataset}

The datasets prepared for experimentation on different NLP tasks performed during the research, are available in our repository.

\textbf{NER Dataset:} 

\url{https://gitlab.com/malik.zohaib90/nlp-on-legal-datasets/-/tree/master/data/ner}

\textbf{Opinions Classification Dataset:}

\url{https://gitlab.com/malik.zohaib90/nlp-on-legal-datasets/-/tree/master/data/classification}

The Legal Vocabulary prepared for preparation of Vocab-BERT is available at:

\url{https://gitlab.com/malik.zohaib90/nlp-on-legal-datasets/-/tree/master/data/legal_vocabulary}
\end{appendices}
\newpage

%%%%%%%%%%%%%%%%%%%%%%%%%%%%%%%%%%%%%%%%%%%%%%%%%%%%%%%%%%%%%%%%%%%%%%%%%%%%%%%%%%%%%%%%%
\backmatter

% -- Bibliography
\printbibliography

% -- Eidesstattliche Erklärung (= Affadavit)
% \afterpage{\blankpage}

% \include{includes/eidesstattlicheErklaerung}

\end{document}